\documentclass[10pt]{wlscirep}
\usepackage[utf8]{inputenc}
\usepackage[T1]{fontenc}
\usepackage{algorithmic}
\usepackage[ruled]{algorithm2e}
\usepackage{bm}
\usepackage{chngcntr}
\usepackage{threeparttable}
\usepackage{amsmath}
\usepackage{hyperref}
\usepackage{xr-hyper}
\usepackage{cleveref}
\usepackage{booktabs}
\usepackage{amssymb}
\usepackage{pdflscape}
\usepackage{makecell}
\usepackage{array}

\newcommand{\SuppTableSimple}{\fontsize{9.0}{11.0}\selectfont}
\newcommand{\SuppTableStandard}{\fontsize{8.0}{9.5}\selectfont}
\newcommand{\SuppTableCompact}{\fontsize{7.5}{9.0}\selectfont}
\newcommand{\SuppTableDense}{\fontsize{7.0}{8.5}\selectfont}
\newcommand{\SuppTableNotes}{\fontsize{7.2}{8.7}\selectfont}

\crefname{table}{}{}
\Crefname{figure}{}{}

\title{FOCUS: A Subspecialty Foundation Model for Intraoperative Central Nervous System Pathology}

\author[1,*]{Xing Liu} 
\author[2,3,*]{Yifan Guo}
\author[2,3,*]{Hongtai Lv} 
\author[4,*]{Lei Lou} 
\author[2,3,*]{Yuyi Tan} 
\author[1]{Jiang Du} 
\author[1]{Xijia Zhang} 
\author[1]{Ming Zhang} 
\author[4]{Chao Pang} 
\author[4]{Minglei Li} 
\author[5]{Wei Liu} 
\author[2]{Xiaobo Zhu} 
\author[2]{Song Zhang} 
\author[2,6]{Rong Yu} 
\author[2,3]{Hanlin Shu} 
\author[2,3]{Wenzhuo Xu} 
\author[1]{Tao Jiang}
\author[7,$\#$]{Cheng Lu} 
\author[2,8,$\#$]{Jie Tian} 
\author[1,$\#$]{Qing Chang} 
\author[2,3,$\#$]{Zhenyu Liu}

\affil[1]{Department of Neuropathology, Beijing Neurosurgical Institute, Capital Medical University, No. 119 South Fourth Ring West Road, Fengtai District, Beijing,100070, P.R.China.}
\affil[2]{CAS Key Laboratory of Molecular Imaging, Institute of Automation, Chinese Academy of Sciences, No. 95 Zhongguancun East Road, Haidian District, Beijing, 100190, P.R. China. }
\affil[3]{School of Artificial Intelligence, University of Chinese Academy of Sciences, No. 1 Yanqihu East Road, Huairou District, Beijing, 101408, P.R. China. }
\affil[4]{Department of Pathology, The First Hospital of Hebei Medical University, No. 89 Donggang Road, Shijiazhuang City, Hebei Province, 050031, P.R. China.}
\affil[5]{Department of Neurosurgery, The First Hospital of Hebei Medical University, No. 89 Donggang Road, Shijiazhuang City, Hebei Province, 050031, P.R. China. }
\affil[6]{School of Advanced Interdisciplinary Sciences, University of Chinese Academy of Sciences, No. 1 Yanqihu East Road, Huairou District, Beijing, 101408, P.R. China.}
\affil[7]{Guangdong Provincial People's Hospital, No. 106 Zhongshan 2nd Road, Guangzhou, Guangdong Province, 510180, P.R. China.}
\affil[8]{Beijing Advanced Innovation Center for Big Data-Based Precision Medicine, School of Engineering Medicine, Beihang University, No. 37 Xueyuan Road, Haidian District, Beijing 100191, P.R. China.}

\affil[*]{These authors contributed equally to this work.}
\affil[$\#$]{Correspondence: zhenyu.liu@ia.ac.cn (Zhenyu Liu), changqing@bjtth.org(Qing Chang), tian@ieee.org(Jie Tian), lucheng@gdph.org.cn(Cheng Lu)}

\begin{abstract}
Intraoperative frozen-section diagnosis of central nervous system (CNS) lesions guides surgical decisions but is challenged by artefacts, diagnostic complexity and time constraints. Most pathology foundation models have limited exposure to CNS frozen tissue. Here we present FOCUS, a subspecialty foundation model pretrained on 27,930 frozen-section whole-slide images enriched for CNS tissue. FOCUS ranked first across all 11 diagnostic and 12 biomarker-prediction tasks in the internal test cohort, achieving mean AUCs of 0.969 and 0.877, respectively. This advantage was retained in the temporal and external test cohorts. We further organized a WHO-aligned diagnostic hierarchy, where FOCUS also outperformed comparator models. In a reader study, FOCUS assistance enabled junior readers to approach or exceed unassisted senior readers in hierarchical diagnostic accuracy and granularity. More importantly, FOCUS assistance enabled readers to correctly identify a mean of 63.0\% of initially missed lymphoma cases (73.2\% among junior readers), potentially avoiding unnecessary extensive resection. These findings highlight the potential of FOCUS to provide clinically actionable support for intraoperative CNS pathology and inform surgical decisions.

\end{abstract}
\begin{document}
\flushbottom
\maketitle

\thispagestyle{empty}

\section{Introduction}
Intraoperative neuropathological assessment is central to the surgical management of central nervous system (CNS) lesions~\cite{gal2005100, taxy2009frozen, powell2005intraoperative}. During neurosurgery, frozen-section diagnosis can distinguish neoplastic from non-neoplastic lesions and establish a broad diagnostic category. These findings can confirm that the surgical target is appropriate and help differentiate lesions for which cytoreductive resection is generally pursued from those in which extensive resection should be limited or reconsidered, such as lymphoma~\cite{somerset2011approach, lee2015basics}. This role is particularly important in CNS surgery because the benefit of resection must be balanced against the risk of permanent neurological injury~\cite{weller2021eano, brown2016association, molinaro2020association, karschnia2024oncological}.

Despite its clinical importance, intraoperative neuropathological diagnosis remains challenging. Ice-crystal artefacts, staining variability and poor nuclear detail introduced in frozen-section processing can obscure CNS-specific morphological features~\cite{powell2005intraoperative, lee2015basics}. Interpretation further requires specialized neuropathology expertise because accurate distinction often depends on subtle CNS-specific cytological and architectural characteristics~\cite{powell2005intraoperative, lee2015basics, plesec2007frozen, plesec2009frozen}. These complex assessments must be completed within a short intraoperative window~\cite{wagner2024confocal}. Diagnostic reliability consequently remains imperfect, with frozen-section discrepancies reported across diverse CNS lesions~\cite{plesec2007frozen, plesec2009frozen, cakir2019intraoperative}. Incorrect or uncertain diagnoses may lead to inappropriate resection, increase neurological risk, require additional sampling or prolong surgery~\cite{vermeulen2023ultrafast, wagner2024confocal}.

Beyond morphology, the 2021 WHO classification groups CNS tumours into families, types and subtypes, with molecular features defining many entities~\cite{louis20212021}. However, molecular findings required for integrated diagnoses are generally unavailable during frozen-section consultation. Rapid nanopore-based methylation profiling may yield intraoperative classifications but requires dedicated sample processing and sequencing~\cite{djirackor2021intraoperative, vermeulen2023ultrafast}. Consequently, intraoperative diagnoses often remain at broader diagnostic levels rather than specific molecularly defined entities.

Computational pathology offers a route to extract diagnostically and biologically informative signals from routine histology~\cite{shaktah2025application,sun2023tumor,li2024point}. Existing task-specific models support diagnosis and molecular prediction from frozen sections~\cite{zhang2024multicenter,yuan2025cryoaid, jin2026hetairos,han2026molecular}, but are constrained by limited dataset sizes. Frozen-to-FFPE translation can reduce visible artefacts but may generate artificial morphology~\cite{ozyoruk2022deep,ho2025f2fldm}. Pathology foundation models such as H-Optimus-1~\cite{scalbert2026hoptimus1} learn robust representations through large-scale self-supervised learning, enabling transfer to downstream tasks with limited supervision~\cite{ma2025pathbench}. However, applying these models to intraoperative CNS frozen sections introduces two distinct challenges. First, most pathology foundation models are pretrained mainly on formalin-fixed paraffin-embedded (FFPE) slides, which differ from frozen sections in artefacts and cellular detail~\cite{ozyoruk2022deep,gorman2023artificial}. Recent advances, notably CRISP~\cite{zhao2026clinically}, have addressed this gap through frozen-section pretraining, demonstrating broad applicability across intraoperative diagnostic tasks. Nevertheless, CNS diagnosis often depends on CNS-specific features, which are not specifically enriched during pan-organ pretraining~\cite{powell2005intraoperative,lee2015basics,louis20212021,weller2021eano}. Intraoperative neuropathology therefore calls for representations adapted to both frozen-section morphology and CNS-specific diagnostic patterns.

Here, we introduce \textbf{FOCUS}, a \textbf{F}oundation model for intra\textbf{O}perative frozen-section pathology of the \textbf{C}entral nervo\textbf{US} system (CNS). Built on H-Optimus-1, FOCUS was developed through parameter-efficient self-supervised adaptation on approximately 11 million patches from 27,930 frozen-section whole-slide images, nearly half from CNS tissue. We evaluated FOCUS against eight pathology foundation models on 11 diagnostic tasks, a WHO-aligned four-level diagnostic hierarchy and 12 biomarker-prediction tasks across internal, temporal and external validation cohorts. FOCUS ranked first on all internal validations, while maintaining advantages in temporal and external cohorts. In paired reader studies involving six neuropathologists, FOCUS assistance improved readers' hierarchical diagnostic accuracy and granularity and increased lymphoma detection, potentially helping avoid unnecessary extensive resection. Together, these evaluations support the potential of FOCUS as a frozen-section-adapted, CNS-specific foundation model for robust and clinically actionable intraoperative diagnostic support.

\section{Results}
\subsection{Study design}

We developed FOCUS as a subspecialty foundation model for intraoperative central nervous system (CNS) pathology (Fig.~\ref{fig1}{a}). The pretraining corpus comprised 27,930 whole-slide images (WSIs) from seven public datasets spanning 27 anatomical sites or tissue categories, including 13,689 brain slides (49.0\%) (Fig.~\ref{fig1}{b,c}, Supplementary Table~\cref{tabs_dist_pretrain_source,tabs_dist_pretrain_organ}). To preserve morphological diversity while reducing redundancy, hierarchical clustering-based data curation (HCDC) \cite{chen2025revisiting} was applied to UNI-derived patch embeddings \cite{chen2024towards}, yielding approximately 11 million patches for pretraining (Fig.~\ref{fig1}{d}). Starting from H-Optimus-1 \cite{scalbert2026hoptimus1}, FOCUS was adapted within a DINOv2 teacher--student framework \cite{oquab2023dinov2} using low-rank adaptation (LoRA) \cite{hu2022lora} and Frozen-Aware Multi-View (FAMV) augmentation (Fig.~\ref{fig1}{e}).

We then evaluated the fixed FOCUS encoder against eight pathology foundation models across 23 slide-level tasks, comprising 11 diagnostic and 12 biomarker-prediction tasks (Fig.~\ref{fig1}{f--h}). Ten of the 11 diagnostic tasks were organized into a WHO-aligned four-level hierarchy covering primary site, tumour lineage, tumour family and tumour subtype. Lymphoma differentiation was evaluated separately because of its importance for intraoperative surgical decision-making. We further investigated whether integrating patient age and anatomical sampling location with histopathological features could improve hierarchical diagnosis (Fig.~\ref{fig3}{f,g}). Model generalizability was assessed in independent temporal and external validation cohorts, with additional evaluations of diagnostically uncertain and discordant cases. Paired reader studies involving three junior and three senior neuropathologists assessed the clinical utility of FOCUS assistance in hierarchical diagnosis and lymphoma detection. Finally, attention mapping and morphological prototype analysis examined prediction-associated tissue patterns, while few-shot learning, pan-cancer transfer and robustness analyses assessed broader representational capabilities.

\subsection{FOCUS supports intraoperative CNS tumour diagnosis}

We first evaluated whether the subspecialty-specific representations learned by FOCUS could distinguish CNS tumour types relevant to intraoperative diagnosis. All models used the same dual-stream multiple-instance learning (DSMIL) aggregator to isolate representation quality~\cite{li2021dual}; alternative aggregators are evaluated in Supplementary Tables~\cref{tab:task_definitions,tab:focus_val_auc_ci,tab:focus_test_auc_ci}. Across the 11 tasks in the internal test set, FOCUS achieved the highest AUC or macro-AUC for every comparison and significantly surpassed other baselines ($P < 0.001$), with a mean AUC of 0.969 (95\% CI, 0.960--0.978; Fig.~\ref{fig2}{a}, Supplementary Tables~\cref{tab:dsmil_macro_metrics_task1,tab:dsmil_macro_metrics_task2,tab:dsmil_macro_metrics_task3,tab:dsmil_macro_metrics_task4,tab:dsmil_macro_metrics_task5,tab:dsmil_macro_metrics_task6,tab:dsmil_macro_metrics_task7,tab:dsmil_macro_metrics_task8,tab:dsmil_macro_metrics_task9,tab:dsmil_macro_metrics_task10,tab:dsmil_macro_metrics_task12}).

We next tested whether this diagnostic performance persisted with limited training examples and under distribution shift. When the validation and test sets were held fixed and only 1, 2, 4, 8, 16 or 32 support images per class were available, FOCUS achieved the highest mean AUC at every shot (from 0.683 to 0.907; Fig.~\ref{fig2}{b}, Extended Data Fig.~\ref{figs_diag_kshot}). We then evaluated the same representations across temporal and institutional distribution shifts. FOCUS achieved an average AUC of 0.955 (95\% CI, 0.940--0.967) in the temporal validation cohort and 0.915 (95\% CI, 0.868--0.953) in the external validation cohort (Fig.~\ref{fig2}{c,d}, Supplementary Tables~\cref{tab:external1_diagnosis_auc_acc,tab:external2_diagnosis_auc_acc}). Thus, the diagnostic advantage extended from the internal test set to low-data settings and temporally and institutionally distinct cohorts.

Accurate intraoperative identification of CNS lymphoma is critical for guiding surgical decisions. We therefore evaluated lymphoma detection separately in an independent lymphoma differentiation cohort. FOCUS achieved the highest AUC and ACC among the evaluated models, with values of 0.967 (95\% CI, 0.936--0.998) and 0.902 (95\% CI, 0.829--0.946), respectively (Fig.~\ref{fig2}{e,f}, Supplementary Table~\cref{tab:lymphoma_differentiation_metrics}).  To assess discrimination at a clinically relevant high-sensitivity operating point, we also assessed specificity at 95\% sensitivity for CNS tumour and lymphoma detection. For CNS tumour detection, FOCUS achieved the highest specificity across all three cohorts (Fig.~\ref{fig2}{g}, Supplementary Table~\cref{tab:task1_three_cohorts_task12_pooled}). Given the limited number of lymphoma cases in individual cohorts, we pooled all four cohorts for lymphoma detection and FOCUS achieved the highest specificity of 0.813 (95\% CI, 0.689--0.957; Fig.~\ref{fig2}{h}, Supplementary Table~\cref{tab:task1_three_cohorts_task12_pooled}).

To examine morphological correlates of model predictions, we visualized attention maps on representative frozen sections. FOCUS highlighted hypercellular regions with nuclear atypia and hyperchromasia in tumour detection (Fig.~\ref{fig2}{i}) and dense infiltrates of atypical lymphoid-appearing cells in lymphoma differentiation (Fig.~\ref{fig2}{j}), while assigning lower attention to sparsely cellular or vacuolated regions. Additional examples are shown in Supplementary Fig.~\cref{fig:histology_attention_examples,fig:molecular_attention_examples}. Together, these results demonstrate strong diagnostic discrimination of FOCUS across multiple intraoperative tasks.

\subsection{FOCUS preserves performance across the diagnostic hierarchy}

To assess FOCUS's ability to refine intraoperative diagnoses from broad categories to specific tumour entities, we organized ten of the 11 diagnostic tasks into a four-level hierarchy aligned with the WHO classification. In the internal test set, FOCUS achieved the highest accuracy of 0.702 at the deepest levels and the largest mean maximum correct depth of 3.37. The same pattern was reflected in the hierarchy-aware summary metrics defined in \nameref{sec:methods}, for which FOCUS achieved the highest hierarchical F1 score (H-F1) of 0.865 (95\% CI, 0.845--0.882) and depth-weighted hierarchical F1 score (DH-F1) of 0.825 (95\% CI, 0.801--0.847; Fig.~\ref{fig3}{a}, Supplementary Tables~\cref{tab:dsmil_levelwise_h_f1,tab:dsmil_levelwise_dh_f1,tab:dsmil_levelwise_accuracy}). These findings extend the task-level results to the consistency and depth of the diagnostic pathway.

We next tested whether this advantage persisted across temporal and institutional shifts. In the temporal validation cohort ($n=309$), FOCUS retained the highest H-F1 (0.816; 95\% CI, 0.785--0.845) and DH-F1 (0.760; 95\% CI, 0.724--0.796), and its accuracy advantage over the nearest comparator increased toward level 4 from +0.017 to +0.029 (Fig.~\ref{fig3}{b}, Supplementary Tables~\cref{tab:dsmil_h_f1_result_external1_2e-5_adam_1e-5_binary,tab:dsmil_dh_f1_result_external1_2e-5_adam_1e-5_binary,tab:dsmil_accuracy_result_external1_2e-5_adam_1e-5_binary}). Similar results were observed in the external cohort (Fig.~\ref{fig3}{c}, Supplementary Tables~\cref{tab:dsmil_h_f1_result_external2_2e-5_adam_1e-5_binary,tab:dsmil_dh_f1_result_external2_2e-5_adam_1e-5_binary,tab:dsmil_accuracy_result_external2_2e-5_adam_1e-5_binary}), further supporting FOCUS's stable performance under temporal and institutional shifts.

The value of retaining the correct diagnostic pathway becomes particularly apparent when cases are uncertain or diagnostically discordant. Among uncertain cases ($n=77$), FOCUS achieved the highest H-F1 of 0.813 (95\% CI, 0.753--0.873) and DH-F1 of 0.761 (95\% CI, 0.689--0.831; Fig.~\ref{fig3}{e}, Supplementary Tables~\cref{tab:dsmil_h_f1_uncertainty_group,tab:dsmil_dh_f1_uncertainty_group,tab:dsmil_accuracy_uncertainty_group}). The distinction was more apparent in diagnostically discordant cases ($n=66$). Although performance decreased for all models, FOCUS retained the highest H-F1 of 0.637 (95\% CI, 0.583--0.691) and DH-F1 of 0.503 (95\% CI, 0.441--0.568; Fig.~\ref{fig3}{d}, Supplementary Tables~\cref{tab:dsmil_h_f1_discordance_group,tab:dsmil_dh_f1_discordance_group,tab:dsmil_accuracy_discordance_group}). In these challenging subsets, FOCUS more often retained the correct trajectory when fine-grained categorization was unresolved.

Finally, we examined whether age and tumour location provided information complementary to frozen-section morphology, as neuropathologists routinely consider these factors. Adding these variables improved macro-AUC in seven of the ten component tasks (Fig.~\ref{fig3}{f}). At the hierarchical level, clinical integration increased H-F1 by 2.36 percentage points and DH-F1 by 2.91 points (Fig.~\ref{fig3}{g}, Supplementary Tables~\cref{tab:clinical_information_task_delta,tab:clinical_information_metric_delta}). These findings indicate that clinical context contributed to fine-grained distinctions.

\subsection{FOCUS links pathological morphology to molecular alterations}

Beyond morphological classification, intraoperative frozen sections may contain information associated with molecular alterations that define CNS tumour entities. We therefore examined whether FOCUS representations could support biomarker prediction directly from histology. In the internal cohort, FOCUS achieved the highest AUC in each of 12 biomarker-prediction tasks, with an average AUC of 0.877 (95\% CI, 0.830--0.924), significantly outperforming other baselines ($P$ values ranging from $0.0017$ to $0.0024$; Fig.~\ref{fig5}{a}, Supplementary Tables~\cref{tab:dsmil_macro_metrics_task17,tab:dsmil_macro_metrics_task18,tab:dsmil_macro_metrics_task19,tab:dsmil_macro_metrics_task20,tab:dsmil_macro_metrics_task21,tab:dsmil_macro_metrics_task22,tab:dsmil_macro_metrics_task24,tab:dsmil_macro_metrics_task25,tab:dsmil_macro_metrics_task26,tab:dsmil_macro_metrics_task28,tab:dsmil_macro_metrics_task31,tab:dsmil_macro_metrics_task32}). The molecular representation also remained useful when supervision was restricted: across 1--32 shots per class, FOCUS rose from a mean AUC of 0.552 at one shot to 0.737 at 32 shots and had the highest mean AUC from two shots onward (Fig.~\ref{fig5}{b}, Extended Data Fig.~\ref{figs_biom_kshot}).

We then evaluated whether this task-level advantage persisted across temporal and institutional shifts. Averaged across the 12 molecular tasks, FOCUS achieved an AUC of 0.866 (95\% CI, 0.823--0.897) in the temporal validation cohort, compared with 0.832 (95\% CI, 0.780--0.874) for the nearest model, GigaPath (Fig.~\ref{fig5}{c}, Supplementary Table~\cref{tab:external1_biomarker_auc_acc}). In the independent external cohort, where 10 tasks contained sufficient class support, FOCUS retained the highest mean AUC at 0.902 (95\% CI, 0.819--0.971), followed by H-Optimus-0 at 0.871 (95\% CI, 0.807--0.923; Fig.~\ref{fig5}{d}, Supplementary Table~\cref{tab:external2_biomarker_auc_acc}).

To characterize morphological correlates of molecular predictions, we grouped high-contribution patches into ten shared prototypes (Fig.~\ref{fig5}{e,f}, Supplementary Table~\cref{tab:prototype-morphology}). The strongest associations involved IDH and 1p/19q (Fig.~\ref{fig5}{g}, Extended Data Fig.~\ref{fig:prototype_enrichment}). IDH-wild-type tumours favoured atypical, proliferative P05, whereas IDH-mutant tumours favoured less cellular, perineuronal P08 ($-0.258$ and $+0.203$; both $P<0.001$). Within IDH-mutant tumours, 1p/19q codeletion shifted enrichment from pleomorphic P07 to oligodendroglial-like P02, characterized by monomorphic cells and branching capillaries ($-0.185$ and $+0.168$; both $P<0.001$). TERT mutation in IDH-wild-type tumours modestly enriched vascular P02 ($+0.068$; $P=0.026$), whereas CDKN2A/B deletion enriched vascular P06 and proliferative P05 ($+0.191$ and $+0.124$; both $P<0.001$). These findings link FOCUS representations to genotype-associated glioma morphology.

\subsection{FOCUS aids pathologists in diagnosis and lymphoma detection}

We next moved beyond standalone model evaluation to assess whether FOCUS could improve intraoperative diagnostic performance through human–AI collaboration. Three junior (mean experience, 6 years) and three senior (mean experience, 15 years) neuropathologists interpreted randomly selected, diagnostically uncertain and discordant cases with and without FOCUS assistance. As a standalone reader, FOCUS exceeded the unaided junior- and senior-reader means in both H-F1 and DH-F1 across all three cohorts (Fig.~\ref{fig4}{a,b}). This performance translated into gains with assistance: in the randomly selected and uncertainty cohorts, assisted junior-reader performance approached or exceeded the unaided senior-reader level. In the randomly selected cohort, for example, mean H-F1 increased from 0.696 to 0.786 for junior readers and from 0.812 to 0.847 for senior readers, with parallel gains in DH-F1. Improvements in both hierarchical metrics were significant for five of six and four of six readers in the two cohorts, respectively (Supplementary Tables~\cref{tab:reader-study-performance-g1,tab:reader-study-performance-g2}). The discordance cohort showed a different experience-related pattern: unaided junior readers outperformed unaided senior readers in both H-F1 and DH-F1. FOCUS assistance nevertheless increased mean H-F1 in senior readers from 0.497 to 0.542 (Fig.~\ref{fig4}{a,b}, Supplementary Table~\ref{tab:reader-study-performance-g3}).

To determine whether these gains reflected more specific correct diagnoses, we next examined diagnostic depth. As a standalone reader, FOCUS reached mean maximum correct depths of 2.94, 2.69, and 2.14 in the three cohorts, respectively, exceeding the mean of the unaided readers in paired permutation tests in all three cohorts ($P < 0.001$; Fig.~\ref{fig4}{c}). After assistance, five readers had positive mean depth changes and one was unchanged in the randomly selected cohort, while all six readers had positive mean changes in the uncertainty and discordance cohorts (Fig.~\ref{fig4}{d}, Supplementary Tables~\cref{tab:reader-study-depth-g1,tab:reader-study-depth-g2,tab:reader-study-depth-g3}).

Layer-wise accuracy localized these gains within the diagnostic hierarchy. We focused on level 3, which provides clinically informative tumour categories without requiring precise subtyping. At this level, standalone FOCUS exceeded both unaided reader-group means across all three cohorts; in the randomly selected cohort, its accuracy was 0.860, compared with 0.719 for unaided senior readers. AI assistance primarily improved accuracy at deeper levels, with little change at the first level (Fig.~\ref{fig4}{e}, Supplementary Tables~\cref{tab:reader-study-level-accuracy-g1,tab:reader-study-level-accuracy-g2,tab:reader-study-level-accuracy-g3}). At level 3, junior-reader accuracy in the randomly selected cohort increased from 0.532 to 0.702, approaching the unaided senior-reader mean of 0.719, with similar gains observed in the uncertainty and discordance cohorts. Across 1,458 paired reader--case assessments, AI assistance increased the maximum correct depth in 222 assessments, and rectified 51 incorrect diagnoses (Fig.~\ref{fig4}{f}). Alongside these gains, FOCUS assistance also improved inter-reader agreement among the six readers, with Fleiss' $\kappa$ increasing from 0.225 to 0.299 (Supplementary Table~\cref{tab:reader-agreement}).

We next evaluated intraoperative lymphoma detection, which may help avoid unnecessary extensive resection. In a separate cohort of 102 cases, including 81 lymphomas and 21 non-lymphomas, standalone FOCUS exceeded the unaided junior- and senior-reader means in both accuracy and lymphoma recall, achieving 0.902 and 0.963, respectively (Fig.~\ref{fig4}{g,h}, Supplementary Table~\cref{tab:reader-study-lymphoma}). FOCUS assistance increased mean lymphoma recall from 0.816 to 0.947 among junior readers ($P < 0.001$) and from 0.840 to 0.914 among senior readers ($P = 0.002$; Fig.~\ref{fig4}{h}, Supplementary Table~\cref{tab:reader-study-lymphoma}). Junior readers showed the larger gain, narrowing the accuracy gap with senior readers from 8.4 to 3.9 percentage points (Fig.~\ref{fig4}{g}, Supplementary Table~\cref{tab:reader-study-lymphoma}). These recall gains included correction of lymphoma cases initially classified as non-lymphoma, with mean reader-level correction proportions of 73.2\% among junior readers, 52.7\% among senior readers and 63.0\% across all readers. In two representative cases of diffuse large B-cell lymphoma, preoperative imaging and intraoperative frozen-section interpretation favoured non-lymphoma diagnoses, whereas FOCUS correctly identified lymphoma (Extended Data Fig.~\cref{fig:case_study}{a,b}).

FOCUS also provided a faster post-preparation route to a diagnostic reference. Conventional slide review generally required 5--10 min, with a further 5--10 min for difficult cases requiring senior consultation. The post-preparation AI pathway comprised slide scanning, which required approximately 1--1.5 min, followed by automated analysis. Combining the scanning interval with the measured device-specific processing times  (1.6--2.3 min, Supplementary Table~\cref{tab:focus-processing-time}) showed that the AI pathway remained shorter than routine slide review across all tested hardware configurations, with a larger time advantage for cases requiring senior review. Together, these findings support FOCUS as a timely intraoperative reference that improves comprehensive diagnosis and especially lymphoma recognition.

\subsection{Transferability and robustness of FOCUS representations}

We next examined whether subspecialty pretraining preserved transferable morphology and whether its performance depended on specific components or image scale. Because the pan-cancer evaluation test sets overlapped with CRISP’s pretraining dataset, we excluded CRISP to avoid potential data leakage. Across nine pan-cancer tasks with external validation, FOCUS achieved a higher area under the receiver operating characteristic curve (AUC) in 49 of 63 model--task comparisons, with a mean paired difference of $+0.0098$ (Fig.~\ref{fig6}{a,b}, Supplementary Tables~\cref{tab:dsmil_metrics_blca_benign_malignant,tab:dsmil_metrics_brca_benign_malignant,tab:dsmil_metrics_cesc_tumor_grading,tab:dsmil_metrics_coadread_benign_malignant,tab:dsmil_metrics_hnsc_benign_malignant,tab:dsmil_metrics_kirc_kirp_kich_benign_malignant,tab:dsmil_metrics_kirc_kirp_kich_histological_subtype,tab:dsmil_metrics_kirc_kirp_kich_tumor_grading,tab:dsmil_metrics_lihc_benign_malignant,tab:dsmil_metrics_lihc_tumor_grading,tab:dsmil_metrics_luad_lusc_histological_subtype,tab:dsmil_metrics_lung_benign_malignant,tab:dsmil_metrics_ov_benign_malignant,tab:dsmil_metrics_paad_benign_malignant,tab:dsmil_metrics_prad_benign_malignant,tab:dsmil_metrics_prad_gleason_grade_group,tab:dsmil_metrics_stad_benign_malignant,tab:dsmil_metrics_thca_benign_malignant,tab:dsmil_metrics_ucec_ucs_benign_malignant}). Across ten tasks evaluated only by internal validation, 27 comparisons favoured FOCUS, 36 were tied, and 7 favoured a comparator (Fig.~\ref{fig6}{c}). Despite no significant differences ($P>0.05$), comparable performance suggests that FOCUS retained transferable morphological representations beyond its training domain.

To assess whether FOCUS performance depended on a specific MIL aggregation method, we first benchmarked four representative MIL methods—ABMIL~\cite{ilse2018attention}, CLAM~\cite{lu2021data}, TransMIL~\cite{shao2021transmil} and DSMIL~\cite{li2021dual}—across all 23 downstream tasks. MIL methods did not differ significantly in validation or test performance ($P > 0.05$; Supplementary Table~\cref{tab:focus_val_auc_ci,tab:focus_test_auc_ci}), indicating that the performance of FOCUS was not dependent on a particular aggregator. DSMIL was selected as default aggregator based on validation performance. We next directly tested the learned representations independently of aggregator using patch-level linear probing on six histological regions (3,000 patches each) from 20 pretraining-excluded Ivy-GAP WSIs. FOCUS ranked first among all competitors and achieved an accuracy of 0.730 $\pm$ 0.010 across five folds (Fig.~\ref{fig6}{d}, Supplementary Table~\ref{tab:patchcls-fold-results}). FOCUS also showed stable mean performance across the tested input dimensions (224$\times$224, 448$\times$448, 896$\times$896). Relative to 224 pixels, the mean paired AUC difference was $+0.003$ at 448 pixels and $-0.002$ at 896 pixels ($P >0.05$; Fig.~\ref{fig6}{e}, Supplementary Table~\ref{tab:multisize-stability}).

Finally, paired ablations evaluated the contribution of the proposed strategy (Fig.~\ref{fig6}{f}, Supplementary Table~\ref{tab:ablation-results}). The hierarchical clustering-based data curation (HCDC) comparison favoured the full pipeline in 22 of 23 matched tasks, with a mean paired gain of $+0.0125$ ($P < 0.001$). The Frozen-Aware Multi-View augmentation (FAMV) comparison produced a smaller increment: 17 of 23 tasks improved, with a mean gain of $+0.0012$ ( $P = 0.033$). HCDC therefore showed the consistent association with performance across tasks, whereas FAMV provided a modest refinement.

\section*{Discussion}

Intraoperative CNS pathology guides time-sensitive surgical decisions, yet frozen-section artefacts, limited sampling and uneven access to subspecialty expertise complicate diagnosis~\cite{powell2005intraoperative,plesec2007frozen,wilson2018pathologyaccess}. Diagnostic errors may lead to inappropriate resection, while uncertainty can prolong surgery; molecular information is also generally unavailable intraoperatively~\cite{somerset2011approach,louis20212021}. Existing AI approaches have shown promise in histopathological diagnosis~\cite{zhang2024multicenter,yuan2025cryoaid,zhao2026clinically}, but task-specific models have limited diagnostic scope, transformation methods may not reliably preserve diagnostic information, and most foundation models lack specialization for intraoperative CNS pathology~\cite{chen2024towards,vorontsov2024foundation,xu2024whole}. Here, we developed FOCUS, a subspecialty foundation model pretrained on 27,930 CNS-enriched frozen-section WSIs. We evaluated FOCUS across 11 diagnostic and 12 molecular prediction tasks, with internal, temporal and external validation. To assess its clinical utility, we further examined WHO-aligned hierarchical diagnosis and lymphoma detection, and conducted paired reader studies involving six neuropathologists to evaluate AI-assisted interpretation.

The findings highlight the clinical value of hierarchical diagnosis, which provides meaningful diagnostic information even when definitive subtype classification is uncertain~\cite{louis20212021}. FOCUS preserved more accurate diagnostic paths at deeper hierarchical levels, with advantages maintained in uncertain and discordant cases. In the reader study, assistance enabled junior pathologists to approach or exceed the hierarchical performance of unaided senior pathologists. These findings suggest a potential role for FOCUS as a virtual senior-level second opinion, providing traceable diagnostic predictions and localized tissue evidence for review by local pathologists. Such assistance may be particularly valuable in regional or smaller neurosurgical centres lacking dedicated neuropathologists. Integrated into tele-neuropathology workflows, AI-assisted interpretation of digitized frozen sections could complement remote specialist consultation and extend timely subspecialty support beyond tertiary referral centres~\cite{baskota2020teleneuropathology}. This approach offers a potential pathway towards reducing geographical disparities in access to high-quality intraoperative neuropathological assessment~\cite{wilson2018pathologyaccess,fleming2021diagnosticscommission}.

A particularly consequential application of this diagnostic support is the intraoperative detection of CNS lymphoma, for which extensive surgical resection is generally not indicated~\cite{hoangxuan2023eano_pcnsl,ferreri2024eha_esmo_pcnsl}. Timely recognition can limit unnecessary resection and guide tissue allocation for lymphoma-specific investigations. In the dedicated lymphoma detection cohort, FOCUS assistance enabled junior readers to correct a mean of 73.2\% of initially misclassified lymphoma cases, with an overall mean correction rate of 63.0\% across all readers. By improving lymphoma recognition at this critical decision point, FOCUS could support more informed surgical management and facilitate timely downstream diagnostic pathways.

Beyond morphological diagnosis, FOCUS offers provisional molecular information generally unavailable during surgery. Its strong performance across 12 biomarker-prediction tasks, together with morphology-associated molecular prototypes, suggests that frozen sections contain informative genotype--phenotype signals. Early signals for alterations such as IDH mutation or 1p/19q codeletion could refine the intraoperative differential, prioritize molecular diagnoses and guide tissue allocation for confirmatory testing~\cite{louis20212021,weller2021eano}. Integrating these molecular insights with hierarchical morphological diagnosis could provide a more comprehensive characterization of CNS lesions intraoperatively, extending frozen-section assessment beyond conventional morphological interpretation.

Several limitations remain. First, the retrospective reader study did not assess surgical decisions or patient outcomes, and AI assistance occasionally introduced errors, including false-positive lymphoma predictions that could lead to inappropriate conservative management of non-lymphoma tumours (Extended Data Fig.~\cref{fig:case_study}{c}). Second, the hierarchy excluded rare CNS entities and open-set presentations. Finally, real-world deployment, particularly in resource-limited centres, will depend on rapid benchtop scanners, GPU computing resources, and standardized slide preparation and quality control, including where personnel lack specialized histotechnology training~\cite{baskota2020teleneuropathology,kaushal2023remote}. Prospective multicentre studies are therefore needed to assess diagnostic safety, end-to-end workflow feasibility and clinical outcomes across diverse healthcare settings. Overall, FOCUS provides a framework for combining hierarchical diagnosis and provisional molecular characterization intraoperatively, with the potential to improve the consistency and equitable accessibility of subspecialty neuropathological support across healthcare systems.

\clearpage

\begin{figure}
	\centering
	\includegraphics[width=.985\linewidth]{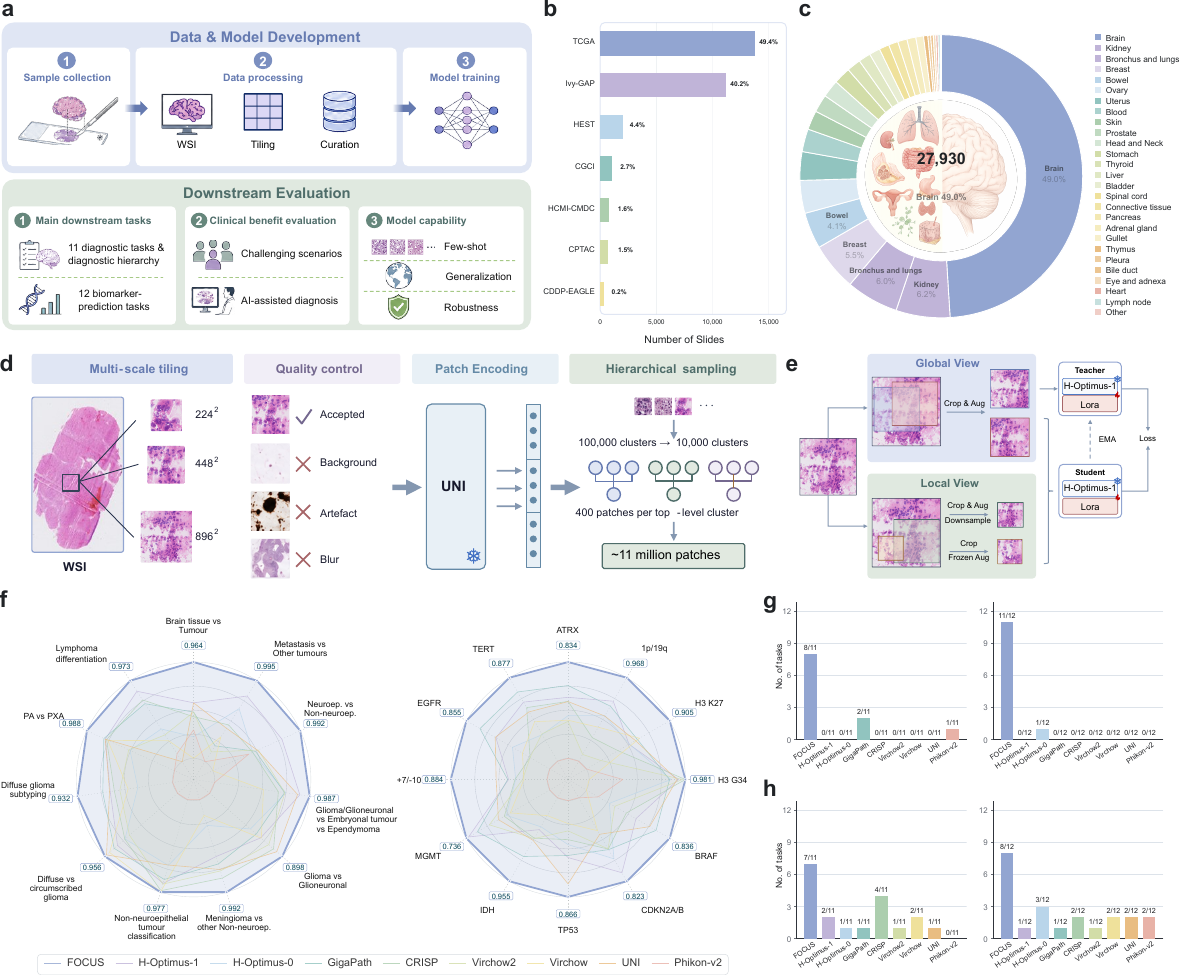}
    \refstepcounter{figure}
	\label{fig1}
\end{figure}
\noindent\textbf{Figure 1. Overview of FOCUS.} \textbf{a,} Study workflow from data collection and model training to diagnostic, molecular and clinical evaluation. \textbf{b,} Distribution of 27,930 pretraining whole-slide images (WSIs) across seven public datasets. \textbf{c,} Distribution across 27 anatomical categories; brain specimens accounted for 49.0\% of the pretraining WSIs. \textbf{d,} Multi-scale patch curation, UNI encoding and hierarchical sampling from 100,000 fine-grained and 10,000 top-level clusters, yielding approximately 11 million patches. \textbf{e,} Teacher--student adaptation using low-rank adaptation (LoRA) and Frozen-Aware Multi-View (FAMV) augmentation. \textbf{f,} AUC across 11 diagnostic tasks (left) and 12 biomarker-prediction tasks (right) for FOCUS and eight pathology foundation models in the internal test cohort; boxed values indicate FOCUS performance. \textbf{g,} Number of diagnostic and biomarker-prediction tasks on which each model ranked first in the independent temporal validation cohort. \textbf{h,} Number of diagnostic and biomarker-prediction tasks in which each model ranked first in the external validation cohort. For tasks in which AUC could not be calculated because one class was absent, accuracy was used instead.

\clearpage

\begin{figure}[htp]
	\centering
	\includegraphics[width=.985\linewidth]{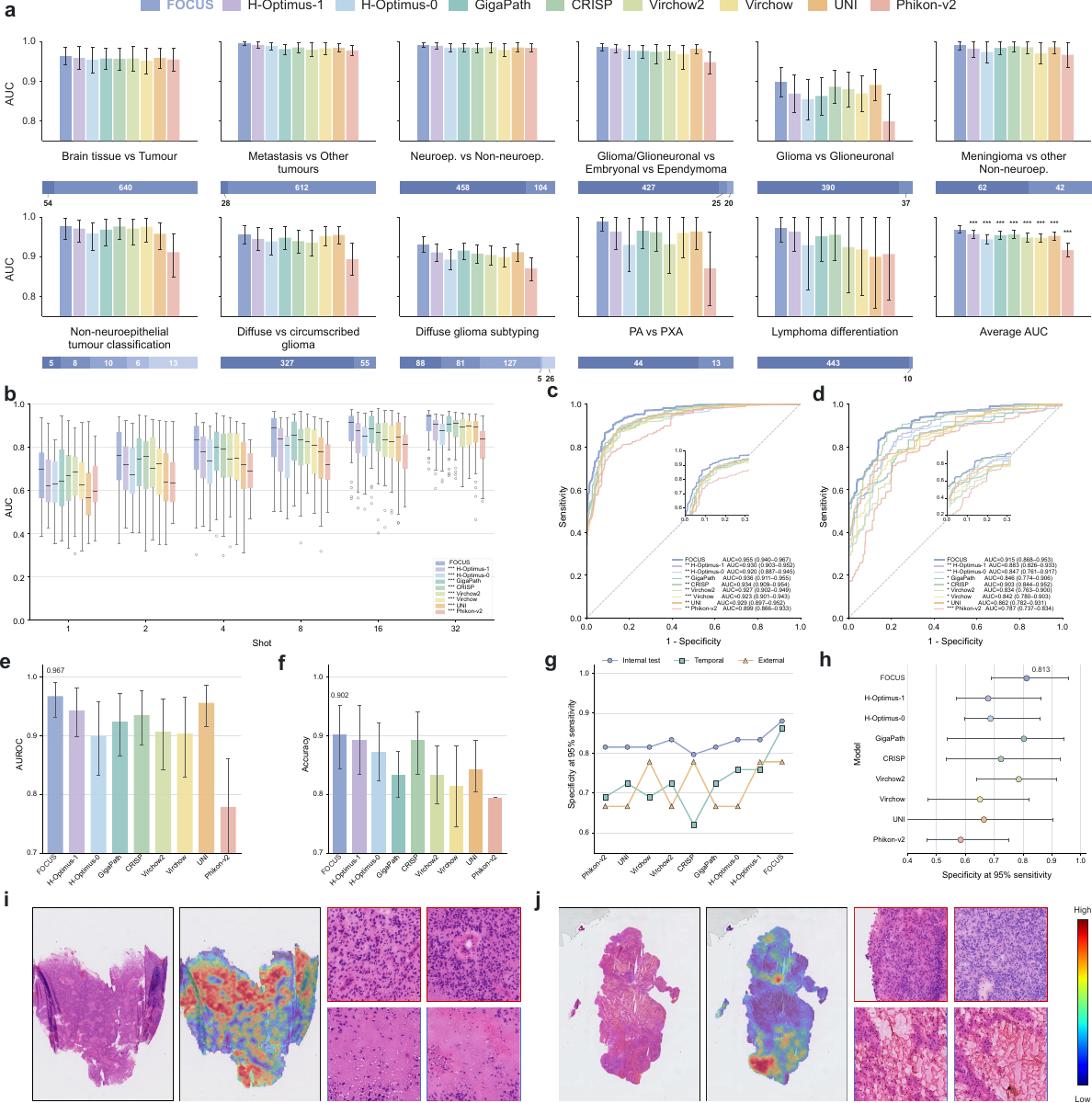}
    \refstepcounter{figure}
	\label{fig2}
\end{figure}
\noindent\textbf{Figure 2. Diagnostic performance of FOCUS for intraoperative CNS pathology.} \textbf{a,} AUC for binary tasks or macro-AUC for multiclass tasks across 11 diagnostic tasks in the internal cohort; error bars denote 95\% confidence intervals and horizontal bars show class counts. Statistical comparisons used two-sided paired Wilcoxon signed-rank tests. \textbf{b,} Few-shot AUCs using 1--32 examples per class across five repeats. Boxes show interquartile ranges, centre lines show medians, and whiskers extend to 1.5 times the interquartile range. \textbf{c,d,} Mean ROC curves across the 11 tasks in the temporal and external cohorts, respectively. Comparisons between FOCUS and each baseline used two-sided paired Wilcoxon signed-rank tests across the 11 tasks. \textbf{e,} AUC for lymphoma differentiation in the dedicated lymphoma differentiation cohort. \textbf{f,} Accuracy for lymphoma differentiation in the dedicated lymphoma differentiation cohort. \textbf{g,} Specificity at 95\% sensitivity for tumour detection across the internal test, temporal validation and external validation cohorts. \textbf{h,} Specificity at 95\% sensitivity after pooling slide-level predictions from the internal test, temporal validation, external validation and lymphoma differentiation cohorts. \textbf{i,j,} Attention maps and representative high- and low-attention regions for brain tissue versus tumour and lymphoma differentiation. Red and blue outlines indicate high- and low-attention regions, respectively. All statistical tests were corrected for multiple comparisons using the Benjamini--Hochberg procedure. *Adjusted $P<0.05$, **adjusted $P<0.01$ and ***adjusted $P<0.001$.

\clearpage

\begin{figure}
	\centering
	\includegraphics[width=.985\linewidth]{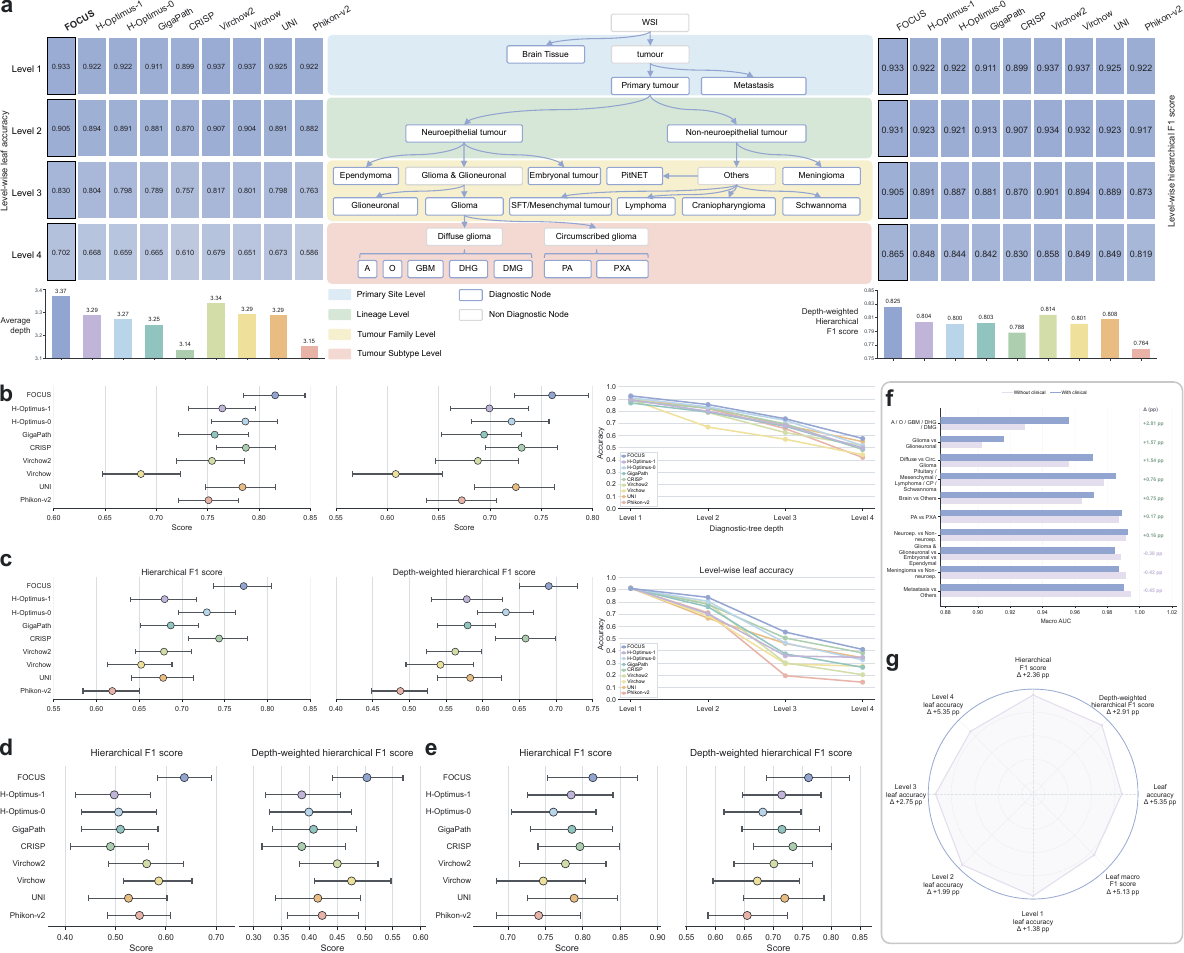}
    \refstepcounter{figure}
	\label{fig3}
\end{figure} 
\noindent\textbf{Figure 3. Hierarchical diagnostic performance of FOCUS.} \textbf{a,} Ten of the 11 diagnostic tasks were organized into a four-level central nervous system (CNS) tumour hierarchy. In the internal test set, heatmaps show cumulative level-wise leaf accuracy (ACC) and hierarchical F1 score (H-F1) after truncation at each level; lower bar plots show mean maximum correct depth and depth-weighted hierarchical F1 score (DH-F1). \textbf{b,c,} H-F1, DH-F1 and level-wise ACC in the temporal ($n=309$ slides) and independent external ($n=246$ slides) validation cohorts, respectively. \textbf{d,e,} H-F1 and DH-F1 in diagnostically uncertain ($n=77$ cases) and diagnostically discordant ($n=66$ cases) subsets, respectively. \textbf{f,} Changes in task-level macro-AUC after adding age and tumour location to morphology. \textbf{g,} Changes in hierarchical metrics and level-wise ACC after clinical-information integration. Points and error bars in \textbf{b--e} indicate point estimates and 95\% confidence intervals.

\clearpage

\begin{figure}[htp]
	\centering
	\includegraphics[width=.95\linewidth]{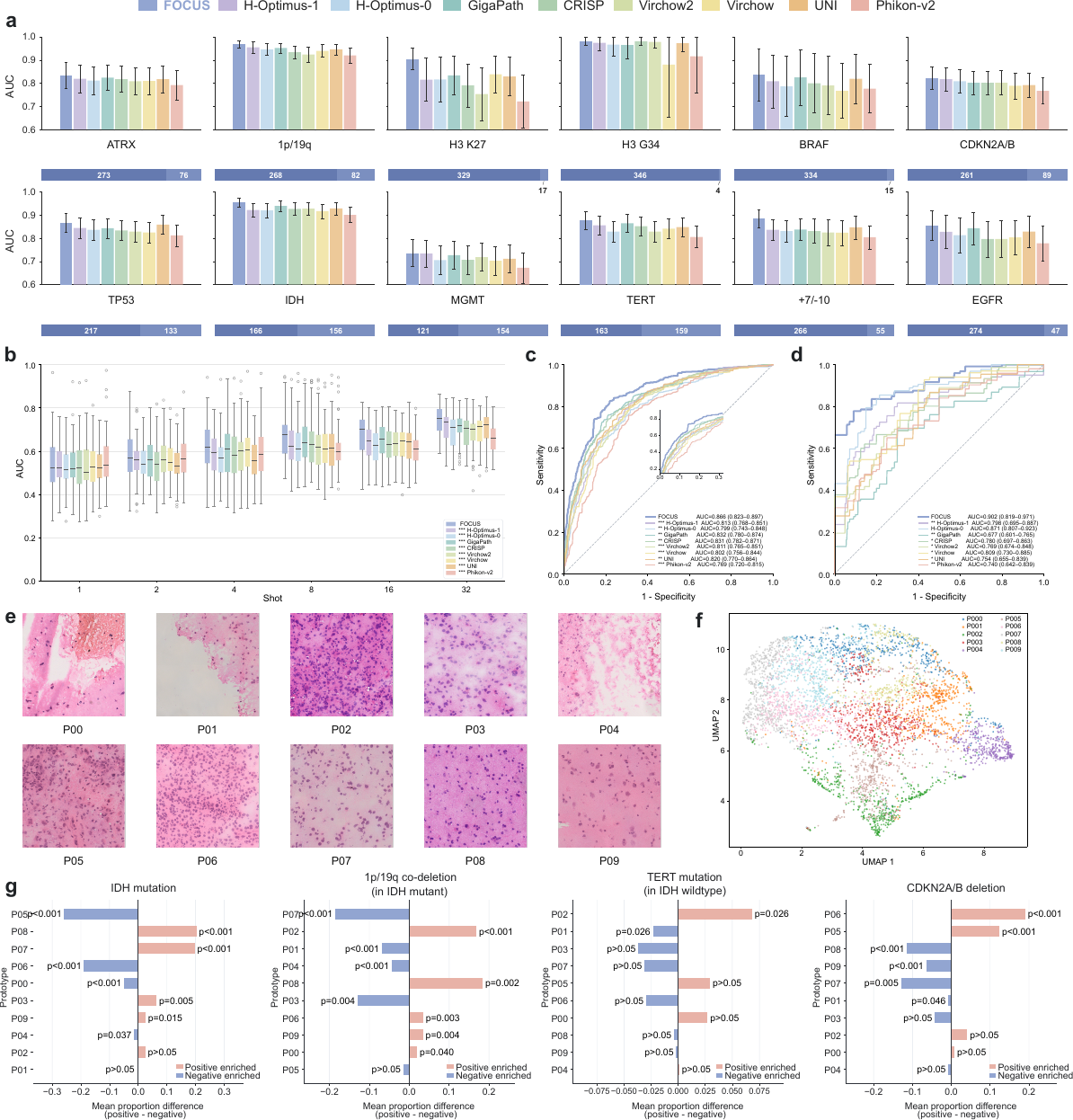}
    \refstepcounter{figure}
	\label{fig5}
\end{figure}
\noindent\textbf{Figure 4. Biomarker-prediction performance and prototype morphology of FOCUS.} \textbf{a,} AUCs for 12 biomarker-prediction tasks in the internal test cohort. Error bars denote 95\% confidence intervals; horizontal bars show class counts. Statistical comparisons used two-sided paired Wilcoxon signed-rank tests. \textbf{b,} Few-shot AUCs using 1--32 examples per class across five repeats; validation and test sets were fixed. Boxes show interquartile ranges, centre lines show medians, and whiskers extend to 1.5 times the interquartile range. \textbf{c,d,} Mean ROC curves in the temporal and independent external cohorts, respectively. Comparisons between FOCUS and each baseline used two-sided paired Wilcoxon signed-rank tests across the evaluable tasks. \textbf{e,f,} Representative patches from ten shared prototypes and their UMAP projection. \textbf{g,} Mean slide-level prototype-proportion differences for IDH mutation, 1p/19q codeletion within IDH-mutant gliomas, TERT mutation within IDH-wild-type gliomas and CDKN2A/B deletion. Positive values indicate enrichment in the molecular-positive group. Prototype proportions were compared using two-sided Mann--Whitney $U$ tests. All statistical tests were corrected for multiple comparisons using the Benjamini--Hochberg procedure. *Adjusted $P<0.05$, **adjusted $P<0.01$ and ***adjusted $P<0.001$.

\clearpage

\begin{figure}[htp]
	\centering
	\includegraphics[width=.985\linewidth]{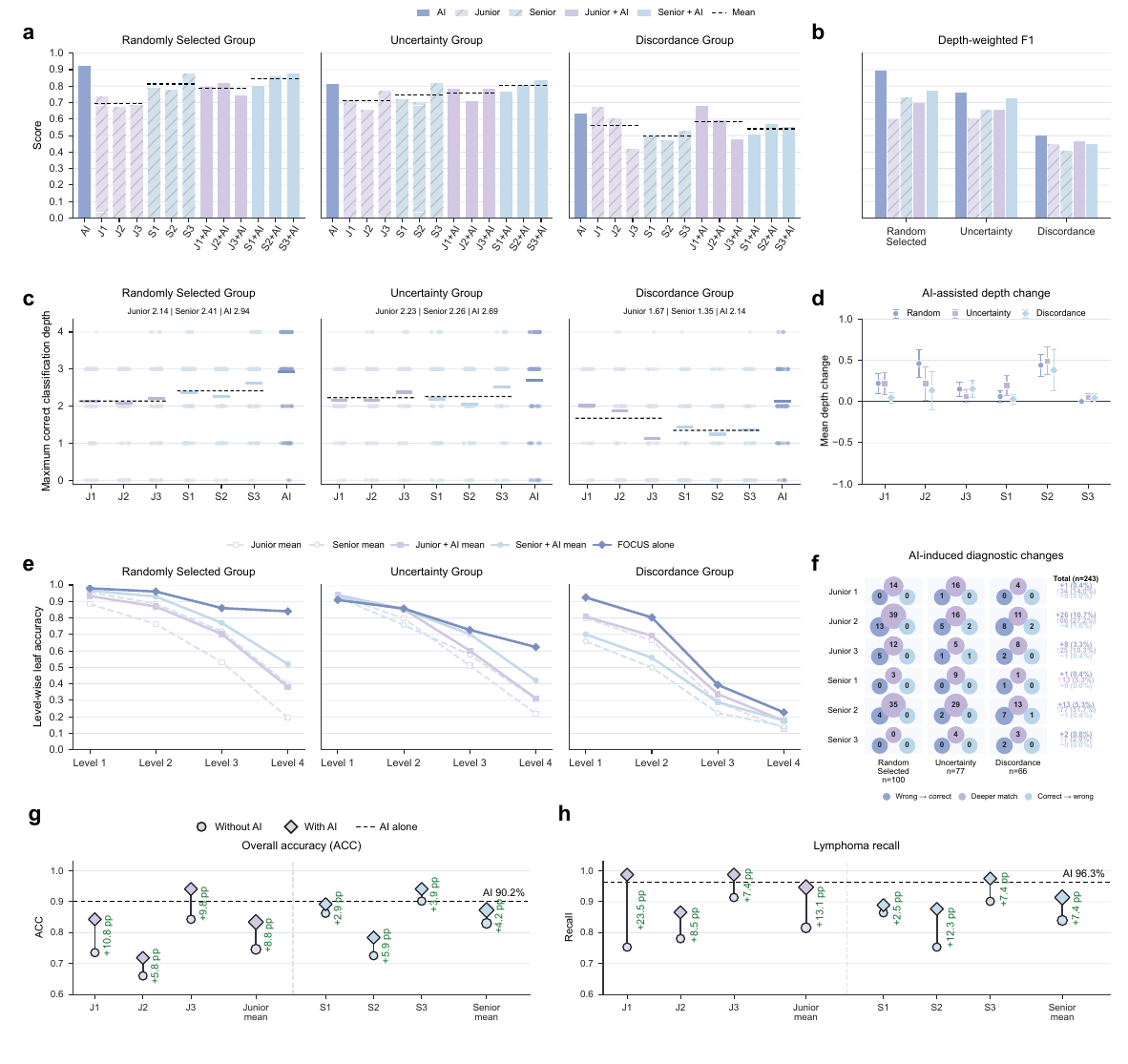}
    \refstepcounter{figure}
	\label{fig4}
\end{figure}

\noindent\textbf{Figure 5. FOCUS-assisted hierarchical diagnosis in the reader study.} Six neuropathologists (three junior and three senior) independently reviewed 243 cases and then completed an AI-assisted review after a two-week washout period. The reader-study cohort comprised 100 randomly selected cases, 77 cases whose intraoperative reports contained explicit diagnostic uncertainty and 66 cases with discordance between intraoperative and final diagnoses. \textbf{a,} H-F1 for FOCUS, unaided readers and AI-assisted readers; dashed lines denote group means. \textbf{b,} Mean DH-F1 across the three case cohorts and reader groups. \textbf{c,} Mean maximum correct classification depth for FOCUS and unaided readers. \textbf{d,} Reader-specific change in maximum correct depth after AI assistance; points and error bars show mean changes and 95\% confidence intervals. \textbf{e,} Mean level-wise leaf accuracy for unaided and AI-assisted junior and senior readers. \textbf{f,} Case-level diagnostic changes after AI assistance. Blue indicates wrong-to-correct conversions, purple indicates deeper correct classifications, and light blue indicates correct-to-wrong changes. \textbf{g,h,} Changes in reader accuracy and recall, respectively, before and after AI assistance in the lymphoma differentiation cohort. Circles and diamonds denote performance without and with AI assistance, respectively; green labels indicate absolute percentage-point changes, and dashed horizontal lines show the performance of AI alone.

\clearpage

\begin{figure}[htp]
	\centering
	\includegraphics[width=.985\linewidth]{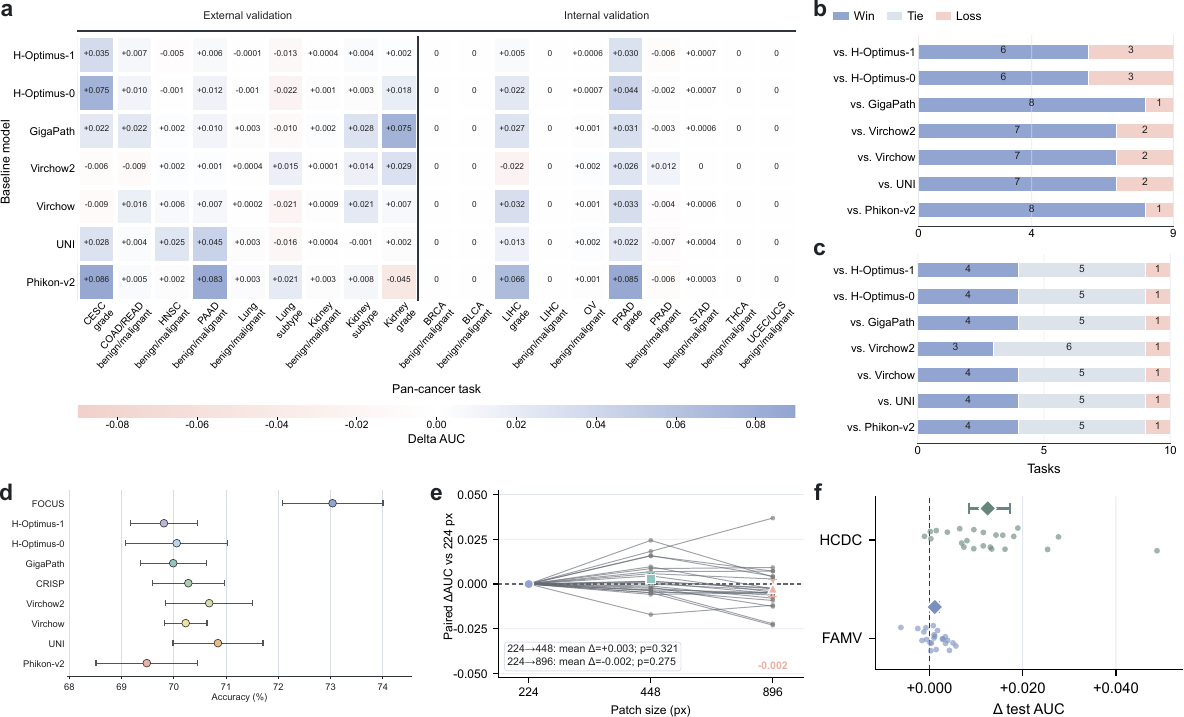}
    \refstepcounter{figure}
	\label{fig6}
\end{figure}

\noindent\textbf{Figure 6. Transferability and robustness of FOCUS representations.} \textbf{a,} AUC differences between FOCUS and seven pathology foundation models across pan-cancer tasks with external (left) or internal-only (right) validation. Differences are calculated as FOCUS minus comparator, such that positive values favour FOCUS. \textbf{b,c,} Win--tie--loss summaries for externally and internally validated tasks, respectively; ties denote $|\Delta\mathrm{AUC}|\leq10^{-6}$. \textbf{d,} Five-fold linear-probe accuracy (mean $\pm$ s.d.) for six Ivy-GAP histological regions sampled from 20 pretraining-excluded WSIs; all encoders were frozen. \textbf{e,} Paired AUC changes across 23 tasks for 448- and 896-pixel inputs relative to the 224-pixel reference; coloured markers show mean changes with 95\% task-bootstrap confidence intervals. $P$ values are from two-sided paired Wilcoxon signed-rank tests across the 23 tasks. \textbf{f,} Paired changes in test AUC attributable to hierarchical clustering-based data curation (HCDC) and FAMV, calculated against the corresponding ablation condition without each component; dots denote tasks and diamonds with bars denote means with 95\% task-bootstrap confidence intervals. P values are from two-sided paired Wilcoxon signed-rank tests across the 23 matched tasks.

\clearpage

\section{Methods}
\label{sec:methods}

\subsection{Ethics Declarations}

This study was reviewed and approved by the Institutional Review Board of Beijing Tiantan Hospital, Capital Medical University (no. KY2024-212-02), and was conducted in accordance with the Declaration of Helsinki and relevant institutional ethical guidelines. All institutional patient data and pathology images were de-identified before analysis. Given the retrospective nature of the study and the use of de-identified clinical data and archival pathology specimens collected during routine clinical care, the requirement for informed consent was waived by the institutional review board.

\subsection{Dataset Curation and Cohorts}

To develop and validate FOCUS, we constructed six types of datasets: public frozen-section slides for foundation-model pretraining, an institutional cohort for downstream task development, an independent temporal validation cohort, an external validation cohort, an independent lymphoma differentiation cohort, and public pan-cancer cohorts for evaluating cross-domain generalization.

The foundation model was pretrained on 27,930 publicly available frozen-section whole-slide images (WSIs) assembled from seven public resources. These comprised TCGA \cite{weinstein2013tcga}, Ivy-GAP \cite{puchalski2018ivy}, HEST-1k \cite{jaume2024hest}, CPTAC \cite{edwards2015cptac}, and the CGCI, HCMI-CMDC and CDDP-EAGLE projects hosted by the NCI Genomic Data Commons \cite{grossman2016gdc}. Brain and central nervous system specimens accounted for approximately 50\% of the pretraining data, while the remaining slides covered a broad range of non-CNS organs and tumour types. The contributions of individual data sources are summarized in Fig.~\ref{fig1}{b} and Supplementary Table~\ref{tabs_dist_pretrain_source}, and the organ-level distribution is shown in Fig.~\ref{fig1}{c} and Supplementary Table~\ref{tabs_dist_pretrain_organ}. All slides were used for self-supervised pretraining without task-specific diagnostic or molecular labels. To prevent data leakage, no slides from the CNS downstream validation cohorts were included in the pretraining dataset.

The CNS downstream development cohort was collected from Beijing Tiantan Hospital (BTH) between October 2022 and May 2024, and included 3,193 patients with 3,471 slides. This cohort was used to develop task-specific models for CNS intraoperative frozen-section pathology, including diagnostic tasks and biomarker prediction tasks. The following inclusion criteria were applied to all cases: (1) at least one available intraoperative frozen section; (2) the digitized slide was free of major scanning artefacts, including defocus, blurring, or other artefacts; and (3) availability of complete clinical information, intraoperative frozen-section diagnosis, and postoperative pathological diagnosis. The detailed patient enrollment workflow is summarized in Supplementary Fig.~\ref{figs_inclu_exclu}. The whole dataset was split into training and test sets at an 8:2 ratio at the patient level and all downstream tasks shared the same partition. Because slides from the same patient may contain either normal brain tissue or tumour tissue, all tasks in this study were performed at the slide level. Supplementary Table~\ref{tabs_downstream_label_dist} displays the detailed partition and label distribution of each task. For each downstream task, 20\% of the training set was further held out as a validation set for hyperparameter selection and early stopping. The test set was kept independent from model selection and was used only for final internal evaluation. Given that different slides from the same patient may contain distinct histopathological phenotypes, such as non-neoplastic brain tissue or tumour, we performed downstream tasks at the slide level.

To assess model generalizability under temporal difference and institutional distribution shifts, we established one temporal validation cohort and one external validation cohort. The temporal validation cohort comprised 309 slides from 285 patients who underwent surgery at Beijing Tiantan Hospital (BTH) between August 2025 and January 2026. These cases were collected during a time period distinct from that of the development cohort and were digitized using a different scanning platform, thereby enabling evaluation of robustness to temporal and scanner-related shifts. The external validation cohort comprised 246 slides from 223 patients who underwent surgery at the First Hospital of Hebei Medical University (FHHMU) between March 2024 and May 2026. The inclusion and exclusion process for these validation cohorts is shown in Supplementary Fig.~\ref{figs_inclu_exclu}. Both cohorts were reserved exclusively for final model evaluation and were not involved in foundation-model pretraining or downstream model training. The same task definitions, diagnostic hierarchy, molecular-label definitions, and exclusion criteria were applied across the BTH development cohort, BTH temporal validation cohort, and FHHMU external validation cohort to ensure consistent evaluation.

To assess the ability of FOCUS and pathologists to distinguish CNS lymphoma from other tumours, a distinction that can influence intraoperative surgical decisions, we established an independent lymphoma differentiation cohort at BTH using cases collected between December 2023 and August 2026. Cases were included if either preoperative imaging or intraoperative frozen-section reports left the distinction between lymphoma and other tumours unresolved. Detailed patient enrollment workflow is summarized in Supplementary Fig.~\ref{figs_inclu_exclu}. The cohort comprised 102 frozen-section WSIs from 96 patients, including 81 lymphoma and 21 non-lymphoma slides. Neither patients nor slides overlapped with those in any other study cohort. We evaluated the standalone performance of FOCUS and assessed changes in the diagnostic performance of pathologists with FOCUS assistance.

Public pan-cancer validation cohorts were used to evaluate the broader generalization ability of the foundation model beyond CNS frozen-section pathology. The data sources and task definitions are summarized in Supplementary Tables~\cref{tabs_dist_tcga_source,tabs_tcga_label_dist}. For these public cohorts, the training split may overlap with the pretraining data because both were derived from public datasets. However, the test and external test splits were kept fully independent from the pretraining dataset. Therefore, performance on these held-out splits was used to assess out-of-distribution generalization under non-CNS and pan-cancer settings.

\subsection{Downstream Task Definitions}

We evaluated 23 downstream tasks in CNS frozen-section pathology, comprising 11 diagnostic classification tasks and 12 biomarker prediction tasks. The diagnostic tasks followed the World Health Organization (WHO) classification framework~\cite{louis20212021} for CNS tumours and covered clinically relevant levels of diagnostic granularity, ranging from broad lineage-level categorisation to fine-grained tumour subtype discrimination. The biomarker prediction tasks assessed whether molecular alterations could be predicted directly from frozen-section WSIs. We constructed all tasks from a unified set of predefined training and test cohorts and, for each task, selected the relevant cases according to the corresponding diagnostic category or biomarker availability. We assigned reference labels for the diagnostic tasks using final postoperative pathological diagnoses, together with  immunohistochemical findings, molecular results and clinical history. We then harmonised these labels to the predefined CNS diagnostic hierarchy. Reference labels for biomarker prediction were obtained from clinically validated molecular pathology reports, including sequencing-based assays, methylation testing and copy-number analysis. Because different slides from the same patient may exhibit distinct histopathological phenotypes, we defined and evaluated the downstream tasks primarily at the slide level.

\noindent\textbf{Non-neoplastic brain tissue versus tumour.} This task evaluated whether the model could distinguish non-neoplastic brain tissue from tumour-containing frozen sections, representing the fundamental determination of whether sampled tissue contains tumour. The task was formulated as a binary classification problem with two classes: non-neoplastic brain tissue and tumour.

\noindent\textbf{Metastasis versus other CNS tumours.} This task assessed the ability of the model to distinguish metastasis from other CNS tumours. This distinction is clinically important because metastasis and primary CNS tumours may differ substantially in surgical strategy, ancillary testing, and subsequent oncological management. The task was formulated as a binary classification problem with two classes: metastasis and other CNS tumours.

\noindent\textbf{Neuroepithelial versus non-neuroepithelial tumour.} This task evaluated broad tumour-lineage classification in CNS frozen sections. Distinguishing neuroepithelial from non-neuroepithelial tumours provides a basis for subsequent classification into more specific tumour groups. The task was formulated as a binary classification problem with two classes: neuroepithelial tumour and non-neuroepithelial tumour.

\noindent\textbf{Glioma or glioneuronal tumour versus embryonal tumour versus ependymoma.} This task evaluated discrimination among three major neuroepithelial tumour groups encountered in CNS pathology. The task was formulated as a three-class classification problem with the following classes: glioma or glioneuronal tumour, embryonal tumour, and ependymoma.

\noindent\textbf{Glioma versus glioneuronal tumour.} This task evaluated fine-grained discrimination between gliomas and glioneuronal tumours, which may show overlapping histomorphological features in frozen sections but differ in biological characteristics and diagnostic implications. The task was formulated as a binary classification problem with two classes: glioma and glioneuronal tumour.

\noindent\textbf{Meningioma versus other non-neuroepithelial tumours.} This task assessed whether the model could distinguish meningioma from other common non-neuroepithelial CNS tumours. This distinction is clinically relevant because meningiomas may need to be differentiated intraoperatively from other non-neuroepithelial lesions with overlapping morphological features. The task was formulated as a binary classification problem with two classes: meningioma and other non-neuroepithelial tumours, with the latter comprising pituitary neuroendocrine tumour (PitNET), SFT/mesenchymal tumour, lymphoma, craniopharyngioma, and schwannoma.

\noindent\textbf{Non-neuroepithelial tumour classification.} This task evaluated fine-grained classification among major non-neuroepithelial CNS tumour categories. The task was formulated as a five-class classification problem with the following classes: pituitary neuroendocrine tumour (PitNET), SFT/mesenchymal tumour, lymphoma, craniopharyngioma, and schwannoma.

\noindent\textbf{Diffuse versus circumscribed glioma.} This task assessed whether the model could distinguish diffuse from circumscribed gliomas. These groups differ in their histopathological characteristics, molecular alterations, and subsequent integrated diagnostic work-up. The task was formulated as a binary classification problem with two classes: diffuse glioma and circumscribed glioma.

\noindent\textbf{Diffuse glioma molecular-subtype classification.} This task evaluated fine-grained classification among major molecularly defined diffuse glioma entities. The task was formulated as a five-class classification problem with the following classes: astrocytoma, IDH-mutant; oligodendroglioma, IDH-mutant and 1p/19q-codeleted; glioblastoma, IDH-wildtype; diffuse hemispheric glioma, H3 G34-mutant; and diffuse midline glioma, H3 K27-altered.

\noindent\textbf{Pilocytic astrocytoma versus pleomorphic xanthoastrocytoma.} This task evaluated discrimination between two representative circumscribed glioma entities. Their distinction can be challenging in frozen sections because of potentially overlapping morphological features, while they differ in clinicopathological context and molecular associations. The task was formulated as a binary classification problem with two classes: pilocytic astrocytoma and pleomorphic xanthoastrocytoma.

\noindent\textbf{Lymphoma differentiation.} This task evaluated the clinically important differentiation of lymphoma from other malignant CNS tumours with potentially overlapping frozen-section morphology, including glioma, metastasis, and embryonal tumour. Distinguishing lymphoma from these entities is particularly important intraoperatively because a diagnosis of lymphoma may favour limited biopsy rather than extensive tumour resection. The task was formulated as a binary classification problem, with lymphoma as one class and glioma, metastasis, and embryonal tumour combined as the other class.

\noindent\textbf{ATRX alteration prediction.} This task assessed the prediction of ATRX alteration status, which is commonly associated with astrocytic lineage in diffuse gliomas. The task included two classes: ATRX-unaltered and ATRX-altered.

\noindent\textbf{1p/19q codeletion prediction.} This task assessed the prediction of 1p/19q codeletion status, a defining molecular feature of oligodendroglioma in the appropriate diagnostic context. The task included two classes: 1p/19q-non-codeleted and 1p/19q-codeleted.

\noindent\textbf{H3 K27 alteration prediction.} This task evaluated the prediction of H3 K27 alteration status, which is diagnostically relevant to diffuse midline glioma and related CNS tumour entities. The task included two classes: H3 K27-unaltered and H3 K27-altered.

\noindent\textbf{H3 G34 mutation prediction.} This task assessed the prediction of H3 G34 mutation status, a molecular alteration associated with diffuse hemispheric glioma. The task included two classes: H3 G34-wildtype and H3 G34-mutant.

\noindent\textbf{BRAF mutation prediction.} This task evaluated the prediction of BRAF mutation status, which is clinically relevant in several circumscribed gliomas and glioneuronal tumours. The task included two classes: BRAF-wildtype and BRAF-mutant.

\noindent\textbf{CDKN2A/B deletion prediction.} This task evaluated the prediction of CDKN2A/B deletion status, which has diagnostic and prognostic implications in selected glioma entities. The task included two classes: CDKN2A/B-retained and CDKN2A/B-deleted.

\noindent\textbf{TP53 mutation prediction.} This task assessed the prediction of TP53 mutation status, which is associated with several CNS tumour entities and contributes to molecular characterization. The task included two classes: TP53-wildtype and TP53-mutant.

\noindent\textbf{IDH1/IDH2 mutation prediction.} This task evaluated the prediction of IDH1/IDH2 mutation status, a central molecular feature in the integrated classification and prognostic stratification of adult-type diffuse gliomas. The task included two classes: IDH1/IDH2-wildtype and IDH1/IDH2-mutant, with the latter defined by a mutation in either gene.

\noindent\textbf{MGMT promoter methylation prediction.} This task assessed the prediction of MGMT promoter methylation status, an important predictive biomarker for response to alkylating chemotherapy in gliomas. The task included two classes: MGMT promoter-unmethylated and MGMT promoter-methylated.

\noindent\textbf{Chromosome 7 gain/chromosome 10 loss prediction.} This task assessed the prediction of concurrent chromosome 7 gain and chromosome 10 loss (+7/-10), a characteristic copy-number signature associated with IDH-wildtype glioblastoma. The task included two classes: +7/-10-negative and +7/-10-positive.

\noindent\textbf{EGFR amplification prediction.} This task evaluated the prediction of EGFR amplification status, a molecular feature frequently associated with IDH-wildtype glioblastoma and relevant to integrated diagnosis. The task included two classes: EGFR-non-amplified and EGFR-amplified.

\noindent\textbf{TERT promoter mutation prediction.} This task evaluated the prediction of TERT promoter mutation status, a molecular alteration with diagnostic and prognostic relevance in diffuse gliomas and other CNS tumours. The task included two classes: TERT promoter-wildtype and TERT promoter-mutant.

\subsection{Whole-slide image preprocessing}

\noindent\textbf{Tissue patch extraction and quality control.} The purpose of preprocessing was to remove non-informative regions while retaining a broad spectrum of diagnostically relevant morphology. Whole-slide images were first processed at the highest available magnification using Trident, a standardized pathology image processing framework \cite{zhang2025standardizing,vaidya2025molecular}, to preserve cellular details for subsequent representation learning. Tissue-containing regions were identified from each WSI, and image patches were extracted at three spatial scales, including $224 \times 224$, $448 \times 448$ and $896 \times 896$ pixels. This multi-scale extraction strategy enabled the model to capture both cell-level nuclear patterns and larger tissue-level architecture, both of which are required for central nervous system tumour interpretation.

Patch-level filtering was then performed in the hue-saturation-value color space. Using threshold-based rules, patches dominated by non-tissue regions, including background, pen markings of different colors and staining contaminants, were excluded before representation learning. This step reduced computational redundancy and limited the influence of non-tissue artefacts. Importantly, this filtering step was not used to impose any diagnostic label, but only to define a cleaner unlabelled image corpus for self-supervised pretraining. After filtering, the remaining patches were used as candidates for diversity-preserving sampling.

\noindent\textbf{Hierarchical clustering-based data curation (HCDC).} To construct a compact yet morphologically diverse pretraining corpus, we used a feature clustering and sampling procedure separately for patches extracted at each spatial scale. Candidate patches of $224 \times 224$, $448 \times 448$ and $896 \times 896$ pixels were independently encoded using UNI, a general-purpose pathology foundation model, to obtain scale-specific patch-level feature embeddings \cite{chen2024towards}. These embeddings were used only for image-space organization and patch selection, rather than as supervision for the proposed model. Following a hierarchical data curation strategy \cite{chen2025revisiting}, candidate patches at each scale were first grouped into 100,000 fine-grained clusters and then merged into 10,000 higher-level clusters. This scale-wise clustering design reduced the dominance of repetitive morphology from large tissue regions while preserving rare or heterogeneous patterns that may be underrepresented by naive random sampling from WSIs.

From the resulting cluster structure, patches were randomly sampled within each top-level cluster, with up to 400 patches selected from each cluster when available. This strategy generated approximately 11 million image patches for self-supervised training. Compared with naive random sampling from all WSIs, cluster-guided sampling better matches the objective of foundation model pretraining, because it increases morphological coverage without requiring additional manual annotation. The whole preprocessing pipeline is shown in Fig.~\ref{fig1}{d}.

\subsection{Foundation model development}

\noindent\textbf{Model initialization and parameter-efficient adaptation.} FOCUS was initialized from H-Optimus-1, a large-scale pathology foundation model, and adapted to CNS frozen-section pathology using low-rank adaptation (LoRA) \cite{scalbert2026hoptimus1,hu2022lora}. For each frozen projection matrix $W_0$, LoRA introduces a trainable low-rank update:

\begin{equation}
    W = W_0 + \frac{\alpha}{r}\Delta W,\quad \Delta W = BA,
\end{equation}

where $A$ and $B$ are low-rank trainable matrices. During pretraining, all backbone parameters were frozen, and LoRA modules with rank $r=8$ were inserted only into the query, key and value projections of each transformer block. For a ViT-G backbone with 40 transformer layers and an embedding dimension of 1536, this configuration introduces approximately 1.97 million trainable parameters, corresponding to only 0.69\% of the full QKV projection parameters and approximately 0.18\% of a 1.1-billion-parameter backbone. This parameter-efficient design enabled adaptation to frozen-section artefacts and CNS tumour morphology while preserving the general pathology representation learned by the pretrained backbone.

\noindent\textbf{Self-supervised teacher-student training.} The model was trained within a DINOv2-style self-supervised teacher-student framework \cite{oquab2023dinov2}, as illustrated in Fig.~\ref{fig1}{e}. The teacher network was updated by exponential moving average of the student parameters, and the student was optimized using the DINOv2 self-distillation objective. Because the training patches were unlabelled, this framework allowed the model to learn visual representations from large-scale frozen-section morphology without requiring patch-level annotation.

The data augmentation strategy called Frozen-Aware Multi-View (FAMV) was specifically modified for frozen pathology. Global views were used to capture tissue-level context and were augmented with stronger color perturbations, reflecting the staining variability commonly seen in frozen sections. Local views were used to emphasize cellular and nuclear details, with half of them receiving stronger texture degradation to mimic the loss of fine nuclear information and local blur caused by frozen-section preparation. In addition, a multi-scale view alignment mechanism was introduced: the other half of local views were generated by downsampling a large-field crop covering the same region as the global view, thereby forcing the model to align representations of the same tissue region observed at different resolutions. This design was intended to improve sensitivity to local cellular features while maintaining consistency with the surrounding histological architecture.

\noindent\textbf{Training objective and implementation.} The final pretraining objective followed the DINOv2 training recipe and combined image-level self-distillation with patch-level masked prediction \cite{oquab2023dinov2}. For image-level learning, the student was trained to match the teacher probability distribution over prototype assignments computed from the class tokens of different global and local views. For patch-level learning, masked patch tokens from the student were aligned with the corresponding visible patch tokens from the teacher, encouraging the encoder to recover local morphology from incomplete observations. Sinkhorn-Knopp centering and KoLeo regularization were used to stabilize prototype assignments and improve feature dispersion, respectively. In addition to these standard DINOv2 objectives, our multi-scale view alignment encouraged representations of the same tissue region at different resolutions to remain consistent.

The pretraining was performed for 250,000 iterations on 4 $\times$ 80 GB NVIDIA A100 GPUs, using a total batch size of 128. The student network was optimized with AdamW and a cosine learning-rate schedule after a 25,000-step warm-up. The peak and final learning rates were $5 \times 10^{-5}$ and $1 \times 10^{-6}$, respectively. The teacher network was updated as an exponential moving average of the student network, with the momentum coefficient increased from 0.992 to 1.000 during training. Multi-crop self-supervision used two global views and eight local views per image, with automatic mixed precision enabled throughout training. Full pretraining hyperparameters are provided in Supplementary Table~\ref{tabs_config_pretrain}.

\subsection{Downstream model development and evaluation}

\noindent\textbf{Slide-level feature aggregation and task-specific classifiers.} For each downstream task, WSIs were first tiled into image patches using the same preprocessing pipeline. Patch-level embeddings were extracted by the frozen FOCUS encoder. Because the available labels were defined at the slide or case level rather than at the patch level, downstream prediction was formulated as a weakly supervised multiple-instance learning (MIL) problem. Patch embeddings from each slide were aggregated into a slide-level representation using a MIL module, followed by a task-specific prediction head. The MIL aggregator was selected by comparing representative aggregation methods, including ABMIL~\cite{ilse2018attention}, CLAM~\cite{lu2021data}, TransMIL~\cite{shao2021transmil} and DSMIL~\cite{li2021dual}, on the validation set, and the selected aggregator DSMIL was then used for downstream model development. This design followed the common computational pathology workflow in which preprocessing, feature extraction, feature aggregation and final prediction are separated into modular components~\cite{XU2025103456}.

The same downstream training protocol was applied to FOCUS and all comparator foundation models. The evaluated baselines included UNI~\cite{chen2024towards}, Phikon-v2~\cite{filiot2024phikon}, Virchow~\cite{vorontsov2024foundation}, Virchow2~\cite{zimmermann2024virchow2}, CRISP~\cite{zhao2026clinically}, GigaPath~\cite{xu2024whole}, H-Optimus-0~\cite{hoptimus0} and H-Optimus-1~\cite{scalbert2026hoptimus1}. To ensure that the comparison primarily reflected the quality of the learned representations, the downstream aggregator, task-specific head, evaluation metrics and task definitions were kept consistent across models whenever possible.

For each downstream task, the MIL aggregator and prediction head were trained for 20 epochs using the Adam optimizer with a learning rate of \(2 \times 10^{-5}\) and a weight decay of \(1 \times 10^{-5}\). Early stopping was applied with a patience of 5 epochs, and the model was optimized using cross-entropy loss. When applicable, class imbalance was addressed using weighted sampling. Hyperparameters were selected on the validation set and fixed before final evaluation on the held-out test set and external validation cohorts.

\noindent\textbf{Few-shot evaluation.} To assess label efficiency, few-shot experiments were performed for both diagnostic and molecular prediction tasks. For each downstream task, the task-specific training set was subsampled to contain $n \in \{1,2,4,8,16,32\}$ labelled examples per class or per task-specific sampling unit, while the test set remained unchanged. A downstream classifier was then trained using the limited labelled subset, and the AUC was calculated on the same held-out test set. This design directly tested whether FOCUS could provide useful representations when only a small number of labelled frozen-section slides were available, which is particularly relevant for rare CNS tumour subtypes and molecular endpoints with limited sample sizes.

\noindent\textbf{Hierarchical diagnosis system.} The diagnostic tasks were organized as a clinically oriented hierarchical system rather than as a single flat multiclass task. All reference diagnoses were harmonized according to the 2021 WHO Classification of tumours of the Central Nervous System, which provides the current taxonomic basis for integrating histological and molecular information in CNS tumour diagnosis \cite{louis20212021}. Only slides with reference diagnoses that could be mapped unambiguously to the prespecified hierarchy were included in hierarchical evaluation. Each slide was assigned to a leaf diagnosis and to its corresponding ancestral nodes in the hierarchy. This hierarchical structure is particularly relevant to frozen-section assessment. Even when an exact subtype cannot be established, assignment to the correct diagnostic branch can narrow the differential diagnosis while avoiding unsupported specificity.

The hierarchy was structured to approximate the sequential decision process used during frozen-section interpretation. The first decision separated non-neoplastic brain tissue from tumour. Among tumour slides, metastasis was distinguished from other CNS tumours, because recognizing a metastatic lesion during surgery immediately changes the interpretation of tumour origin and subsequent clinical communication. The remaining tumour spectrum was then divided into neuroepithelial and non-neuroepithelial tumours. Within the neuroepithelial branch, glioma or glioneuronal tumour was distinguished from embryonal tumour and ependymoma. A more specific distinction between glioma and glioneuronal tumour was then performed, as this separation can be subtle on frozen sections but remains important for pathological diagnosis. Within the non-neuroepithelial branch, meningioma was first separated from other common non-neuroepithelial tumours, owing to its high frequency and characteristic role in extra-axial tumour diagnosis. The remaining cases were further classified as pituitary neuroendocrine tumour (PitNET), SFT/mesenchymal tumour, lymphoma, craniopharyngioma or schwannoma. For glioma-specific evaluation, diffuse glioma was distinguished from circumscribed glioma. Diffuse gliomas were further classified as astrocytoma, IDH-mutant; oligodendroglioma, IDH-mutant and 1p/19q-codeleted; glioblastoma, IDH-wildtype; diffuse hemispheric glioma, H3 G34-mutant; or diffuse midline glioma, H3 K27-altered. Circumscribed gliomas were further evaluated using the pilocytic astrocytoma versus pleomorphic xanthoastrocytoma task. This organization follows the clinical logic that broad branch-level decisions are often required before fine-grained tumour typing can be reliably attempted in frozen sections.

In addition to the ten hierarchy-derived diagnostic tasks, we included one clinically motivated cross-branch differential diagnosis: lymphoma versus glioma or metastasis or embryonal tumour. This task was not treated as an additional taxonomic level because it represents a common and consequential intraoperative differential diagnosis. Glioma, metastasis, embryonal tumour and lymphoma can overlap in frozen morphology but imply substantially different surgical and therapeutic strategies.

\noindent\textbf{Integration of clinical information.} Because histopathological interpretation in routine clinical practice is informed by relevant clinical context, we incorporated patient age and anatomical sampling location into the multiple-instance learning framework. Sampling locations were first extracted from the original Chinese clinical records and mapped to International Classification of Diseases for Oncology, Third Edition (ICD-O-3) topography codes using a constrained large language model workflow. Candidate mappings were restricted to a predefined catalogue of ICD-O-3 topography sites relevant to neurosurgical specimens and were subjected to programmatic checks for completeness and validity. For compound sampling locations, multiple valid anatomical codes were retained when applicable. For slide-level modelling, patch-level features were aggregated using DSMIL to derive a unified whole-slide representation. Patient age was standardized using statistics calculated exclusively from the training cohort and encoded together with a missingness indicator, after which the resulting features were projected into a 32-dimensional embedding. ICD-O-3 codes were encoded as fixed-vocabulary binary multi-hot vectors and projected into a 256-dimensional embedding, allowing multiple anatomical sites to be represented for a single sample when necessary. The slide, age, and anatomical-location embeddings were then concatenated and passed through a multilayer classifier to generate predictions. This design enabled the model to jointly integrate histomorphological features with clinically relevant age and anatomical information within a unified multimodal MIL framework.

\noindent\textbf{Challenging scenarios analysis.} We performed challenging-scenario analyses to evaluate whether FOCUS could provide additional support in intraoperative settings where frozen-section diagnosis is most prone to uncertainty or error. Two prespecified subgroups were analyzed using paired intraoperative frozen-section reports and final postoperative pathology reports. Candidate cases were identified by report review and manually checked to ensure that they matched the intended definitions. All included cases had a definitive postoperative diagnosis, which served as the reference standard.

The first subgroup, termed the \textbf{Uncertainty} subgroup, comprised 77 patients selected from the internal test set whose intraoperative reports contained explicit uncertainty expressions, including ``possible'', ``favor'', ``suspicious for'', ``cannot exclude'' or ``defer to permanent section''. These cases represented situations in which limited sampling, frozen-section artefacts or time-constrained interpretation precluded a definitive diagnosis, potentially prompting additional sampling, prolonging surgery or favoring a more conservative surgical strategy. The second subgroup, termed the \textbf{Discordance} subgroup, comprised 66 patients selected from the internal test set whose original intraoperative diagnosis differed from the final postoperative diagnosis after both were mapped to the same CNS diagnostic hierarchy \cite{plesec2007frozen}. This subgroup represented potential intraoperative diagnostic errors that could affect sampling, tumour-origin assessment, resection strategy or communication with the surgical team.

For both subgroups, FOCUS predictions and original intraoperative diagnoses were compared with the final postoperative diagnosis using the same task definitions and evaluation metrics as the corresponding diagnostic tasks. In discordant cases, a model prediction was considered potentially corrective if it matched the final postoperative diagnosis when the original frozen-section diagnosis did not. In ambiguous cases, model utility was assessed by whether FOCUS provided a definite prediction concordant with the final postoperative diagnosis.

\noindent\textbf{Reader study of pathologist--AI collaboration.} We conducted a paired reader study to evaluate the impact of FOCUS assistance on intraoperative frozen-section diagnosis. Six pathologists, comprising three junior (mean experience, 6 years; individual experience, 8, 3 and 7 years) and three senior (mean experience, 15 years; individual experience, 10, 13 and 22 years) readers, participated in both assessments. The reader study evaluated diagnostic tasks only, because pathologists could not determine molecular marker status intraoperatively. The hierarchical diagnostic assessment included 243 frozen-section WSIs from three cohorts: 100 randomly selected slides, 77 slides with explicit expressions of diagnostic uncertainty in the original intraoperative reports, and 66 slides with intraoperative diagnoses discordant with final postoperative pathological diagnoses. The lymphoma differentiation assessment included an independent cohort of 102 frozen-section WSIs from 96 patients, comprising 81 lymphoma and 21 non-lymphoma slides. Reference diagnoses for both assessments were derived from final postoperative pathological diagnoses.

Both assessments followed a paired design. Readers first interpreted the WSIs without access to FOCUS predictions and, after a two-week washout period, reassessed the same slides with FOCUS assistance. Each reader provided a diagnostic report for each slide under both conditions. Reader reports were standardized through keyword matching to predefined diagnostic categories. This procedure was applied to both assessments, with the resulting labels used for hierarchical and binary evaluation, respectively. All evaluations were performed at the slide level, enabling paired comparisons of unaided and AI-assisted interpretations for each reader and slide.

The primary metrics for the hierarchical assessment were hierarchical F1 (H-F1) and depth-weighted hierarchical F1 (DH-F1). Reader diagnoses and FOCUS predictions were mapped to the same four-level diagnostic hierarchy as the reference diagnoses. For each slide, the maximum correct depth was defined as the number of consecutive diagnostic nodes shared by the predicted and reference paths, starting from the root of the hierarchy. H-F1 summarized agreement between predicted and reference paths across slides, whereas DH-F1 assigned greater weight to agreement at deeper diagnostic levels. We additionally calculated the maximum correct classification depth for each slide and classification accuracy at each of the four diagnostic levels. Differences in H-F1 and DH-F1 between unaided and AI-assisted interpretations were assessed using two-sided paired randomization tests with 10,000 permutations. Reported $P$ values were adjusted for multiple comparisons using the Benjamini--Hochberg correction. For each reader and cohort, we calculated the slide-level change in maximum correct classification depth as the AI-assisted depth minus the unaided depth. Mean changes were reported with normal-approximation 95\% confidence intervals. We further quantified diagnostic transitions following FOCUS assistance, including wrong-to-correct conversions, increases in maximum correct depth, and correct-to-wrong conversions. Wrong-to-correct conversions were also counted among slides with increased maximum correct depth. All paired analyses included only slides with identical reference diagnostic paths under both conditions. Performance was analysed separately by reader experience level and cohort.

For the lymphoma differentiation assessment, standardized reader diagnoses were classified as lymphoma-positive or lymphoma-negative, irrespective of diagnostic certainty. We calculated binary accuracy and lymphoma recall for each reader before and after FOCUS assistance, using final postoperative pathological diagnoses as the reference. Group-level performance was summarized by averaging the three reader-level values separately for junior and senior readers. We also evaluated the standalone performance of FOCUS across all 102 slides using the same metrics. Differences between unaided and AI-assisted performance were assessed using two-sided paired randomization tests on group-mean reader rates, with all readings from each patient exchanged together. The resulting $P$ values were adjusted using the Benjamini--Hochberg procedure across the four group-by-metric comparisons.

\noindent\textbf{Generalization and representation-level robustness.} To determine whether CNS-specific frozen-section pretraining preserved broader pathological transferability, FOCUS was further evaluated on publicly available pan-cancer frozen-section slides. These data were used exclusively for pan-cancer diagnostic evaluation and were not included in CNS-specific model selection. This analysis assessed whether the model retained transferable representations across tissue types while adapting to CNS intraoperative pathology.

Representation-level robustness was additionally assessed through patch-level classification using Ivy-GAP data. Twenty WSIs that were not included in foundation model pretraining were randomly selected. Patches of 224 $\times$ 224 pixels were extracted and assigned to six predefined histological regions: infiltrating tumour, cellular tumour, perinecrotic zone, pseudopalisading cells around necrosis, hyperplastic blood vessels in cellular tumour and necrosis. To construct a balanced evaluation dataset, 3,000 patches were sampled from each category. A linear-probing protocol was used, in which the encoder remained frozen and only a linear classifier was trained. The classifier was optimized using SGD with a learning rate of $5 \times 10^{-5}$ and a weight decay of $1 \times 10^{-4}$. In contrast to slide-level MIL tasks, this analysis directly evaluated the ability of the frozen encoder to distinguish local histological phenotypes without the influence of slide-level feature aggregation.

Multi-scale robustness was evaluated by extracting representations from matched tissue regions at three spatial scales, corresponding to patch sizes of $224 \times 224$, $448 \times 448$ and $896 \times 896$ pixels. Performance across scales was compared using an identical downstream classifier configuration to determine whether the learned representations remained stable under changes in the relative contributions of cellular detail and tissue-level context.

\subsection{Interpretability analysis}

\noindent\textbf{Attention-based localization of diagnostic evidence.} To examine whether the slide-level predictions were supported by pathologically meaningful regions, we generated attention heatmaps from the downstream multiple-instance learning models. For each whole-slide image, tissue patches were first encoded by the frozen FOCUS encoder, and the resulting patch-level embeddings were passed to the task-specific aggregator. The normalized attention weights assigned by the aggregator were then projected back to the original spatial coordinates of the slide to obtain a slide-level heatmap, following the attention-based weakly supervised learning paradigm commonly used in computational pathology~\cite{lu2021data}.

\noindent\textbf{Prototype enrichment analysis.} To identify recurrent patch phenotypes associated with slide-level molecular status, we performed a weakly supervised prototype enrichment analysis using only WSI-level labels and model-derived patch contribution scores. For each molecular task, slides from the discovery split were used to construct the prototype dictionary, whereas slides from the evaluation split were reserved for enrichment testing. Patch contribution scores were obtained from the task-specific MIL model attention branch and normalized within each slide. Discovery patches were then sampled from configurable pools, including high-attention patches from molecular-positive slides, high-attention patches from molecular-negative slides, and randomly selected background patches. Within the shared-prototype setting, selected patches from all tasks were pooled and duplicate selections with the same WSI identifier and patch index were removed before clustering. Patch features were transformed using $\ell_2$ normalization, and clustered into 10 prototypes using mini-batch $k$-means~\cite{sculley2010webscale}. Each prototype was represented by its centroid in the normalized feature space, and the nearest patches to each centroid were retained with their WSI identifier, coordinates, molecular label, contribution score and crop path for downstream pathological review.

To quantify whether a prototype was preferentially associated with a molecular state, each evaluation slide was projected onto the learned prototype dictionary. Specifically, for a slide $s$, only the top-$N$ finite-attention patches were used for summary statistics, with $N=100$ by default. These patches were assigned to the nearest prototype centroid, yielding the slide-level prototype proportion $P_{s,k}=n_{s,k}/N_s$, where $n_{s,k}$ denotes the number of selected patches assigned to prototype $k$ and $N_s$ denotes the number of selected patches available for slide $s$. The mean attention score among patches assigned to each prototype was also recorded when at least one patch was present. For each prototype and task, the distribution of $P_{s,k}$ was compared between molecular-positive and molecular-negative evaluation slides using a two-sided Mann-Whitney U test~\cite{mann1947test}. The reported effect size was the difference in mean prototype proportion between positive and negative groups, and prototypes were annotated as positive- or negative-enriched according to the sign of this difference. To account for testing multiple prototypes within each task, nominal $P$ values were adjusted using the Benjamini-Hochberg false discovery rate procedure~\cite{benjamini1995controlling}.

\subsection{Metrics and statistical analysis}

\noindent\textbf{Evaluation metrics.} For binary diagnostic and molecular prediction tasks, model discrimination was primarily evaluated using the area under the receiver operating characteristic curve (AUC). Accuracy, sensitivity, specificity and F1 score were reported as secondary metrics to characterize threshold-dependent performance. For threshold-dependent metrics, the decision threshold was selected on the validation set using the Youden index and then fixed before evaluation on the corresponding test or external validation cohort.

For multiclass diagnostic tasks, macro-AUC was calculated by averaging one-versus-rest AUC values across evaluable classes. Accuracy, macro-sensitivity, macro-specificity and macro-F1 score were additionally reported to evaluate average performance. For few-shot experiments, downstream models were trained using $n\in\{1,2,4,8,16,32\}$ labelled samples per class, and AUC was calculated on the same held-out test set to evaluate label efficiency under controlled supervision.

For clinically critical differential diagnostic tasks, we additionally reported specificity at 95\% sensitivity to assess the model's ability to minimize false-positive predictions while maintaining a high detection rate for target diseases. The decision threshold was determined separately for each evaluation dataset based on its empirical ROC curve, ensuring a sensitivity of at least 95\%.

For the patch-level linear-probing experiment, accuracy, macro-F1 score and Balanced accuracy were reported as the mean and standard deviation across five cross-validation folds.

For hierarchical diagnosis, the diagnostic taxonomy was constructed according to the World Health Organization classification of central nervous system tumours and organized as a tree covering common intraoperative diagnostic entities. Because a prediction that remains within the correct diagnostic branch is clinically less severe than a prediction assigned to a distant branch, we evaluated hierarchical performance in addition to conventional leaf-level classification. For each sample (i), let $C_t(i)$ and $C_p(i)$ denote the sets of diagnostic nodes along the paths from the root to the reference and predicted labels, respectively, excluding the common non-diagnostic root node. Let $I_i=C_t(i)\cap C_p(i)$ denote their shared nodes. Hierarchical precision (H-Precision) and hierarchical recall (H-Recall) were calculated across the evaluation set as

\begin{equation}
H\text{-}Precision =
\frac{\sum_i |I_i|}
{\sum_i |C_p(i)|},
\qquad
H\text{-}Recall =
\frac{\sum_i |I_i|}
{\sum_i |C_t(i)|}.
\end{equation}

The hierarchical F1 score (H-F1) was then defined as the harmonic mean of hierarchical precision and recall:

\begin{equation}
H\text{-}F1 =
\frac{2H\text{-}Precision \cdot H\text{-}Recall}
{H\text{-}Precision + H\text{-}Recall}.
\end{equation}

To place greater emphasis on agreement at deeper and more specific diagnostic levels, we additionally calculated the depth-weighted hierarchical F1 score (DH-F1). Each node (v) was assigned a weight according to its depth in the diagnostic tree,

\begin{equation}
w(v)=\mathrm{depth}(v) \in \{1,2,3,4\},
\qquad
W(C)=\sum_{v\in C}w(v),
\end{equation}

where $W(C)$ denotes the cumulative depth weight of a node set $C$. Depth-weighted hierarchical precision (DH-Precision) and recall (DH-Recall) were defined as

\begin{equation}
DH\text{-}Precision =
\frac{\sum_i W(I_i)}
{\sum_i W(C_p(i))},
\qquad
DH\text{-}Recall =
\frac{\sum_i W(I_i)}
{\sum_i W(C_t(i))}.
\end{equation}

The corresponding DH-F1 was calculated as

\begin{equation}
DH\text{-}F1 =
\frac{2DH\text{-}Precision \cdot DH\text{-}Recall}
{DH\text{-}Precision + DH\text{-}Recall}.
\end{equation}

Because deeper nodes correspond to more fine-grained diagnoses, DH-F1 assigns greater importance to agreement at deeper levels of the diagnostic hierarchy. We further calculated the maximum correct depth for each sample, defined as the deepest level at which the predicted and reference diagnostic paths remained identical, and reported its mean across all samples as an additional measure of how far predictions progressed correctly through the hierarchy. This hierarchical evaluation followed the rationale of hierarchical pathological WSI classification, where diagnostic labels are naturally organized into clinically meaningful parent-child relationships\cite{li2025diagnostic}.

\noindent\textbf{Statistical analysis.} To compare overall model performance across multiple tasks, two-sided paired Wilcoxon signed-rank tests were performed, and p-values were adjusted for multiple comparisons using the Benjamini--Hochberg correction. For the diagnostic tasks and biomarker-prediction tasks, comparisons of AUC were performed separately in the internal, temporal, and external validation cohorts. Benjamini--Hochberg correction was applied across evaluable task-specific comparisons. In pan-cancer analyses, comparisons of AUC were performed separately within the external-validation and internal-only task groups. AUC differences no greater than $10^{-6}$ were classified as ties and excluded from these tests. The multi-size analysis and ablation study used 23 complete task pairs and reported nominal $P$ values.

Confidence intervals were estimated using analysis-specific procedures. For binary-task AUCs, 95\% confidence intervals (CIs) were calculated using DeLong's method. For multiclass AUCs and other performance metrics, including specificity at 95\% sensitivity, hierarchical and level-wise metrics, 95\% CIs were estimated using 2,000 nonparametric bootstrap resamples. Cross-task mean AUC differences and task-level effects in the patch size and ablation analyses were evaluated by bootstrap resampling of complete task pairs. Prototype proportions were compared between molecular-positive and molecular-negative slides using two-sided Mann--Whitney $U$ tests, with Benjamini--Hochberg correction applied across the ten prototypes within each molecular task.

Comparisons of H-F1 and DH-F1 used paired, two-sided permutation tests with 10,000 case-level permutations. For reader study, AI-versus-reader and AI-assisted-versus-unaided comparisons were corrected by Benjamini--Hochberg within each study group. For reader-level changes in maximum correct classification depth, 95\% CIs were calculated using the normal approximation to the paired-case mean differences. Maximum correct depth and level-wise accuracy compared AI with the mean of the six readers. Their permutation distributions used randomly reversed case-level paired differences. Benjamini--Hochberg correction was applied within the corresponding analysis families.

Inter-reader agreement was quantified using Fleiss' $\kappa$ before and after AI assistance. Diagnoses were counted as concordant only when their complete hierarchical paths matched. Analyses were performed for each group, the pooled dataset, and junior and senior readers separately. Changes in $\kappa$ were assigned 95\% confidence intervals using 2,000 paired bootstrap resamples clustered by slide.

All tests were two-sided. Statistical significance was defined as an adjusted $P<0.05$ when multiplicity correction was applied and a nominal $P<0.05$ otherwise.

\subsection{Data availability}

Publicly available frozen-section WSIs used for model pretraining and public validation were obtained from TCGA, CPTAC, HCMI, CGCI and CDDP-EAGLE through the NIH Genomic Data Commons (\url{https://portal.gdc.cancer.gov/}), from Ivy-GAP through the Allen Institute for Brain Science (\url{https://glioblastoma.alleninstitute.org/}), and from HEST-1k through the official repository (\url{https://github.com/mahmoodlab/HEST}). These datasets remain subject to the access conditions and terms of use specified by their respective data providers.

The institutional cohorts from Beijing Tiantan Hospital and the First Hospital of Hebei Medical University were retrospectively collected, fully de-identified and used under the relevant institutional review board approvals. Owing to patient privacy requirements and institutional data governance policies, the raw WSIs and associated clinical, pathological and molecular data are not publicly available. Researchers seeking access to specific de-identified subsets for academic and non-commercial purposes may submit a formal data-use proposal to the corresponding author. Requests will be reviewed on a case-by-case basis and will require approval from the participating institutions, execution of appropriate data-use agreements and compliance with applicable ethical, privacy and regulatory requirements.

\subsection{Code availability}
The source code and the FOCUS model can be accessed at \url{https://github.com/Extord1108/FOCUS}.

\bibliography{refs}

\section{Acknowledgements}

This work was supported by the New Generation Artificial Intelligence-National Science and Technology Major Project (2026ZD0125800), the National Key Research and Development Program of China (2024YFF1207400, and 2023YFC3402800), the National Natural Science Foundation of China (62333022, 82272084, 62027901, 92359302, and 92259301), Beijing Natural Science Foundation (JQ23034), Guangdong Basic and Applied Basic Research Foundation (2025A1515220109), Beijing Reform and Development Fund (2060361) and Beijing Neurosurgical Institute Major Project Plan. The funders had no role in study design, data collection and analysis, decision to publish or preparation of the manuscript.

\section{Authors' contributions}

X.L, Y.F.G and Z.Y.L conceived and designed the study. 
X.L, L.L, X.J.Z and W.L collected the clinical and pathological data, and coordinated the external validation cohorts. 
X.L and J.D provided neuropathological expertise. 
Y.F.G, H.T.L, and Y.Y.T curated the pretraining dataset, developed the foundation model and conducted the model training.
Y.F.G, H.T.L, Y.Y.T, R.Y, H.L.S and W.Z.X conducted the downstream experiments and analyzed the results. 
X.B.Z and S.Z performed quality control of the codebase and the results. 
Y.F.G designed the reader study, and X.L, L.L, J.D, M.Z, C.P and M.L participated as neuropathologists. 
Y.F.G prepared the original manuscript draft. 
X.L, T.J, C.L, J.T, Q.C and Z.Y.L supervised the study and critically revised the manuscript. 
X.L, J.T, C.L and Z.Y.L acquired funding. 
All authors reviewed and approved the final manuscript.

\section{Competing interests}
The authors declare no competing interests.

\setcounter{secnumdepth}{0}

\flushbottom

\appendix
\setcounter{figure}{0}
\setcounter{table}{0}

\renewcommand{\figurename}{Extended Data Figure}
\clearpage
\section{Extended Data}

\begin{figure}[htbp]
	\centering
	\includegraphics[width=.95\linewidth]{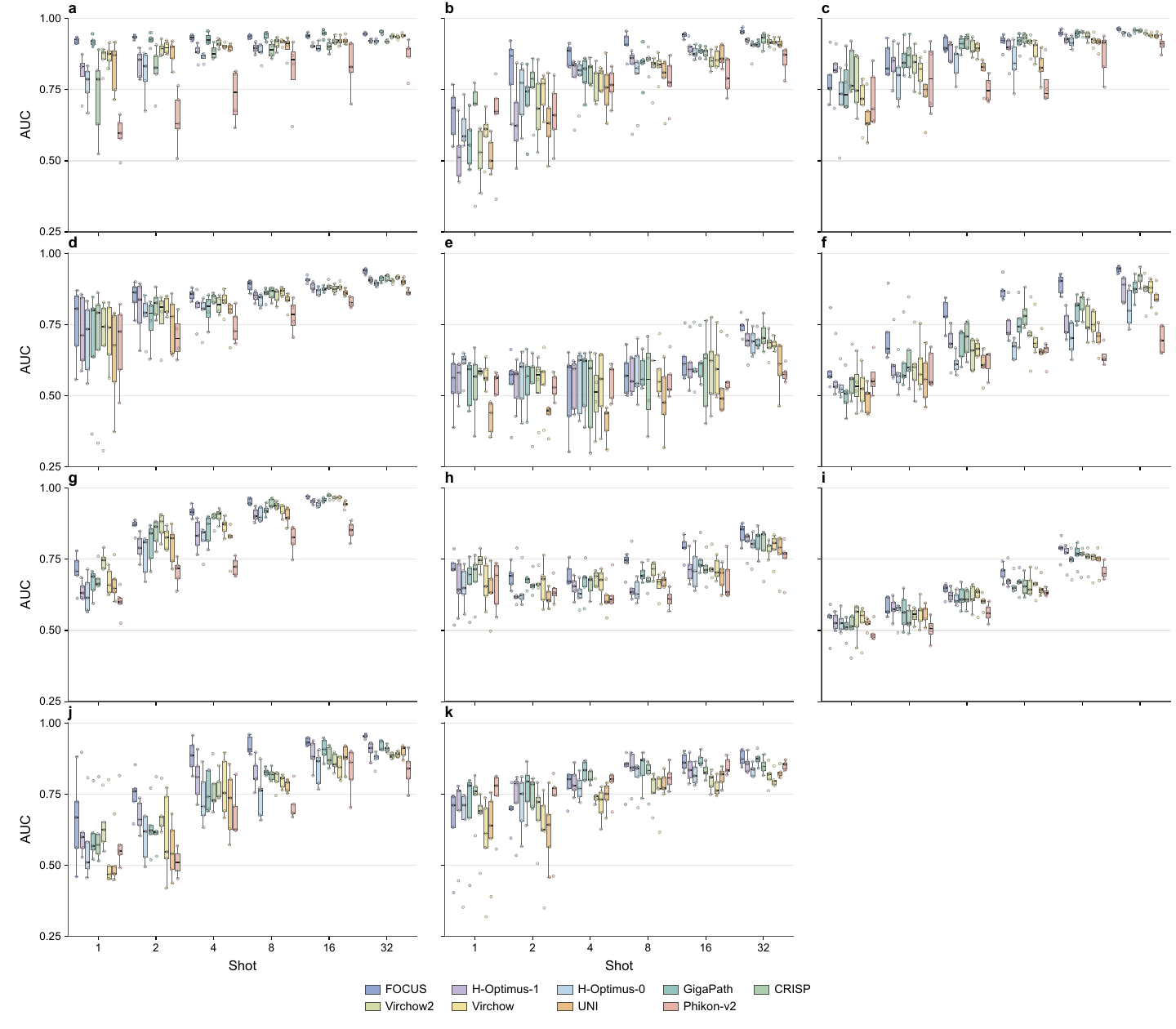}
    \refstepcounter{figure}
	\label{figs_diag_kshot}
\end{figure}
\noindent\textbf{Extended Data Figure 1. Few-shot performance across diagnostic classification tasks.} Box plots compare the area under the receiver operating characteristic curve (AUC) of nine pathology foundation models at 1, 2, 4, 8, 16 and 32-shot settings. Panels correspond to \textbf{a,} Non-neoplastic brain tissue vs tumour; \textbf{b,} Metastasis vs other CNS tumours; \textbf{c,} Neuroepithelial vs non-neuroepithelial tumour; \textbf{d,} Glioma or glioneuronal tumour vs embryonal tumour vs ependymoma; \textbf{e,} Glioma vs glioneuronal tumour; \textbf{f,} Meningioma vs other non-neuroepithelial tumours; \textbf{g,} PitNET vs SFT/mesenchymal tumour vs lymphoma vs craniopharyngioma vs schwannoma; \textbf{h,} Diffuse vs circumscribed glioma; \textbf{i,} Astrocytoma, IDH-mutant vs oligodendroglioma, IDH-mutant and 1p/19q-codeleted vs glioblastoma, IDH-wildtype vs diffuse hemispheric glioma, H3 G34-mutant vs diffuse midline glioma, H3 K27-altered; \textbf{j,} Pilocytic astrocytoma vs pleomorphic xanthoastrocytoma; \textbf{k,} Lymphoma differentiation. Boxes indicate the interquartile range, centre lines denote medians, whiskers extend to 1.5 times the interquartile range, and dots represent individual sampling runs. The shared AUC axis ranges from 0.25 to 1.00, and models are identified by the shared colour legend.

\clearpage

\begin{figure}[htbp]
	\centering
	\includegraphics[width=.95\linewidth]{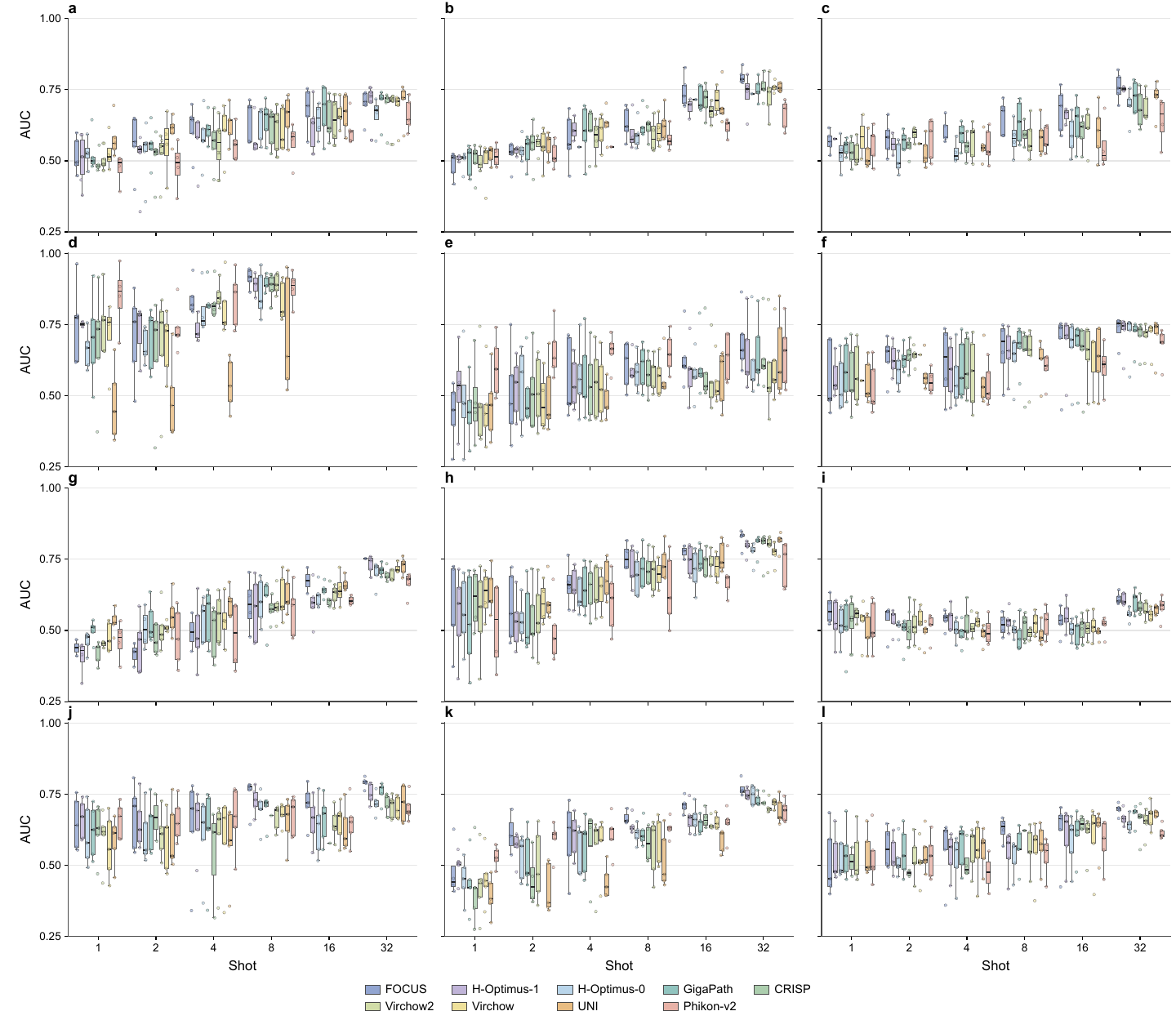}
    \refstepcounter{figure}
	\label{figs_biom_kshot}
\end{figure}
\noindent\textbf{Extended Data Figure 2. Few-shot performance across biomarker prediction tasks.} Box plots compare the area under the receiver operating characteristic curve (AUC) of nine pathology foundation models at 1, 2, 4, 8, 16 and 32-shot settings. Panels correspond to biomarker prediction tasks for \textbf{a,} ATRX alteration; \textbf{b,} 1p/19q codeletion; \textbf{c,} H3 K27 alteration; \textbf{d,} H3 G34 mutation; \textbf{e,} BRAF mutation; \textbf{f,} CDKN2A/B deletion; \textbf{g,} TP53 mutation; \textbf{h,} IDH1/IDH2 mutation; \textbf{i,} MGMT promoter methylation; \textbf{j,} chromosome 7 gain/chromosome 10 loss (+7/-10); \textbf{k,} EGFR amplification; and \textbf{l,} TERT promoter mutation. Boxes indicate the interquartile range, centre lines denote medians, whiskers extend to 1.5 times the interquartile range, and dots represent individual sampling runs. The shared AUC axis ranges from 0.25 to 1.00, and models are identified by the shared colour legend.

\clearpage

\begin{figure}[htbp]
	\centering
	\includegraphics[width=.8\linewidth]{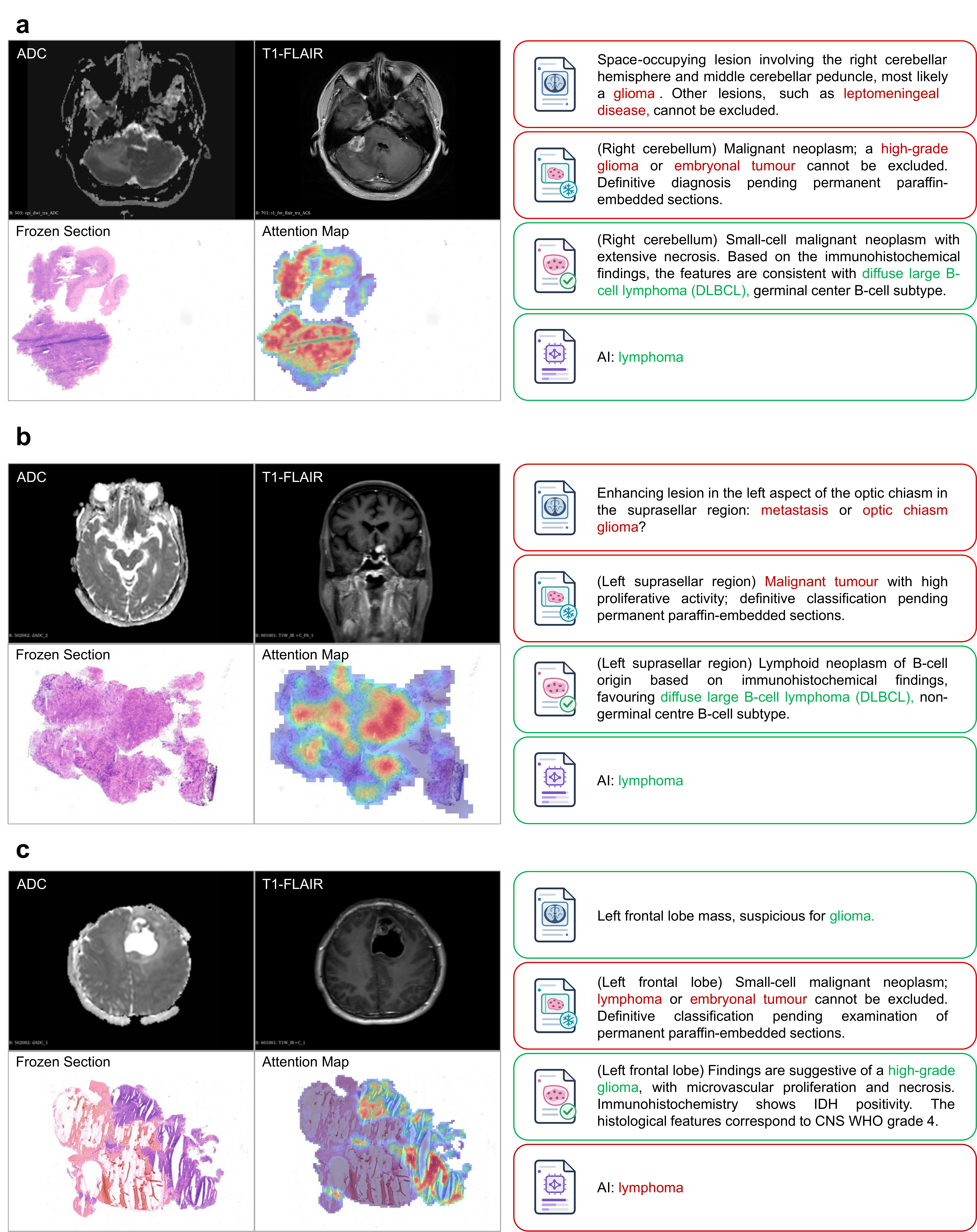}
    \refstepcounter{figure}
	\label{fig:case_study}
\end{figure}
\noindent\textbf{Extended Data Figure 3. Representative correct and incorrect AI predictions in lymphoma differentiation.} Three representative cases illustrate the strengths and limitations of lymphoma differentiation in this cohort. \textbf{a,b,} FOCUS correctly identified two diffuse large B-cell lymphomas despite non-lymphoma impressions from imaging and frozen pathology. \textbf{c,} Preoperative imaging correctly suggested glioma, but the frozen-section report raised lymphoma and FOCUS also predicted lymphoma. This densely cellular tumour lacked a distinctive growth pattern and comprised predominantly small round blue cells with high nuclear-to-cytoplasmic ratios and frequent apoptosis. These lymphoma-like features likely complicated differentiation from a neuroepithelial tumour for both FOCUS and the pathologist on frozen section.

\clearpage

\begin{figure}[htbp]
	\centering
	\includegraphics[width=.95\linewidth]{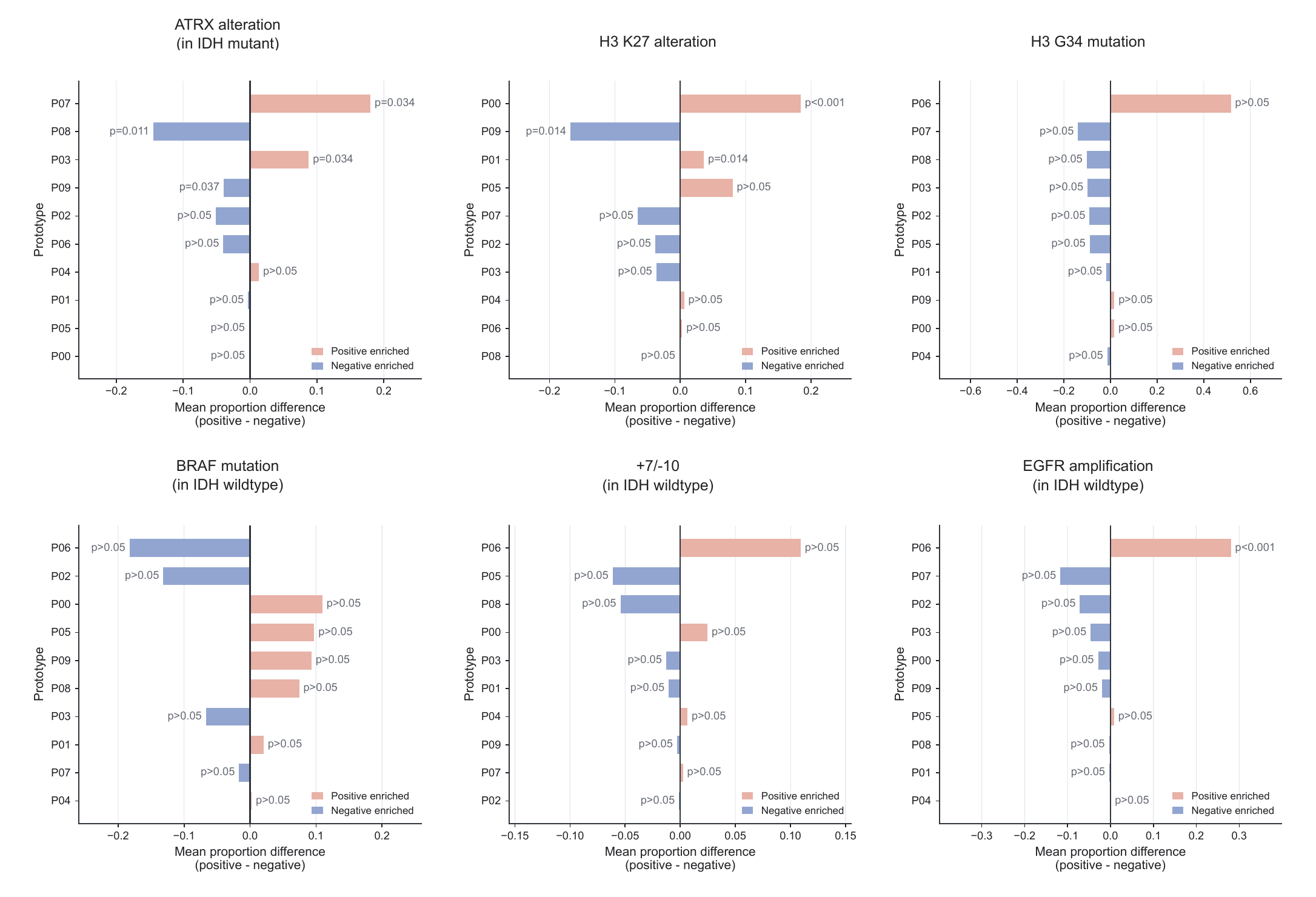}
    \refstepcounter{figure}
	\label{fig:prototype_enrichment}
\end{figure}
\noindent\textbf{Extended Data Figure 4. Prototype enrichment for molecular alterations.} Prototype-proportion differences are shown for ATRX alteration in IDH-mutant gliomas, H3 K27 alteration, H3 G34 mutation, BRAF mutation in IDH-wild-type gliomas, chromosome 7 gain/chromosome 10 loss in IDH-wild-type gliomas, and EGFR amplification in IDH-wild-type gliomas. Bars show the mean prototype-proportion difference between molecular-positive and molecular-negative slides. Salmon and blue bars indicate positive and negative enrichment, respectively. Each panel uses the shared ten-prototype dictionary; annotations report the corresponding enrichment-test \(P\) values.

\clearpage

\setcounter{figure}{0}
\setcounter{table}{0}

\renewcommand{\thefigure}{S\arabic{figure}}

\renewcommand{\thetable}{S\arabic{table}}

\renewcommand{\theHfigure}{supp.\arabic{figure}}
\renewcommand{\theHtable}{supp.\arabic{table}}


\renewcommand{\figurename}{Figure}
\renewcommand{\tablename}{Table}

\renewcommand{\thefigure}{S\arabic{figure}}
\renewcommand{\thetable}{S\arabic{table}}

\section{Supplementary Materials}

\begin{figure}[htbp]
	\centering
	\includegraphics[width=.95\linewidth]{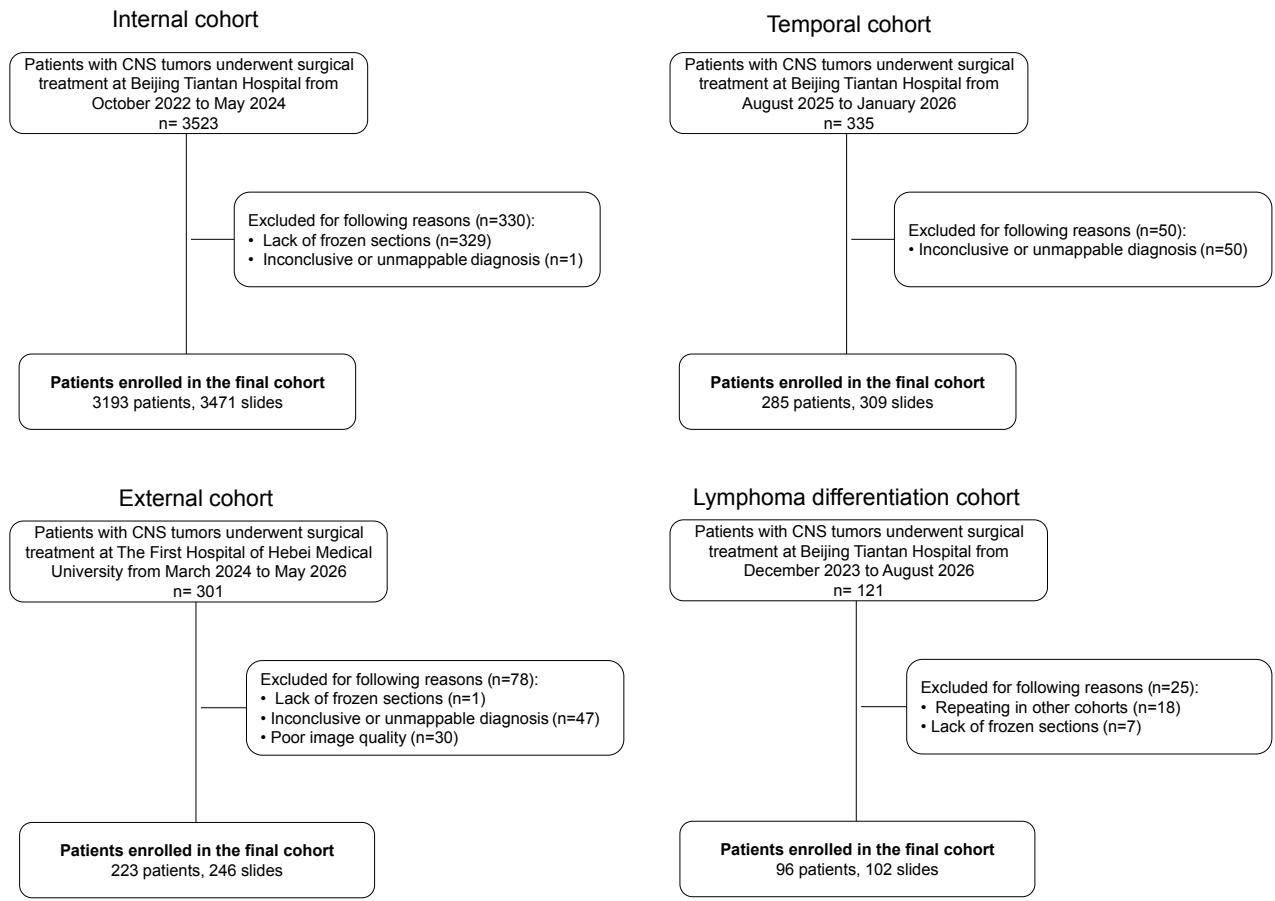}
	\caption{\textbf{Patient selection and cohort composition.} Flow diagrams summarize the inclusion and exclusion of patients who underwent surgical treatment for central nervous system (CNS) tumours.}
	\label{figs_inclu_exclu}
\end{figure} 

\begin{figure}[htbp]
	\centering
	\includegraphics[width=\linewidth]{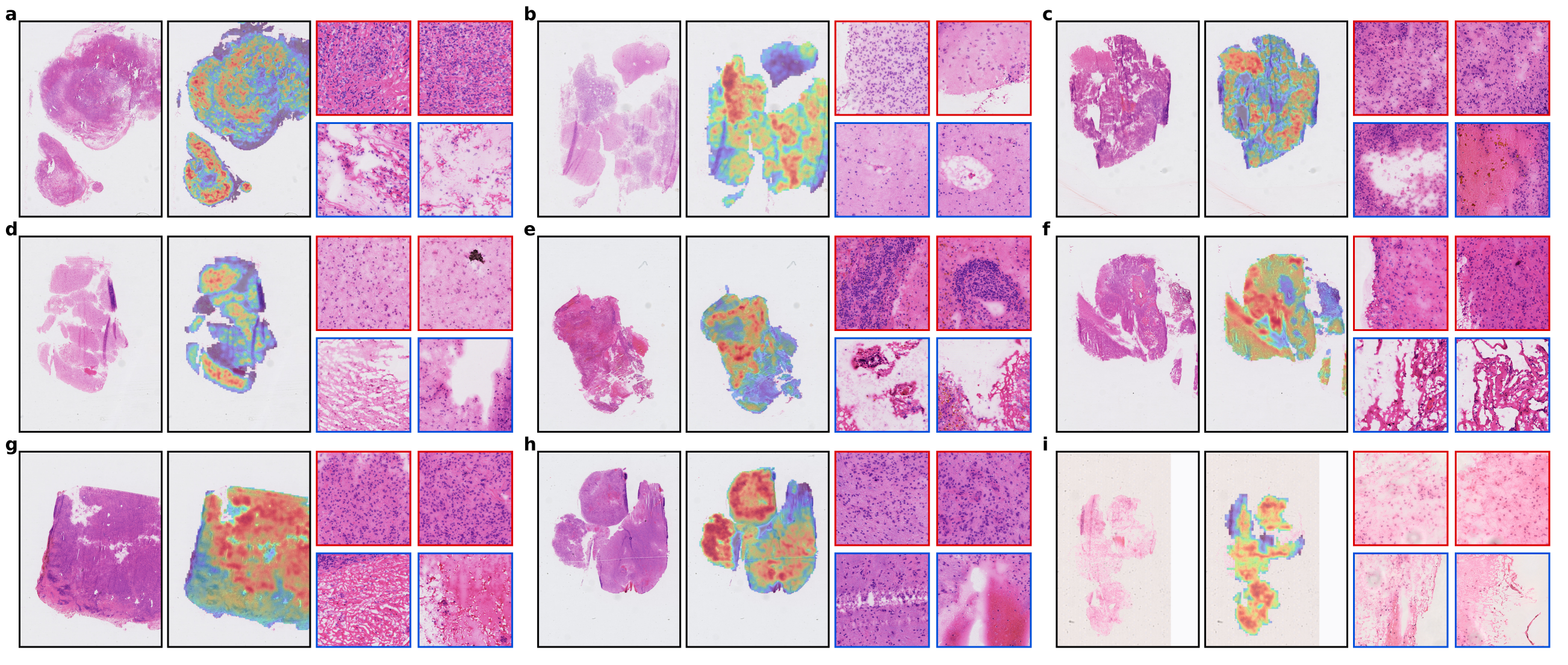}
	\caption{\textbf{Representative attention-map examples for diagnostic classification tasks.} Each subpanel shows a WSI thumbnail, the corresponding attention heatmap, two high-response patches (red frames), and two low-response patches (blue frames). Panel-task mapping: \textbf{a,} neuroepithelial versus non-neuroepithelial tumour classification; \textbf{b,} glioma or glioneuronal tumour versus embryonal tumour versus ependymoma classification; \textbf{c,} glioma versus glioneuronal tumour classification; \textbf{d,} meningioma versus other non-neuroepithelial tumours classification; \textbf{e,} PitNET versus SFT/mesenchymal tumour versus lymphoma versus craniopharyngioma versus schwannoma classification; \textbf{f,} diffuse versus circumscribed glioma classification; \textbf{g,} diffuse glioma molecular-subtype classification; \textbf{h,} pilocytic astrocytoma versus pleomorphic xanthoastrocytoma classification; \textbf{i,} lymphoma differentiation.}
	\label{fig:histology_attention_examples}
\end{figure}

\begin{figure}[htbp]
	\centering
	\includegraphics[width=\linewidth]{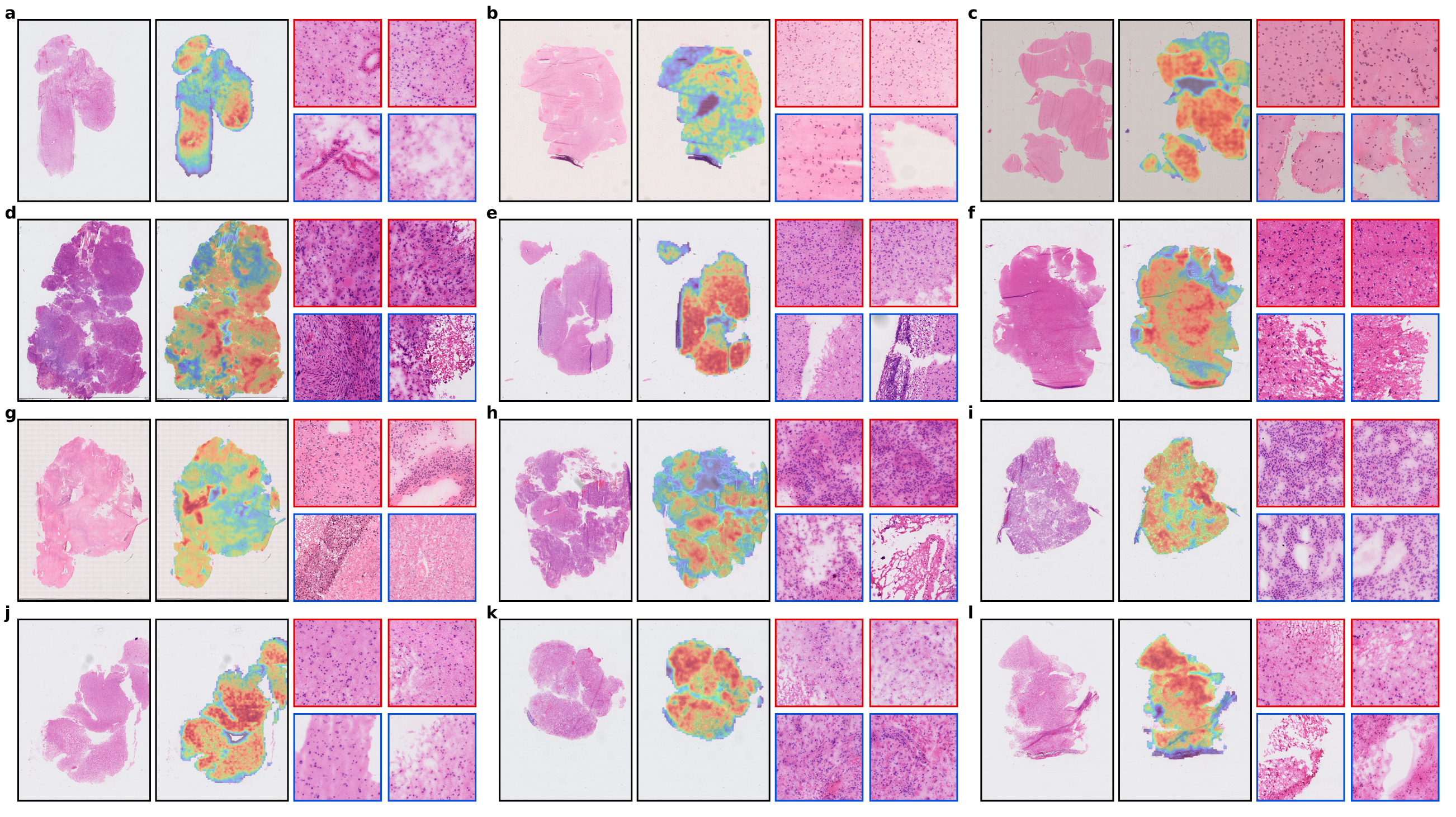}
	\caption{\textbf{Representative attention-map examples for biomarker prediction tasks.} Each subpanel shows a WSI thumbnail, the corresponding attention heatmap, two high-response patches (red frames), and two low-response patches (blue frames). Panel-task mapping: \textbf{a,} ATRX alteration prediction; \textbf{b,} 1p/19q codeletion prediction; \textbf{c,} H3 K27 alteration prediction; \textbf{d,} H3 G34 mutation prediction; \textbf{e,} BRAF mutation prediction; \textbf{f,} CDKN2A/B deletion prediction; \textbf{g,} TP53 mutation prediction; \textbf{h,} IDH1/IDH2 mutation prediction; \textbf{i,} MGMT promoter methylation prediction; \textbf{j,} chromosome 7 gain/chromosome 10 loss (+7/-10) prediction; \textbf{k,} EGFR amplification prediction; \textbf{l,} TERT promoter mutation prediction.}
	\label{fig:molecular_attention_examples}
\end{figure}

\clearpage

\begin{table}[htbp]
\centering
\caption{Distribution of pretraining slides by data source.}
\label{tabs_dist_pretrain_source}
\SuppTableSimple
\begin{tabular}{lrr}
\toprule
Data source & Slides & Percentage (\%) \\
\midrule
TCGA & 13,807 & 49.43 \\
Ivy-GAP & 11,230 & 40.21 \\
HEST & 1,228 & 4.40 \\
CGCI & 763 & 2.73 \\
HCMI-CMDC & 439 & 1.57 \\
CPTAC & 414 & 1.48 \\
CDDP\_EAGLE & 49 & 0.18 \\
\midrule
Total & 27,930 & 100.00 \\
\bottomrule
\end{tabular}
\end{table}

\begin{table}[htbp]
\centering
\caption{Distribution of pretraining slides by organ.}
\label{tabs_dist_pretrain_organ}
\SuppTableSimple
\begin{tabular}{lrr}
\toprule
Organ & Slides & Percentage (\%) \\
\midrule
Brain & 13,689 & 49.01 \\
Kidney & 1,725 & 6.18 \\
Bronchus and lungs & 1,672 & 5.99 \\
Breast & 1,525 & 5.46 \\
Bowel & 1,159 & 4.15 \\
Ovary & 1,048 & 3.75 \\
Uterus & 912 & 3.27 \\
Blood & 727 & 2.60 \\
Skin & 612 & 2.19 \\
Prostate & 568 & 2.03 \\
Head and Neck & 558 & 2.00 \\
Stomach & 542 & 1.94 \\
Thyroid & 447 & 1.60 \\
Liver & 388 & 1.39 \\
Bladder & 346 & 1.24 \\
Spinal cord & 318 & 1.14 \\
Connective tissue & 304 & 1.09 \\
Pancreas & 295 & 1.06 \\
Adrenal gland & 285 & 1.02 \\
Gullet & 247 & 0.88 \\
Thymus & 137 & 0.49 \\
Pleura & 88 & 0.32 \\
Bile duct & 84 & 0.30 \\
Eye and adnexa & 80 & 0.29 \\
Heart & 70 & 0.25 \\
Lymph node & 69 & 0.25 \\
Other & 35 & 0.13 \\
\midrule
Total & 27,930 & 100.00 \\
\bottomrule
\end{tabular}
\end{table}

\begin{landscape}
\begin{table}[p]
\centering
\small
\setlength{\tabcolsep}{1.5pt}
\caption{Downstream task partitions and label distributions in the development cohort and two independent validation cohorts.}
\label{tabs_downstream_label_dist}

\begin{tabular}{>{\raggedright\arraybackslash}p{0.33\linewidth}rrrr}
\toprule
Tasks & Training cohort & Test cohort & Temporal cohort & External cohort\\
\midrule

Non-neoplastic brain tissue/tumour
& 2777 (200/2577) & 694 (54/640) & 309 (29/280) & 246 (9/237)\\

Metastasis/\allowbreak \newline other CNS tumours
& 2577 (112/2465) & 640 (28/612) & 280 (14/266) & 237 (13/224)\\

Neuroepithelial/\allowbreak \newline non-neuroepithelial tumour
& 2291 (1876/415) & 576 (472/104) & 255 (202/53) & 221 (58/163)\\

Glioma or GNT/\allowbreak embryonal tumour/\allowbreak \newline ependymoma
& 1876 (1700/90/86) & 472 (427/25/20) & 202 (176/17/9) & 58 (52/0/6)\\

Glioma/GNT
& 1700 (1553/147) & 427 (390/37) & 176 (162/14) & 52 (50/2)\\

Meningioma/\allowbreak \newline other non-neuroepithelial tumours
& 415 (246/169) & 104 (62/42) & 53 (35/18) & 163 (72/91)\\

PitNET/\allowbreak SFT or mesenchymal tumour/\allowbreak \newline 
lymphoma/\allowbreak craniopharyngioma/\allowbreak schwannoma
& 169 (18/30/38/30/53) & 42 (5/8/10/6/13) & 16 (2/5/2/4/3) & 91 (51/4/4/8/24)\\

Diffuse/circumscribed glioma
& 1529 (1308/221) & 382 (327/55) & 162 (148/14) & 50 (44/6)\\

Astrocytoma /\allowbreak  Oligodendroglioma/\allowbreak \newline 
GBM /\allowbreak  DHG /\allowbreak  DMG
& 1308 (354/323/507/21/103) & 327 (88/81/127/5/26) & 124 (34/35/46/2/7) & 30 (6/3/21/0/0)\\

PA/PXA
& 221 (171/50) & 55 (43/12) & 12 (8/4) & 6 (5/1)\\

Glioma or metastasis or embryonal tumour /\allowbreak \newline  Lymphoma &
1793 (1755/38) & 453 (443/10) & 198 (196/2) & 95 (89/6)\\

ATRX unaltered/altered
& 1399 (1094/305) & 349 (273/76) & 196 (158/38) & 19 (16/3)\\

1p/19q non-codeleted/codeleted
& 1398 (1072/326) & 350 (268/82) & 196 (157/39) & 21 (18/3)\\

H3 K27 unaltered/altered
& 1384 (1317/67) & 346 (329/17) & 196 (189/7) & 19 (19/0)\\

H3 G34 wt/mut
& 1398 (1384/14) & 350 (346/4) & 196 (194/2) & 19 (19/0)\\

BRAF wt/mut
& 1399 (1338/61) & 349 (334/15) & 196 (188/8) & 21 (20/1)\\

CDKN2A/B retained/deleted
& 1398 (1047/351) & 350 (261/89) & 196 (147/49) & 19 (16/3)\\

TP53 wt/mut
& 1398 (867/531) & 350 (217/133) & 196 (124/72) & 19 (11/8)\\

IDH1/IDH2 wt/mut
& 1287 (668/619) & 322 (166/156) & 196 (119/77) & 21 (14/7)\\

MGMT promoter\newline unmethylated/\allowbreak methylated
& 1099 (485/614) & 275 (121/154) & 119 (42/77) & 21 (8/13)\\

+7/-10 negative/positive
& 1288 (1106/182) & 321 (266/55) & 196 (174/22) & 19 (14/5)\\

EGFR non-amplified/amplified
& 1288 (1099/189) & 321 (274/47) & 196 (164/32) & 19 (17/2)\\

TERT promoter wt/mut
& 1287 (648/639) & 322 (163/159) & 196 (110/86) & 21 (10/11)\\

\bottomrule
\end{tabular}

\vspace{2pt}
\begin{minipage}{\linewidth}
\footnotesize
Abbreviations: CNS, central nervous system; GNT, glioneuronal tumour; SFT, solitary fibrous tumour; GBM, glioblastoma; DHG, diffuse hemispheric glioma; DMG, diffuse midline glioma; PA, pilocytic astrocytoma; PXA, pleomorphic xanthoastrocytoma; wt, wildtype; mut, mutant; +7/-10, chromosome 7 gain/chromosome 10 loss. Numbers in parentheses indicate the class distribution in the order in which the corresponding labels are listed.
\end{minipage}
\end{table}
\end{landscape}

\begin{table}[htbp]
\centering
\caption{Distribution of downstream tasks slides by data source.}
\label{tabs_dist_tcga_source}
\SuppTableSimple
\begin{tabular}{lrr}
\toprule
Data source & Slides & Percentage (\%) \\
\midrule
TCGA & 6,336 & 53.89 \\
CPTAC & 5,049 & 42.94 \\
HCMI-CMDC & 165 & 1.40 \\
CGCI & 208 & 1.77 \\
\midrule
Total & 11758 & 100.00 \\
\bottomrule
\end{tabular}
\end{table}

\begin{table}[htbp]
\centering
\begin{threeparttable}
\caption{Summary of TCGA downstream tasks and label distributions.}
\label{tabs_tcga_label_dist}
\SuppTableSimple
\begin{tabular}{lrrr}
\hline
Task & Samples & Label distribution & External validation \\
\hline
BLCA (Benign vs. malignant) & 469 & 37 / 432 & -- \\
BRCA (Benign vs. malignant) & 1976 & 398 / 1578 & -- \\
CESC (tumour grading) & 283 & 19 / 139 / 125 & 208 (10 / 102 / 96) \\
COADREAD (Benign vs. malignant) & 1347 & 134 / 1213 & 165 (33 / 132) \\
HNSC (Benign vs. malignant) & 791 & 57 / 734 & 263 (92 / 171) \\
LIHC (tumour grading) & 396 & 60 / 189 / 134 / 13 & -- \\
LIHC (Benign vs. malignant) & 491 & 89 / 402 & -- \\
OV (Benign vs. malignant) & 1373 & 163 / 1210 & -- \\
PAAD (Benign vs. malignant) & 257 & 37 / 220 & 393 (127 / 266) \\
PRAD (Gleason grade group) & 605 & 53 / 190 / 127 / 72 / 163 & -- \\
PRAD (Benign vs. malignant) & 723 & 118 / 605 & -- \\
STAD (Benign vs. malignant) & 755 & 123 / 632 & -- \\
THCA (Benign vs. malignant) & 639 & 97 / 542 & -- \\
UCEC/UCS (Benign vs. malignant) & 868 & 60 / 808 & -- \\
LUAD/LUSC (Histological subtype) & 1576 & 823 / 753 & 1093 (800 / 293) \\
LUNG (Benign vs. malignant) & 2167 & 591 / 1576 & 1093 (428 / 665) \\
KIRC/KIRP/KICH (Histological subtype) & 1616 & 1090 / 526 & 460 (420 / 40) \\
KIRC/KIRP/KICH (tumour grading) & 1074 & 26 / 476 / 414 / 158 & 420 (28 / 211 / 143 / 38) \\
KIRC/KIRP/KICH (Benign vs. malignant) & 2332 & 716 / 1616 & 680 (220 / 460) \\
\hline
\end{tabular}
\begin{tablenotes}[flushleft]
\footnotesize
\item Abbreviations: BLCA, bladder urothelial carcinoma; BRCA, breast invasive carcinoma; CESC, cervical squamous cell carcinoma and endocervical adenocarcinoma; COADREAD, colon and rectal adenocarcinoma; HNSC, head and neck squamous cell carcinoma; LIHC, liver hepatocellular carcinoma; OV, ovarian serous cystadenocarcinoma; PAAD, pancreatic adenocarcinoma; PRAD, prostate adenocarcinoma; STAD, stomach adenocarcinoma; THCA, thyroid carcinoma; UCEC, uterine corpus endometrial carcinoma; UCS, uterine carcinosarcoma; LUAD, lung adenocarcinoma; LUSC, lung squamous cell carcinoma; LUNG, lung carcinoma; KIRC, kidney renal clear cell carcinoma; KIRP, kidney renal papillary cell carcinoma; KICH, kidney chromophobe.
\end{tablenotes}
\end{threeparttable}
\end{table}

\begin{table}[t]
\centering
\caption{Detailed configuration for self-supervised pretraining.}
\label{tabs_config_pretrain}
\SuppTableSimple
\begin{tabular}{p{0.38\linewidth}|p{0.22\linewidth}}
\toprule
\textbf{Hyper-parameter} & \textbf{Value} \\
\midrule
Backbone architecture & ViT-G/14 \\
Levels & 40 \\
Heads & 24 \\
Patch size & 14 \\
FFN layer & SwiGLU \\
Embedding dimension & 1536 \\
Register tokens & 4 \\
Drop path rate & 0.4 \\
\midrule
Global crop scale & 0.9, 1.0 \\
Global crop number \& size & 2, 224 \\
Local crop scale & 0.32, 0.5 \\
Local crop number \& size & 8, 98 \\
Max masking ratio & 0.5 \\
Min masking ratio & 0.1 \\
Gradient clipping max norm & 3.0 \\
Automatic mixed precision & \checkmark \\
\midrule
LoRA rank $r$ & 8 \\
LoRA scaling factor $\alpha$ & 16 \\
LoRA dropout & 0.0 \\
LoRA target module & \texttt{attn.qkv} \\
\midrule
AdamW $\beta$ & (0.9, 0.999) \\
Total batch size & 128 \\
Training iterations & 250,000 \\
Warmup iterations & 25,000 \\
Learning rate schedule & Cosine \\
Learning rate (start) & 0 \\
Learning rate (base) & $5 \times 10^{-5}$ \\
Learning rate (final) & $1 \times 10^{-6}$ \\
Teacher temperature (start) & 0.04 \\
Teacher temperature (final) & 0.07 \\
Teacher momentum (start) & 0.992 \\
Teacher momentum (final) & 1.000 \\
Weight decay (start) & 0.04 \\
Weight decay (end) & 0.4 \\
\bottomrule
\end{tabular}
\end{table}

\begingroup 
\renewcommand{\arraystretch}{1.18}
\begin{table}[htbp]
\centering
\caption{Definitions of the 23 diagnostic and molecular classification tasks.}
\label{tab:task_definitions}
\fontsize{9.0}{10.8}\selectfont
\setlength{\tabcolsep}{3.0pt}
\begin{tabular}{@{}l>{\raggedright\arraybackslash}p{\dimexpr\linewidth-4.5em\relax}@{}}
\hline
Task & Definition \\
\hline
\textbf{T1} & Non-neoplastic brain tissue versus tumour classification \\
\textbf{T2} & Metastasis versus other CNS tumours classification \\
\textbf{T3} & Neuroepithelial versus non-neuroepithelial tumour classification \\
\textbf{T4} & Glioma or glioneuronal tumour versus embryonal tumour versus ependymoma classification \\
\textbf{T5} & Glioma versus glioneuronal tumour classification \\
\textbf{T6} & Meningioma versus other non-neuroepithelial tumours classification \\
\textbf{T7} & Non-neuroepithelial tumour classification \\
\textbf{T8} & Diffuse versus circumscribed glioma classification \\
\textbf{T9} & Diffuse glioma molecular-subtype classification \\
\textbf{T10} & Pilocytic astrocytoma versus pleomorphic xanthoastrocytoma classification \\
\textbf{T11} & Lymphoma differentiation \\
\textbf{T12} & ATRX alteration prediction \\
\textbf{T13} & 1p/19q codeletion prediction \\
\textbf{T14} & H3 K27 alteration prediction \\
\textbf{T15} & H3 G34 mutation prediction \\
\textbf{T16} & BRAF mutation prediction \\
\textbf{T17} & CDKN2A/B deletion prediction \\
\textbf{T18} & TP53 mutation prediction \\
\textbf{T19} & IDH1/IDH2 mutation prediction \\
\textbf{T20} & MGMT promoter methylation prediction \\
\textbf{T21} & Chromosome 7 gain/chromosome 10 loss prediction \\
\textbf{T22} & EGFR amplification prediction \\
\textbf{T23} & TERT promoter mutation prediction \\
\hline
\end{tabular}
\end{table}

\endgroup 
\begingroup
\renewcommand{\arraystretch}{1.18}
\begin{table}[htbp]
\centering
\caption{Internal validation set AUC and 95\% confidence intervals for FOCUS across MIL methods.}
\label{tab:focus_val_auc_ci}
\SuppTableStandard
\setlength{\tabcolsep}{2pt}
\begin{tabular}{@{}lcccc@{}}
\hline
Task & DSMIL & ABMIL & CLAM & TransMIL \\
\hline
T1 & \textbf{0.991 (0.984--0.998)} & 0.990 (0.982--0.998) & \underline{0.991 (0.983--0.998)} & 0.989 (0.981--0.997) \\
T2 & \underline{0.992 (0.986--0.999)} & 0.976 (0.954--0.997) & 0.990 (0.981--0.998) & \textbf{0.994 (0.989--0.999)} \\
T3 & \textbf{0.999 (0.998--1.000)} & 0.996 (0.991--1.000) & 0.997 (0.993--1.000) & \underline{0.999 (0.998--1.000)} \\
T4 & 0.974 (0.929--0.997) & 0.975 (0.941--0.995) & \underline{0.979 (0.948--0.996)} & \textbf{0.985 (0.960--0.999)} \\
T5 & 0.946 (0.915--0.976) & \underline{0.947 (0.918--0.977)} & \textbf{0.952 (0.923--0.981)} & 0.946 (0.914--0.978) \\
T6 & \textbf{0.998 (0.994--1.000)} & 0.995 (0.984--1.000) & 0.994 (0.982--1.000) & \underline{0.996 (0.989--1.000)} \\
T7 & \textbf{1.000 (1.000--1.000)} & \textbf{1.000 (1.000--1.000)} & \textbf{1.000 (1.000--1.000)} & \textbf{1.000 (1.000--1.000)} \\
T8 & 0.976 (0.959--0.992) & 0.978 (0.961--0.994) & \underline{0.981 (0.968--0.995)} & \textbf{0.982 (0.970--0.994)} \\
T9 & 0.959 (0.939--0.975) & \textbf{0.961 (0.943--0.975)} & \underline{0.961 (0.942--0.976)} & 0.956 (0.936--0.972) \\
T10 & 0.989 (0.967--1.000) & 0.986 (0.961--1.000) & \textbf{0.994 (0.981--1.000)} & \underline{0.991 (0.972--1.000)} \\
T11 & 0.994 (0.985--1.000) & \underline{0.997 (0.992--1.000)} & \textbf{0.999 (0.996--1.000)} & 0.995 (0.989--1.000) \\
T12 & \textbf{0.781 (0.710--0.851)} & 0.777 (0.703--0.851) & \underline{0.779 (0.706--0.852)} & 0.751 (0.677--0.826) \\
T13 & 0.971 (0.951--0.990) & \textbf{0.976 (0.958--0.995)} & \underline{0.972 (0.953--0.992)} & 0.964 (0.942--0.986) \\
T14 & 0.882 (0.783--0.982) & \underline{0.896 (0.817--0.974)} & 0.892 (0.803--0.981) & \textbf{0.897 (0.788--1.000)} \\
T15 & \underline{0.815 (0.582--1.000)} & 0.757 (0.473--1.000) & 0.779 (0.498--1.000) & \textbf{0.830 (0.551--1.000)} \\
T16 & 0.938 (0.901--0.975) & 0.940 (0.903--0.977) & \textbf{0.950 (0.914--0.986)} & \underline{0.945 (0.901--0.989)} \\
T17 & \underline{0.818 (0.760--0.877)} & \textbf{0.822 (0.765--0.879)} & 0.816 (0.758--0.873) & 0.808 (0.748--0.868) \\
T18 & 0.871 (0.829--0.912) & \textbf{0.888 (0.849--0.926)} & \underline{0.876 (0.835--0.917)} & 0.873 (0.833--0.913) \\
T19 & 0.960 (0.937--0.982) & \underline{0.960 (0.937--0.984)} & \textbf{0.962 (0.939--0.984)} & 0.949 (0.923--0.974) \\
T20 & \textbf{0.747 (0.682--0.811)} & 0.735 (0.669--0.800) & 0.740 (0.675--0.805) & \underline{0.744 (0.678--0.809)} \\
T21 & 0.830 (0.761--0.898) & \underline{0.830 (0.760--0.901)} & 0.829 (0.757--0.900) & \textbf{0.835 (0.768--0.902)} \\
T22 & \textbf{0.871 (0.807--0.934)} & 0.860 (0.794--0.927) & 0.862 (0.795--0.928) & \underline{0.863 (0.802--0.924)} \\
T23 & \textbf{0.877 (0.836--0.918)} & 0.870 (0.827--0.912) & \underline{0.873 (0.830--0.916)} & 0.856 (0.812--0.901) \\
\hline
Mean & \textbf{0.921 (0.888--0.951)} & 0.918 (0.883--0.949) & \underline{0.920 (0.886--0.951)} & 0.919 (0.885--0.950) \\
\hline
\end{tabular}
\par\smallskip
\noindent\parbox{\textwidth}{\SuppTableNotes Bold indicates the highest AUC within a task and in the mean row. Underlining indicates the second-highest AUC within a task and in the mean row.}
\end{table}
\endgroup

\begingroup
\renewcommand{\arraystretch}{1.18}
\begin{table}[htbp]
\centering
\caption{Internal test set AUC and 95\% confidence intervals for FOCUS across MIL methods.}
\label{tab:focus_test_auc_ci}
\SuppTableStandard
\setlength{\tabcolsep}{2pt}
\begin{tabular}{@{}lcccc@{}}
\hline
Task & DSMIL & ABMIL & CLAM & TransMIL \\
\hline
T1 & 0.964 (0.943--0.985) & \textbf{0.974 (0.960--0.989)} & \underline{0.970 (0.954--0.986)} & 0.968 (0.951--0.986) \\
T2 & \textbf{0.995 (0.991--1.000)} & 0.983 (0.964--1.000) & 0.993 (0.985--1.000) & \underline{0.995 (0.989--1.000)} \\
T3 & \textbf{0.992 (0.986--0.998)} & 0.991 (0.984--0.999) & \underline{0.992 (0.984--0.999)} & 0.988 (0.979--0.997) \\
T4 & \textbf{0.987 (0.977--0.995)} & 0.985 (0.971--0.995) & \underline{0.985 (0.973--0.995)} & 0.985 (0.971--0.995) \\
T5 & 0.898 (0.861--0.936) & 0.902 (0.865--0.939) & \textbf{0.917 (0.883--0.950)} & \underline{0.908 (0.869--0.947)} \\
T6 & \textbf{0.992 (0.979--1.000)} & \underline{0.988 (0.972--1.000)} & 0.987 (0.970--1.000) & 0.987 (0.973--1.000) \\
T7 & 0.977 (0.944--0.997) & \underline{0.984 (0.958--0.999)} & \textbf{0.988 (0.964--1.000)} & 0.970 (0.926--0.997) \\
T8 & 0.956 (0.933--0.979) & 0.964 (0.946--0.981) & \textbf{0.969 (0.953--0.984)} & \underline{0.967 (0.950--0.984)} \\
T9 & \textbf{0.932 (0.912--0.950)} & 0.928 (0.905--0.947) & \underline{0.932 (0.910--0.950)} & 0.928 (0.905--0.948) \\
T10 & \underline{0.988 (0.965--1.000)} & 0.986 (0.962--1.000) & \underline{0.988 (0.965--1.000)} & \textbf{1.000 (1.000--1.000)} \\
T11 & \textbf{0.973 (0.937--1.000)} & \underline{0.972 (0.935--1.000)} & 0.970 (0.928--1.000) & 0.952 (0.886--1.000) \\
T12 & 0.834 (0.779--0.890) & \underline{0.835 (0.780--0.891)} & \textbf{0.837 (0.782--0.891)} & 0.812 (0.757--0.868) \\
T13 & 0.967 (0.951--0.984) & \textbf{0.971 (0.955--0.987)} & \underline{0.970 (0.954--0.986)} & 0.961 (0.941--0.981) \\
T14 & 0.905 (0.858--0.951) & 0.892 (0.835--0.949) & \underline{0.906 (0.856--0.956)} & \textbf{0.910 (0.853--0.967)} \\
T15 & \underline{0.981 (0.962--1.000)} & 0.968 (0.936--1.000) & 0.966 (0.924--1.000) & \textbf{0.986 (0.973--1.000)} \\
T16 & 0.836 (0.722--0.950) & 0.852 (0.742--0.963) & \underline{0.859 (0.766--0.952)} & \textbf{0.874 (0.794--0.954)} \\
T17 & \textbf{0.823 (0.774--0.872)} & \underline{0.821 (0.770--0.871)} & 0.819 (0.768--0.870) & 0.813 (0.762--0.864) \\
T18 & \textbf{0.866 (0.826--0.907)} & \underline{0.859 (0.818--0.900)} & 0.854 (0.812--0.896) & 0.850 (0.806--0.895) \\
T19 & \underline{0.955 (0.935--0.975)} & 0.949 (0.926--0.972) & 0.949 (0.927--0.972) & \textbf{0.962 (0.943--0.980)} \\
T20 & 0.736 (0.677--0.794) & \textbf{0.744 (0.687--0.801)} & \underline{0.738 (0.681--0.796)} & 0.699 (0.637--0.761) \\
T21 & 0.884 (0.846--0.923) & \underline{0.902 (0.868--0.935)} & \textbf{0.904 (0.871--0.937)} & 0.866 (0.823--0.910) \\
T22 & 0.855 (0.790--0.920) & \underline{0.862 (0.801--0.922)} & 0.853 (0.789--0.918) & \textbf{0.864 (0.804--0.923)} \\
T23 & 0.877 (0.840--0.915) & 0.876 (0.839--0.913) & \underline{0.880 (0.843--0.917)} & \textbf{0.891 (0.855--0.926)} \\
\hline
Mean & 0.921 (0.892--0.948) & \underline{0.921 (0.894--0.947)} & \textbf{0.923 (0.894--0.949)} & 0.919 (0.887--0.947) \\
\hline
\end{tabular}
\par\smallskip
\noindent\parbox{\textwidth}{\SuppTableNotes Bold indicates the highest AUC within a task and in the mean row. Underlining indicates the second-highest AUC within a task and in the mean row.}
\end{table}
\endgroup


\begingroup
\renewcommand{\arraystretch}{1.16}
\begin{table}[p]
\centering
\caption{Performance of various models on internal test set for T1: Normal brain (0) vs tumour (1).}
\label{tab:dsmil_macro_metrics_task1}
\SuppTableStandard
\setlength{\tabcolsep}{4pt}
\begin{tabular}{@{}lccccc@{}}
\hline
Model & AUC & Accuracy & Sensitivity & Specificity & F1 score \\
\hline
\textbf{FOCUS} & \textbf{0.964 (0.943--0.985)} & 0.947 (0.928--0.963) & 0.963 (0.946--0.976) & 0.759 (0.635--0.878) & 0.971 (0.960--0.980) \\
H-Optimus-1 & \underline{0.959 (0.930--0.988)} & 0.935 (0.915--0.954) & 0.941 (0.921--0.959) & \textbf{0.870 (0.771--0.957)} & 0.964 (0.952--0.974) \\
H-Optimus-0 & 0.954 (0.923--0.986) & 0.937 (0.916--0.954) & 0.942 (0.922--0.960) & \textbf{0.870 (0.771--0.957)} & 0.965 (0.953--0.975) \\
GigaPath & 0.958 (0.931--0.984) & 0.935 (0.915--0.952) & 0.944 (0.923--0.961) & \underline{0.833 (0.727--0.929)} & 0.964 (0.952--0.974) \\
CRISP & 0.957 (0.929--0.985) & 0.935 (0.915--0.952) & 0.947 (0.927--0.964) & 0.796 (0.678--0.900) & 0.964 (0.953--0.974) \\
Virchow2 & 0.958 (0.927--0.989) & \textbf{0.954 (0.937--0.968)} & \textbf{0.969 (0.954--0.981)} & 0.778 (0.653--0.889) & \textbf{0.975 (0.965--0.983)} \\
Virchow & 0.952 (0.919--0.984) & \underline{0.952 (0.935--0.967)} & \underline{0.967 (0.952--0.980)} & 0.778 (0.651--0.882) & \underline{0.974 (0.964--0.982)} \\
UNI & \underline{0.959 (0.933--0.985)} & 0.945 (0.927--0.963) & 0.963 (0.946--0.977) & 0.741 (0.617--0.857) & 0.970 (0.960--0.979) \\
Phikon-v2 & 0.955 (0.926--0.985) & 0.948 (0.931--0.964) & 0.963 (0.947--0.976) & 0.778 (0.661--0.887) & 0.972 (0.962--0.980) \\
\hline
\end{tabular}
\par\smallskip
\noindent\parbox{\textwidth}{\SuppTableNotes Values are point estimate (95\% CI). Binary-task CIs use DeLong’s method; multiclass AUC and the other metrics use 2,000-patient nonparametric bootstrap percentile intervals. Bold indicates the highest value for each metric within a task; underlining indicates the second-highest value.}
\end{table}

\begin{table}[p]
\centering
\caption{Performance of various models on internal test set for T2: Metastasis (0) vs all other tumours (1).}
\label{tab:dsmil_macro_metrics_task2}
\SuppTableStandard
\setlength{\tabcolsep}{4pt}
\begin{tabular}{@{}lccccc@{}}
\hline
Model & AUC & Accuracy & Sensitivity & Specificity & F1 score \\
\hline
\textbf{FOCUS} & \textbf{0.995 (0.991--1.000)} & \textbf{0.984 (0.975--0.994)} & \underline{0.997 (0.992--1.000)} & \textbf{0.714 (0.520--0.879)} & \textbf{0.992 (0.987--0.997)} \\
H-Optimus-1 & \underline{0.992 (0.985--0.999)} & \underline{0.983 (0.972--0.992)} & 0.995 (0.989--1.000) & \textbf{0.714 (0.538--0.875)} & 0.991 (0.985--0.996) \\
H-Optimus-0 & 0.989 (0.980--0.998) & 0.981 (0.970--0.991) & 0.995 (0.989--1.000) & \underline{0.679 (0.500--0.850)} & 0.990 (0.984--0.995) \\
GigaPath & 0.981 (0.968--0.994) & 0.973 (0.961--0.984) & 0.995 (0.989--1.000) & 0.500 (0.308--0.690) & 0.986 (0.980--0.992) \\
CRISP & 0.986 (0.973--0.998) & 0.980 (0.969--0.989) & \textbf{0.998 (0.995--1.000)} & 0.571 (0.385--0.750) & 0.989 (0.984--0.994) \\
Virchow2 & 0.980 (0.963--0.997) & \underline{0.983 (0.972--0.992)} & \textbf{0.998 (0.995--1.000)} & 0.643 (0.458--0.815) & \underline{0.991 (0.985--0.996)} \\
Virchow & 0.983 (0.967--0.999) & 0.981 (0.970--0.991) & 0.995 (0.989--1.000) & \underline{0.679 (0.500--0.857)} & 0.990 (0.985--0.995) \\
UNI & 0.985 (0.974--0.996) & 0.978 (0.967--0.989) & 0.995 (0.989--1.000) & 0.607 (0.417--0.778) & 0.989 (0.983--0.994) \\
Phikon-v2 & 0.978 (0.965--0.990) & 0.972 (0.959--0.984) & \underline{0.997 (0.992--1.000)} & 0.429 (0.241--0.613) & 0.985 (0.979--0.992) \\
\hline
\end{tabular}
\par\smallskip
\noindent\parbox{\textwidth}{\SuppTableNotes Values are point estimate (95\% CI). Binary-task CIs use DeLong’s method; multiclass AUC and the other metrics use 2,000-patient nonparametric bootstrap percentile intervals. Bold indicates the highest value for each metric within a task; underlining indicates the second-highest value.}
\end{table}

\begin{table}[p]
\centering
\caption{Performance of various models on internal test set for T3: Neuroepithelial (0) vs non-neuroepithelial tumours (1).}
\label{tab:dsmil_macro_metrics_task3}
\SuppTableStandard
\setlength{\tabcolsep}{4pt}
\begin{tabular}{@{}lccccc@{}}
\hline
Model & AUC & Accuracy & Sensitivity & Specificity & F1 score \\
\hline
\textbf{FOCUS} & \textbf{0.992 (0.986--0.998)} & \textbf{0.973 (0.959--0.988)} & \underline{0.923 (0.867--0.971)} & \textbf{0.985 (0.972--0.996)} & \underline{0.928 (0.888--0.964)} \\
H-Optimus-1 & \underline{0.989 (0.981--0.997)} & \textbf{0.973 (0.959--0.986)} & \textbf{0.933 (0.880--0.979)} & \underline{0.983 (0.969--0.993)} & \textbf{0.928 (0.889--0.963)} \\
H-Optimus-0 & 0.985 (0.974--0.996) & 0.968 (0.952--0.982) & 0.913 (0.854--0.967) & 0.980 (0.966--0.991) & 0.913 (0.868--0.951) \\
GigaPath & 0.985 (0.976--0.995) & 0.966 (0.950--0.980) & \underline{0.923 (0.867--0.971)} & 0.976 (0.960--0.989) & 0.910 (0.864--0.948) \\
CRISP & 0.985 (0.973--0.996) & \textbf{0.973 (0.959--0.986)} & \textbf{0.933 (0.880--0.980)} & \underline{0.983 (0.969--0.994)} & \textbf{0.928 (0.889--0.962)} \\
Virchow2 & 0.986 (0.973--0.998) & \underline{0.972 (0.957--0.984)} & \underline{0.923 (0.867--0.972)} & \underline{0.983 (0.970--0.993)} & 0.923 (0.882--0.958) \\
Virchow & 0.979 (0.962--0.996) & 0.964 (0.948--0.979) & 0.913 (0.857--0.967) & 0.976 (0.961--0.989) & 0.905 (0.856--0.945) \\
UNI & 0.986 (0.975--0.997) & 0.968 (0.954--0.982) & 0.894 (0.832--0.952) & \textbf{0.985 (0.972--0.996)} & 0.912 (0.869--0.950) \\
Phikon-v2 & 0.984 (0.974--0.995) & 0.959 (0.943--0.975) & 0.846 (0.776--0.914) & \textbf{0.985 (0.972--0.996)} & 0.884 (0.836--0.931) \\
\hline
\end{tabular}
\par\smallskip
\noindent\parbox{\textwidth}{\SuppTableNotes Values are point estimate (95\% CI). Binary-task CIs use DeLong’s method; multiclass AUC and the other metrics use 2,000-patient nonparametric bootstrap percentile intervals. Bold indicates the highest value for each metric within a task; underlining indicates the second-highest value.}
\end{table}

\begin{table}[p]
\centering
\caption{Performance of various models on internal test set for T4: Glioma or glioneuronal tumour (0) vs embryonal tumour (1) vs ependymoma (2).}
\label{tab:dsmil_macro_metrics_task4}
\SuppTableStandard
\setlength{\tabcolsep}{4pt}
\begin{tabular}{@{}lccccc@{}}
\hline
Model & AUC & Accuracy & Sensitivity & Specificity & F1 score \\
\hline
\textbf{FOCUS} & \textbf{0.987 (0.977--0.995)} & \textbf{0.974 (0.959--0.986)} & \textbf{0.972 (0.958--0.985)} & \textbf{0.779 (0.668--0.887)} & \textbf{0.971 (0.953--0.985)} \\
H-Optimus-1 & \underline{0.983 (0.972--0.993)} & 0.958 (0.938--0.974) & 0.956 (0.934--0.972) & 0.638 (0.502--0.764) & 0.951 (0.926--0.971) \\
H-Optimus-0 & 0.977 (0.958--0.991) & 0.964 (0.946--0.980) & 0.962 (0.943--0.979) & 0.678 (0.550--0.803) & 0.957 (0.933--0.976) \\
GigaPath & 0.977 (0.959--0.991) & \underline{0.966 (0.950--0.980)} & \underline{0.964 (0.947--0.979)} & \underline{0.718 (0.590--0.835)} & \underline{0.961 (0.940--0.979)} \\
CRISP & 0.974 (0.945--0.992) & 0.960 (0.942--0.976) & 0.958 (0.939--0.975) & 0.658 (0.529--0.782) & 0.954 (0.930--0.973) \\
Virchow2 & 0.977 (0.955--0.992) & 0.962 (0.944--0.978) & 0.960 (0.941--0.977) & 0.658 (0.535--0.783) & 0.953 (0.929--0.974) \\
Virchow & 0.968 (0.931--0.991) & 0.962 (0.944--0.978) & 0.960 (0.941--0.977) & 0.638 (0.506--0.765) & 0.953 (0.928--0.973) \\
UNI & 0.982 (0.969--0.993) & 0.962 (0.944--0.978) & 0.960 (0.941--0.977) & 0.658 (0.528--0.788) & 0.954 (0.931--0.975) \\
Phikon-v2 & 0.948 (0.918--0.974) & 0.946 (0.925--0.965) & 0.943 (0.922--0.964) & 0.497 (0.365--0.631) & 0.925 (0.895--0.953) \\
\hline
\end{tabular}
\par\smallskip
\noindent\parbox{\textwidth}{\SuppTableNotes Values are point estimate (95\% CI). Binary-task CIs use DeLong’s method; multiclass AUC and the other metrics use 2,000-patient nonparametric bootstrap percentile intervals. Bold indicates the highest value for each metric within a task; underlining indicates the second-highest value.}
\end{table}

\begin{table}[p]
\centering
\caption{Performance of various models on internal test set for T5: Glioma (0) vs glioneuronal tumour (1).}
\label{tab:dsmil_macro_metrics_task5}
\SuppTableStandard
\setlength{\tabcolsep}{4pt}
\begin{tabular}{@{}lccccc@{}}
\hline
Model & AUC & Accuracy & Sensitivity & Specificity & F1 score \\
\hline
\textbf{FOCUS} & \textbf{0.898 (0.861--0.936)} & 0.899 (0.871--0.925) & 0.378 (0.216--0.533) & 0.949 (0.926--0.969) & 0.394 (0.235--0.531) \\
H-Optimus-1 & 0.869 (0.822--0.917) & \textbf{0.909 (0.881--0.934)} & 0.216 (0.094--0.359) & \textbf{0.974 (0.957--0.990)} & 0.291 (0.130--0.453) \\
H-Optimus-0 & 0.855 (0.804--0.906) & \underline{0.902 (0.874--0.927)} & 0.162 (0.056--0.286) & \underline{0.972 (0.954--0.987)} & 0.222 (0.078--0.370) \\
GigaPath & 0.863 (0.816--0.911) & 0.888 (0.857--0.916) & 0.135 (0.029--0.256) & 0.959 (0.938--0.977) & 0.172 (0.038--0.306) \\
CRISP & 0.887 (0.845--0.928) & 0.892 (0.862--0.918) & \textbf{0.514 (0.345--0.677)} & 0.928 (0.901--0.951) & \textbf{0.452 (0.311--0.575)} \\
Virchow2 & 0.880 (0.836--0.924) & 0.892 (0.859--0.920) & 0.459 (0.294--0.621) & 0.933 (0.907--0.957) & 0.425 (0.276--0.556) \\
Virchow & 0.870 (0.825--0.914) & 0.888 (0.857--0.913) & 0.405 (0.250--0.564) & 0.933 (0.906--0.956) & 0.385 (0.234--0.511) \\
UNI & \underline{0.891 (0.853--0.930)} & 0.895 (0.864--0.920) & \underline{0.486 (0.318--0.644)} & 0.933 (0.907--0.956) & \underline{0.444 (0.296--0.571)} \\
Phikon-v2 & 0.799 (0.729--0.869) & 0.895 (0.867--0.920) & 0.054 (0.000--0.139) & \textbf{0.974 (0.957--0.990)} & 0.082 (0.000--0.200) \\
\hline
\end{tabular}
\par\smallskip
\noindent\parbox{\textwidth}{\SuppTableNotes Values are point estimate (95\% CI). Binary-task CIs use DeLong’s method; multiclass AUC and the other metrics use 2,000-patient nonparametric bootstrap percentile intervals. Bold indicates the highest value for each metric within a task; underlining indicates the second-highest value.}
\end{table}

\begin{table}[p]
\centering
\caption{Performance of various models on internal test set for T6: Meningioma (0) vs other non-neuroepithelial tumours (1).}
\label{tab:dsmil_macro_metrics_task6}
\SuppTableStandard
\setlength{\tabcolsep}{4pt}
\begin{tabular}{@{}lccccc@{}}
\hline
Model & AUC & Accuracy & Sensitivity & Specificity & F1 score \\
\hline
\textbf{FOCUS} & \textbf{0.992 (0.979--1.000)} & \textbf{0.971 (0.933--1.000)} & \textbf{0.976 (0.921--1.000)} & \underline{0.968 (0.919--1.000)} & \textbf{0.965 (0.919--1.000)} \\
H-Optimus-1 & 0.982 (0.961--1.000) & \underline{0.952 (0.904--0.990)} & 0.929 (0.844--1.000) & \underline{0.968 (0.919--1.000)} & 0.940 (0.880--0.987) \\
H-Optimus-0 & 0.974 (0.947--1.000) & 0.923 (0.865--0.971) & 0.905 (0.809--0.978) & 0.935 (0.866--0.985) & 0.905 (0.831--0.961) \\
GigaPath & 0.984 (0.967--1.000) & \underline{0.952 (0.904--0.990)} & \underline{0.952 (0.879--1.000)} & 0.952 (0.889--1.000) & \underline{0.941 (0.878--0.987)} \\
CRISP & \underline{0.988 (0.975--1.000)} & 0.923 (0.865--0.971) & 0.905 (0.809--0.978) & 0.935 (0.869--0.985) & 0.905 (0.831--0.963) \\
Virchow2 & 0.986 (0.971--1.000) & 0.942 (0.894--0.981) & 0.905 (0.806--0.977) & \underline{0.968 (0.919--1.000)} & 0.927 (0.857--0.978) \\
Virchow & 0.971 (0.946--0.997) & 0.913 (0.856--0.962) & 0.905 (0.806--0.977) & 0.919 (0.846--0.983) & 0.894 (0.816--0.957) \\
UNI & 0.986 (0.970--1.000) & 0.942 (0.894--0.981) & 0.929 (0.842--1.000) & 0.952 (0.891--1.000) & 0.929 (0.861--0.978) \\
Phikon-v2 & 0.967 (0.936--0.998) & 0.904 (0.846--0.952) & 0.786 (0.659--0.895) & \textbf{0.984 (0.948--1.000)} & 0.868 (0.783--0.936) \\
\hline
\end{tabular}
\par\smallskip
\noindent\parbox{\textwidth}{\SuppTableNotes Values are point estimate (95\% CI). Binary-task CIs use DeLong’s method; multiclass AUC and the other metrics use 2,000-patient nonparametric bootstrap percentile intervals. Bold indicates the highest value for each metric within a task; underlining indicates the second-highest value.}
\end{table}

\begin{table}[p]
\centering
\caption{Performance of various models on internal test set for T7: PitNET (0) vs SFT/mesenchymal tumour (1) vs lymphoma (2) vs craniopharyngioma (3) vs schwannoma (4).}
\label{tab:dsmil_macro_metrics_task7}
\SuppTableStandard
\setlength{\tabcolsep}{4pt}
\begin{tabular}{@{}lccccc@{}}
\hline
Model & AUC & Accuracy & Sensitivity & Specificity & F1 score \\
\hline
\textbf{FOCUS} & \textbf{0.977 (0.944--0.997)} & \underline{0.925 (0.867--0.972)} & \textbf{0.833 (0.714--0.929)} & 0.938 (0.884--0.979) & \underline{0.830 (0.707--0.929)} \\
H-Optimus-1 & 0.971 (0.937--0.993) & 0.900 (0.834--0.954) & 0.786 (0.667--0.905) & 0.914 (0.853--0.962) & 0.775 (0.627--0.899) \\
H-Optimus-0 & 0.958 (0.917--0.986) & 0.881 (0.815--0.935) & 0.738 (0.595--0.857) & 0.899 (0.833--0.948) & 0.729 (0.573--0.856) \\
GigaPath & 0.968 (0.927--0.994) & 0.921 (0.859--0.969) & \underline{0.810 (0.667--0.929)} & \underline{0.939 (0.885--0.980)} & 0.803 (0.655--0.923) \\
CRISP & \underline{0.976 (0.944--0.997)} & \textbf{0.927 (0.869--0.972)} & \textbf{0.833 (0.714--0.929)} & \textbf{0.956 (0.917--0.987)} & \textbf{0.836 (0.723--0.931)} \\
Virchow2 & 0.971 (0.930--0.995) & 0.894 (0.822--0.950) & 0.762 (0.619--0.881) & 0.933 (0.889--0.971) & 0.762 (0.625--0.881) \\
Virchow & 0.975 (0.936--0.997) & 0.889 (0.823--0.945) & 0.762 (0.643--0.881) & 0.910 (0.849--0.958) & 0.753 (0.608--0.877) \\
UNI & 0.958 (0.918--0.985) & 0.888 (0.824--0.945) & 0.738 (0.595--0.857) & 0.910 (0.860--0.955) & 0.715 (0.554--0.858) \\
Phikon-v2 & 0.911 (0.850--0.957) & 0.756 (0.693--0.810) & 0.476 (0.333--0.643) & 0.781 (0.686--0.852) & 0.364 (0.215--0.542) \\
\hline
\end{tabular}
\par\smallskip
\noindent\parbox{\textwidth}{\SuppTableNotes Values are point estimate (95\% CI). Binary-task CIs use DeLong’s method; multiclass AUC and the other metrics use 2,000-patient nonparametric bootstrap percentile intervals. Bold indicates the highest value for each metric within a task; underlining indicates the second-highest value.}
\end{table}

\begin{table}[p]
\centering
\caption{Performance of various models on internal test set for T8: Diffuse (0) vs circumscribed glioma (1).}
\label{tab:dsmil_macro_metrics_task8}
\SuppTableStandard
\setlength{\tabcolsep}{4pt}
\begin{tabular}{@{}lccccc@{}}
\hline
Model & AUC & Accuracy & Sensitivity & Specificity & F1 score \\
\hline
\textbf{FOCUS} & \textbf{0.956 (0.933--0.979)} & \textbf{0.921 (0.893--0.948)} & 0.600 (0.473--0.733) & \underline{0.976 (0.957--0.991)} & \underline{0.688 (0.575--0.787)} \\
H-Optimus-1 & 0.945 (0.916--0.974) & 0.906 (0.874--0.935) & 0.491 (0.360--0.623) & \underline{0.976 (0.959--0.991)} & 0.600 (0.474--0.713) \\
H-Optimus-0 & 0.939 (0.905--0.972) & 0.914 (0.882--0.942) & 0.527 (0.393--0.661) & \textbf{0.979 (0.963--0.994)} & 0.637 (0.511--0.741) \\
GigaPath & 0.948 (0.919--0.977) & 0.914 (0.885--0.942) & 0.600 (0.467--0.731) & 0.966 (0.945--0.985) & 0.667 (0.548--0.766) \\
CRISP & 0.939 (0.910--0.968) & 0.908 (0.880--0.937) & \textbf{0.655 (0.524--0.778)} & 0.951 (0.928--0.973) & 0.673 (0.560--0.768) \\
Virchow2 & 0.935 (0.902--0.967) & 0.908 (0.880--0.935) & \underline{0.618 (0.491--0.750)} & 0.957 (0.934--0.978) & 0.660 (0.548--0.759) \\
Virchow & 0.952 (0.929--0.975) & \underline{0.919 (0.893--0.945)} & \textbf{0.655 (0.525--0.776)} & 0.963 (0.943--0.982) & \textbf{0.699 (0.588--0.792)} \\
UNI & \underline{0.955 (0.931--0.978)} & 0.914 (0.882--0.942) & 0.582 (0.450--0.714) & 0.969 (0.950--0.987) & 0.660 (0.545--0.761) \\
Phikon-v2 & 0.894 (0.853--0.934) & 0.895 (0.864--0.924) & 0.545 (0.420--0.680) & 0.954 (0.931--0.976) & 0.600 (0.485--0.709) \\
\hline
\end{tabular}
\par\smallskip
\noindent\parbox{\textwidth}{\SuppTableNotes Values are point estimate (95\% CI). Binary-task CIs use DeLong’s method; multiclass AUC and the other metrics use 2,000-patient nonparametric bootstrap percentile intervals. Bold indicates the highest value for each metric within a task; underlining indicates the second-highest value.}
\end{table}

\begin{table}[p]
\centering
\caption{Performance of various models on internal test set for T9: IDH-mutant astrocytoma (0), IDH-mutant and 1p/19q-codeleted oligodendroglioma (1), IDH-wild-type GBM (2), H3G34-mutant DHG (3), or H3K27-altered DMG (4).}
\label{tab:dsmil_macro_metrics_task9}
\SuppTableStandard
\setlength{\tabcolsep}{4pt}
\begin{tabular}{@{}lccccc@{}}
\hline
Model & AUC & Accuracy & Sensitivity & Specificity & F1 score \\
\hline
\textbf{FOCUS} & \textbf{0.932 (0.912--0.950)} & \underline{0.869 (0.843--0.893)} & \textbf{0.761 (0.716--0.804)} & \underline{0.884 (0.860--0.906)} & \textbf{0.750 (0.699--0.796)} \\
H-Optimus-1 & 0.911 (0.889--0.932) & 0.856 (0.830--0.881) & 0.734 (0.688--0.780) & 0.876 (0.855--0.897) & 0.716 (0.663--0.765) \\
H-Optimus-0 & 0.893 (0.868--0.918) & 0.852 (0.825--0.876) & 0.728 (0.679--0.774) & 0.869 (0.846--0.891) & 0.709 (0.655--0.760) \\
GigaPath & \underline{0.916 (0.894--0.936)} & 0.861 (0.833--0.885) & \underline{0.749 (0.700--0.793)} & 0.876 (0.851--0.899) & 0.737 (0.683--0.784) \\
CRISP & 0.908 (0.884--0.930) & 0.845 (0.817--0.870) & 0.719 (0.670--0.765) & 0.861 (0.836--0.884) & 0.696 (0.641--0.746) \\
Virchow2 & 0.905 (0.880--0.928) & 0.850 (0.822--0.876) & 0.731 (0.682--0.777) & 0.865 (0.838--0.889) & 0.715 (0.660--0.766) \\
Virchow & 0.899 (0.874--0.923) & 0.842 (0.816--0.868) & 0.716 (0.667--0.765) & 0.857 (0.833--0.881) & 0.695 (0.639--0.746) \\
UNI & 0.911 (0.888--0.933) & \textbf{0.871 (0.846--0.894)} & \textbf{0.761 (0.716--0.804)} & \textbf{0.887 (0.864--0.907)} & \underline{0.745 (0.694--0.794)} \\
Phikon-v2 & 0.870 (0.840--0.898) & 0.828 (0.801--0.853) & 0.682 (0.630--0.731) & 0.853 (0.831--0.873) & 0.646 (0.589--0.700) \\
\hline
\end{tabular}
\par\smallskip
\noindent\parbox{\textwidth}{\SuppTableNotes Values are point estimate (95\% CI). Binary-task CIs use DeLong’s method; multiclass AUC and the other metrics use 2,000-patient nonparametric bootstrap percentile intervals. Bold indicates the highest value for each metric within a task; underlining indicates the second-highest value.}
\end{table}

\begin{table}[p]
\centering
\caption{Performance of various models on internal test set for T10: Pilocytic astrocytoma (0) vs pleomorphic xanthoastrocytoma (1).}
\label{tab:dsmil_macro_metrics_task10}
\SuppTableStandard
\setlength{\tabcolsep}{4pt}
\begin{tabular}{@{}lccccc@{}}
\hline
Model & AUC & Accuracy & Sensitivity & Specificity & F1 score \\
\hline
\textbf{FOCUS} & \textbf{0.988 (0.965--1.000)} & \underline{0.930 (0.860--0.982)} & \textbf{0.923 (0.750--1.000)} & 0.932 (0.848--1.000) & \underline{0.857 (0.667--0.970)} \\
H-Optimus-1 & 0.963 (0.918--1.000) & \underline{0.930 (0.860--0.982)} & 0.769 (0.500--1.000) & \underline{0.977 (0.927--1.000)} & 0.833 (0.625--0.968) \\
H-Optimus-0 & 0.930 (0.862--0.998) & 0.877 (0.789--0.965) & 0.692 (0.400--0.923) & 0.932 (0.850--1.000) & 0.720 (0.454--0.897) \\
GigaPath & \underline{0.965 (0.921--1.000)} & \underline{0.930 (0.860--0.982)} & \underline{0.846 (0.611--1.000)} & 0.955 (0.881--1.000) & 0.846 (0.640--0.970) \\
CRISP & 0.962 (0.914--1.000) & 0.895 (0.807--0.965) & 0.769 (0.500--1.000) & 0.932 (0.848--1.000) & 0.769 (0.533--0.923) \\
Virchow2 & 0.932 (0.857--1.000) & 0.895 (0.807--0.965) & 0.692 (0.400--0.929) & 0.955 (0.886--1.000) & 0.750 (0.500--0.917) \\
Virchow & 0.960 (0.899--1.000) & \textbf{0.947 (0.877--1.000)} & 0.769 (0.500--1.000) & \textbf{1.000 (1.000--1.000)} & \textbf{0.870 (0.667--1.000)} \\
UNI & 0.963 (0.918--1.000) & \textbf{0.947 (0.877--1.000)} & 0.769 (0.500--1.000) & \textbf{1.000 (1.000--1.000)} & \textbf{0.870 (0.667--1.000)} \\
Phikon-v2 & 0.871 (0.778--0.964) & 0.807 (0.702--0.912) & 0.231 (0.000--0.500) & \underline{0.977 (0.927--1.000)} & 0.353 (0.000--0.621) \\
\hline
\end{tabular}
\par\smallskip
\noindent\parbox{\textwidth}{\SuppTableNotes Values are point estimate (95\% CI). Binary-task CIs use DeLong’s method; multiclass AUC and the other metrics use 2,000-patient nonparametric bootstrap percentile intervals. Bold indicates the highest value for each metric within a task; underlining indicates the second-highest value.}
\end{table}

\begin{table}[p]
\centering
\caption{Performance of various models on internal test set for T11: Glioma, metastasis or embryonal tumour (0) vs lymphoma (1).}
\label{tab:dsmil_macro_metrics_task12}
\SuppTableStandard
\setlength{\tabcolsep}{4pt}
\begin{tabular}{@{}lccccc@{}}
\hline
Model & AUC & Accuracy & Sensitivity & Specificity & F1 score \\
\hline
\textbf{FOCUS} & \textbf{0.973 (0.937--1.000)} & 0.987 (0.976--0.996) & 0.400 (0.100--0.714) & \textbf{1.000 (1.000--1.000)} & 0.571 (0.182--0.833) \\
H-Optimus-1 & \underline{0.963 (0.913--1.000)} & 0.989 (0.978--0.998) & 0.500 (0.167--0.833) & \textbf{1.000 (1.000--1.000)} & 0.667 (0.286--0.909) \\
H-Optimus-0 & 0.929 (0.816--1.000) & 0.989 (0.978--0.998) & 0.500 (0.167--0.833) & \textbf{1.000 (1.000--1.000)} & 0.667 (0.286--0.909) \\
GigaPath & 0.952 (0.888--1.000) & 0.989 (0.978--0.998) & 0.500 (0.167--0.833) & \textbf{1.000 (1.000--1.000)} & 0.667 (0.286--0.909) \\
CRISP & 0.956 (0.891--1.000) & \textbf{0.993 (0.985--1.000)} & \textbf{0.700 (0.400--1.000)} & \textbf{1.000 (1.000--1.000)} & \textbf{0.824 (0.571--1.000)} \\
Virchow2 & 0.924 (0.809--1.000) & 0.985 (0.971--0.993) & 0.500 (0.182--0.833) & 0.995 (0.989--1.000) & 0.588 (0.250--0.833) \\
Virchow & 0.919 (0.803--1.000) & \underline{0.991 (0.982--0.998)} & \textbf{0.700 (0.400--1.000)} & \underline{0.998 (0.993--1.000)} & \underline{0.778 (0.526--0.960)} \\
UNI & 0.900 (0.771--1.000) & \underline{0.991 (0.982--0.998)} & \underline{0.600 (0.266--0.917)} & \textbf{1.000 (1.000--1.000)} & 0.750 (0.421--0.957) \\
Phikon-v2 & 0.907 (0.792--1.000) & 0.982 (0.969--0.993) & 0.200 (0.000--0.500) & \textbf{1.000 (1.000--1.000)} & 0.333 (0.000--0.667) \\
\hline
\end{tabular}
\par\smallskip
\noindent\parbox{\textwidth}{\SuppTableNotes Values are point estimate (95\% CI). Binary-task CIs use DeLong’s method; multiclass AUC and the other metrics use 2,000-patient nonparametric bootstrap percentile intervals. Bold indicates the highest value for each metric within a task; underlining indicates the second-highest value.}
\end{table}

\begin{table}[p]
\centering
\caption{Performance of various models on internal test set for T12: ATRX unaltered (0) vs altered (1).}
\label{tab:dsmil_macro_metrics_task17}
\SuppTableStandard
\setlength{\tabcolsep}{4pt}
\begin{tabular}{@{}lccccc@{}}
\hline
Model & AUC & Accuracy & Sensitivity & Specificity & F1 score \\
\hline
\textbf{FOCUS} & \textbf{0.834 (0.779--0.890)} & \underline{0.831 (0.791--0.868)} & \textbf{0.592 (0.474--0.697)} & 0.897 (0.860--0.933) & \textbf{0.604 (0.500--0.688)} \\
H-Optimus-1 & 0.819 (0.760--0.878) & \underline{0.831 (0.791--0.868)} & \underline{0.539 (0.427--0.653)} & 0.912 (0.877--0.943) & 0.582 (0.482--0.675) \\
H-Optimus-0 & 0.811 (0.750--0.872) & \textbf{0.834 (0.794--0.871)} & \underline{0.539 (0.427--0.654)} & 0.916 (0.881--0.947) & \underline{0.586 (0.481--0.676)} \\
GigaPath & \underline{0.824 (0.770--0.879)} & 0.828 (0.785--0.865) & 0.513 (0.395--0.627) & 0.916 (0.880--0.946) & 0.565 (0.453--0.658) \\
CRISP & 0.817 (0.761--0.874) & 0.825 (0.785--0.863) & 0.500 (0.387--0.611) & 0.916 (0.880--0.947) & 0.555 (0.448--0.650) \\
Virchow2 & 0.808 (0.751--0.866) & 0.822 (0.782--0.862) & 0.421 (0.309--0.535) & 0.934 (0.903--0.961) & 0.508 (0.393--0.612) \\
Virchow & 0.809 (0.750--0.868) & 0.828 (0.785--0.865) & 0.408 (0.300--0.518) & \textbf{0.945 (0.917--0.970)} & 0.508 (0.393--0.613) \\
UNI & 0.817 (0.758--0.877) & \textbf{0.834 (0.794--0.871)} & \underline{0.539 (0.431--0.650)} & 0.916 (0.881--0.946) & \underline{0.586 (0.482--0.676)} \\
Phikon-v2 & 0.792 (0.729--0.856) & 0.825 (0.785--0.865) & 0.421 (0.312--0.529) & \underline{0.938 (0.908--0.965)} & 0.512 (0.400--0.613) \\
\hline
\end{tabular}
\par\smallskip
\noindent\parbox{\textwidth}{\SuppTableNotes Values are point estimate (95\% CI). Binary-task CIs use DeLong’s method; multiclass AUC and the other metrics use 2,000-patient nonparametric bootstrap percentile intervals. Bold indicates the highest value for each metric within a task; underlining indicates the second-highest value.}
\end{table}

\begin{table}[p]
\centering
\caption{Performance of various models on internal test set for T13: 1p/19q non-codeleted (0) vs codeleted (1).}
\label{tab:dsmil_macro_metrics_task18}
\SuppTableStandard
\setlength{\tabcolsep}{4pt}
\begin{tabular}{@{}lccccc@{}}
\hline
Model & AUC & Accuracy & Sensitivity & Specificity & F1 score \\
\hline
\textbf{FOCUS} & \textbf{0.967 (0.951--0.984)} & \textbf{0.911 (0.883--0.940)} & \textbf{0.720 (0.625--0.813)} & \underline{0.970 (0.948--0.989)} & \textbf{0.792 (0.719--0.859)} \\
H-Optimus-1 & \underline{0.954 (0.931--0.978)} & \underline{0.909 (0.877--0.937)} & \textbf{0.720 (0.618--0.815)} & 0.966 (0.944--0.985) & \underline{0.787 (0.708--0.855)} \\
H-Optimus-0 & 0.946 (0.921--0.971) & 0.889 (0.854--0.920) & 0.610 (0.500--0.716) & \textbf{0.974 (0.954--0.992)} & 0.719 (0.628--0.800) \\
GigaPath & 0.951 (0.929--0.974) & 0.897 (0.866--0.929) & \underline{0.683 (0.581--0.784)} & 0.963 (0.939--0.982) & 0.757 (0.672--0.827) \\
CRISP & 0.933 (0.905--0.962) & 0.891 (0.860--0.923) & 0.622 (0.518--0.727) & \textbf{0.974 (0.954--0.992)} & 0.729 (0.642--0.808) \\
Virchow2 & 0.924 (0.891--0.957) & 0.877 (0.843--0.911) & 0.610 (0.506--0.720) & 0.959 (0.933--0.981) & 0.699 (0.611--0.783) \\
Virchow & 0.939 (0.913--0.966) & 0.897 (0.866--0.929) & 0.659 (0.554--0.759) & \underline{0.970 (0.949--0.989)} & 0.750 (0.662--0.825) \\
UNI & 0.946 (0.922--0.969) & 0.889 (0.857--0.920) & 0.659 (0.558--0.761) & 0.959 (0.934--0.981) & 0.735 (0.649--0.810) \\
Phikon-v2 & 0.920 (0.888--0.952) & 0.863 (0.829--0.897) & 0.537 (0.425--0.642) & 0.963 (0.939--0.985) & 0.647 (0.547--0.734) \\
\hline
\end{tabular}
\par\smallskip
\noindent\parbox{\textwidth}{\SuppTableNotes Values are point estimate (95\% CI). Binary-task CIs use DeLong’s method; multiclass AUC and the other metrics use 2,000-patient nonparametric bootstrap percentile intervals. Bold indicates the highest value for each metric within a task; underlining indicates the second-highest value.}
\end{table}

\begin{table}[p]
\centering
\caption{Performance of various models on internal test set for T14: H3K27 unaltered (0) vs altered (1).}
\label{tab:dsmil_macro_metrics_task19}
\SuppTableStandard
\setlength{\tabcolsep}{4pt}
\begin{tabular}{@{}lccccc@{}}
\hline
Model & AUC & Accuracy & Sensitivity & Specificity & F1 score \\
\hline
\textbf{FOCUS} & \textbf{0.905 (0.858--0.951)} & 0.945 (0.919--0.968) & \underline{0.176 (0.000--0.375)} & 0.985 (0.970--0.997) & 0.240 (0.000--0.455) \\
H-Optimus-1 & 0.815 (0.722--0.908) & 0.945 (0.919--0.968) & 0.059 (0.000--0.188) & 0.991 (0.979--1.000) & 0.095 (0.000--0.273) \\
H-Optimus-0 & 0.816 (0.719--0.913) & 0.942 (0.916--0.965) & 0.059 (0.000--0.188) & 0.988 (0.975--0.997) & 0.091 (0.000--0.267) \\
GigaPath & 0.833 (0.750--0.917) & 0.948 (0.922--0.968) & \underline{0.176 (0.000--0.375)} & 0.988 (0.975--0.997) & 0.250 (0.000--0.471) \\
CRISP & 0.791 (0.697--0.884) & \textbf{0.957 (0.934--0.977)} & \textbf{0.294 (0.083--0.533)} & 0.991 (0.979--1.000) & \textbf{0.400 (0.125--0.621)} \\
Virchow2 & 0.753 (0.638--0.867) & 0.945 (0.919--0.968) & \underline{0.176 (0.000--0.375)} & 0.985 (0.970--0.997) & 0.240 (0.000--0.457) \\
Virchow & \underline{0.838 (0.760--0.917)} & \underline{0.954 (0.931--0.974)} & \underline{0.176 (0.000--0.375)} & \underline{0.994 (0.985--1.000)} & \underline{0.273 (0.000--0.500)} \\
UNI & 0.829 (0.746--0.912) & 0.939 (0.913--0.962) & \underline{0.176 (0.000--0.375)} & 0.979 (0.962--0.994) & 0.222 (0.000--0.424) \\
Phikon-v2 & 0.721 (0.607--0.834) & 0.951 (0.925--0.971) & 0.000 (0.000--0.000) & \textbf{1.000 (1.000--1.000)} & 0.000 (0.000--0.000) \\
\hline
\end{tabular}
\par\smallskip
\noindent\parbox{\textwidth}{\SuppTableNotes Values are point estimate (95\% CI). Binary-task CIs use DeLong’s method; multiclass AUC and the other metrics use 2,000-patient nonparametric bootstrap percentile intervals. Bold indicates the highest value for each metric within a task; underlining indicates the second-highest value.}
\end{table}

\begin{table}[p]
\centering
\caption{Performance of various models on internal test set for T15: H3G34 wild type (0) vs mutant (1).}
\label{tab:dsmil_macro_metrics_task20}
\SuppTableStandard
\setlength{\tabcolsep}{4pt}
\begin{tabular}{@{}lccccc@{}}
\hline
Model & AUC & Accuracy & Sensitivity & Specificity & F1 score \\
\hline
\textbf{FOCUS} & \textbf{0.981 (0.962--1.000)} & \underline{0.986 (0.971--0.997)} & \textbf{0.250 (0.000--1.000)} & 0.994 (0.986--1.000) & \textbf{0.286 (0.000--0.667)} \\
H-Optimus-1 & 0.974 (0.941--1.000) & \underline{0.986 (0.971--0.994)} & \underline{0.000 (0.000--0.000)} & \underline{0.997 (0.991--1.000)} & \underline{0.000 (0.000--0.000)} \\
H-Optimus-0 & 0.966 (0.918--1.000) & 0.983 (0.969--0.994) & \underline{0.000 (0.000--0.000)} & 0.994 (0.986--1.000) & \underline{0.000 (0.000--0.000)} \\
GigaPath & 0.965 (0.907--1.000) & \textbf{0.989 (0.977--0.997)} & \underline{0.000 (0.000--0.000)} & \textbf{1.000 (1.000--1.000)} & \underline{0.000 (0.000--0.000)} \\
CRISP & \textbf{0.981 (0.965--0.998)} & \textbf{0.989 (0.977--0.997)} & \underline{0.000 (0.000--0.000)} & \textbf{1.000 (1.000--1.000)} & \underline{0.000 (0.000--0.000)} \\
Virchow2 & 0.978 (0.953--1.000) & \textbf{0.989 (0.977--0.997)} & \underline{0.000 (0.000--0.000)} & \textbf{1.000 (1.000--1.000)} & \underline{0.000 (0.000--0.000)} \\
Virchow & 0.880 (0.656--1.000) & \textbf{0.989 (0.977--0.997)} & \underline{0.000 (0.000--0.000)} & \textbf{1.000 (1.000--1.000)} & \underline{0.000 (0.000--0.000)} \\
UNI & 0.973 (0.935--1.000) & \textbf{0.989 (0.977--0.997)} & \underline{0.000 (0.000--0.000)} & \textbf{1.000 (1.000--1.000)} & \underline{0.000 (0.000--0.000)} \\
Phikon-v2 & 0.915 (0.759--1.000) & \textbf{0.989 (0.977--0.997)} & \underline{0.000 (0.000--0.000)} & \textbf{1.000 (1.000--1.000)} & \underline{0.000 (0.000--0.000)} \\
\hline
\end{tabular}
\par\smallskip
\noindent\parbox{\textwidth}{\SuppTableNotes Values are point estimate (95\% CI). Binary-task CIs use DeLong’s method; multiclass AUC and the other metrics use 2,000-patient nonparametric bootstrap percentile intervals. Bold indicates the highest value for each metric within a task; underlining indicates the second-highest value.}
\end{table}

\begin{table}[p]
\centering
\caption{Performance of various models on internal test set for T16: BRAF wild type (0) vs mutant (1).}
\label{tab:dsmil_macro_metrics_task21}
\SuppTableStandard
\setlength{\tabcolsep}{4pt}
\begin{tabular}{@{}lccccc@{}}
\hline
Model & AUC & Accuracy & Sensitivity & Specificity & F1 score \\
\hline
\textbf{FOCUS} & \textbf{0.836 (0.722--0.950)} & 0.946 (0.920--0.968) & \textbf{0.133 (0.000--0.333)} & 0.982 (0.967--0.994) & 0.174 (0.000--0.385) \\
H-Optimus-1 & 0.808 (0.693--0.922) & 0.954 (0.931--0.974) & \underline{0.067 (0.000--0.222)} & 0.994 (0.985--1.000) & 0.111 (0.000--0.333) \\
H-Optimus-0 & 0.786 (0.656--0.916) & 0.957 (0.934--0.977) & 0.000 (0.000--0.000) & \textbf{1.000 (1.000--1.000)} & 0.000 (0.000--0.000) \\
GigaPath & \underline{0.825 (0.706--0.943)} & \underline{0.960 (0.937--0.980)} & \textbf{0.133 (0.000--0.333)} & \underline{0.997 (0.991--1.000)} & \underline{0.222 (0.000--0.467)} \\
CRISP & 0.800 (0.672--0.928) & 0.954 (0.931--0.974) & 0.000 (0.000--0.000) & \underline{0.997 (0.991--1.000)} & 0.000 (0.000--0.000) \\
Virchow2 & 0.790 (0.665--0.916) & 0.957 (0.934--0.977) & 0.000 (0.000--0.000) & \textbf{1.000 (1.000--1.000)} & 0.000 (0.000--0.000) \\
Virchow & 0.767 (0.649--0.885) & \textbf{0.963 (0.943--0.980)} & \textbf{0.133 (0.000--0.333)} & \textbf{1.000 (1.000--1.000)} & \textbf{0.235 (0.000--0.500)} \\
UNI & 0.819 (0.713--0.925) & 0.957 (0.934--0.977) & \textbf{0.133 (0.000--0.333)} & 0.994 (0.985--1.000) & 0.211 (0.000--0.462) \\
Phikon-v2 & 0.777 (0.672--0.882) & 0.957 (0.934--0.977) & 0.000 (0.000--0.000) & \textbf{1.000 (1.000--1.000)} & 0.000 (0.000--0.000) \\
\hline
\end{tabular}
\par\smallskip
\noindent\parbox{\textwidth}{\SuppTableNotes Values are point estimate (95\% CI). Binary-task CIs use DeLong’s method; multiclass AUC and the other metrics use 2,000-patient nonparametric bootstrap percentile intervals. Bold indicates the highest value for each metric within a task; underlining indicates the second-highest value.}
\end{table}

\begin{table}[p]
\centering
\caption{Performance of various models on internal test set for T17: CDKN2A/B retained (0) vs deleted (1).}
\label{tab:dsmil_macro_metrics_task22}
\SuppTableStandard
\setlength{\tabcolsep}{4pt}
\begin{tabular}{@{}lccccc@{}}
\hline
Model & AUC & Accuracy & Sensitivity & Specificity & F1 score \\
\hline
\textbf{FOCUS} & \textbf{0.823 (0.774--0.872)} & \textbf{0.814 (0.774--0.854)} & \textbf{0.517 (0.417--0.624)} & 0.916 (0.881--0.950) & \textbf{0.586 (0.486--0.675)} \\
H-Optimus-1 & \underline{0.817 (0.767--0.867)} & 0.797 (0.754--0.840) & 0.427 (0.325--0.528) & 0.923 (0.889--0.955) & 0.517 (0.415--0.610) \\
H-Optimus-0 & 0.808 (0.757--0.859) & 0.794 (0.751--0.837) & \underline{0.438 (0.337--0.544)} & 0.916 (0.880--0.947) & 0.520 (0.422--0.614) \\
GigaPath & 0.802 (0.751--0.853) & 0.797 (0.754--0.840) & 0.427 (0.326--0.533) & 0.923 (0.888--0.955) & 0.517 (0.415--0.610) \\
CRISP & 0.802 (0.750--0.853) & 0.780 (0.737--0.823) & 0.236 (0.152--0.326) & \textbf{0.966 (0.942--0.985)} & 0.353 (0.245--0.457) \\
Virchow2 & 0.801 (0.749--0.854) & 0.777 (0.734--0.823) & 0.247 (0.159--0.337) & \underline{0.958 (0.932--0.981)} & 0.361 (0.244--0.466) \\
Virchow & 0.789 (0.733--0.845) & \underline{0.811 (0.771--0.851)} & \underline{0.438 (0.337--0.543)} & 0.939 (0.907--0.966) & \underline{0.542 (0.443--0.634)} \\
UNI & 0.792 (0.738--0.845) & 0.783 (0.740--0.829) & 0.337 (0.237--0.436) & 0.935 (0.904--0.963) & 0.441 (0.331--0.544) \\
Phikon-v2 & 0.767 (0.711--0.823) & 0.780 (0.737--0.826) & 0.281 (0.188--0.380) & 0.950 (0.923--0.974) & 0.394 (0.281--0.500) \\
\hline
\end{tabular}
\par\smallskip
\noindent\parbox{\textwidth}{\SuppTableNotes Values are point estimate (95\% CI). Binary-task CIs use DeLong’s method; multiclass AUC and the other metrics use 2,000-patient nonparametric bootstrap percentile intervals. Bold indicates the highest value for each metric within a task; underlining indicates the second-highest value.}
\end{table}

\begin{table}[p]
\centering
\caption{Performance of various models on internal test set for T18: TP53 wild type (0) vs mutant (1).}
\label{tab:dsmil_macro_metrics_task24}
\SuppTableStandard
\setlength{\tabcolsep}{4pt}
\begin{tabular}{@{}lccccc@{}}
\hline
Model & AUC & Accuracy & Sensitivity & Specificity & F1 score \\
\hline
\textbf{FOCUS} & \textbf{0.866 (0.826--0.907)} & \textbf{0.820 (0.777--0.857)} & 0.737 (0.659--0.809) & 0.871 (0.825--0.912) & \underline{0.757 (0.700--0.809)} \\
H-Optimus-1 & 0.844 (0.800--0.888) & 0.803 (0.760--0.846) & 0.654 (0.570--0.735) & \underline{0.894 (0.852--0.932)} & 0.716 (0.648--0.777) \\
H-Optimus-0 & 0.837 (0.792--0.881) & 0.803 (0.760--0.846) & 0.654 (0.576--0.733) & \underline{0.894 (0.853--0.932)} & 0.716 (0.651--0.775) \\
GigaPath & 0.843 (0.800--0.886) & 0.800 (0.757--0.840) & 0.662 (0.576--0.742) & 0.885 (0.841--0.925) & 0.715 (0.645--0.774) \\
CRISP & 0.834 (0.790--0.878) & 0.791 (0.749--0.831) & \underline{0.752 (0.675--0.820)} & 0.816 (0.763--0.865) & 0.733 (0.672--0.786) \\
Virchow2 & 0.829 (0.783--0.875) & 0.786 (0.743--0.829) & 0.602 (0.520--0.682) & \textbf{0.899 (0.857--0.936)} & 0.681 (0.611--0.745) \\
Virchow & 0.825 (0.780--0.870) & 0.783 (0.737--0.826) & 0.669 (0.585--0.746) & 0.853 (0.804--0.899) & 0.701 (0.632--0.759) \\
UNI & \underline{0.859 (0.817--0.900)} & \underline{0.811 (0.769--0.851)} & \textbf{0.782 (0.706--0.847)} & 0.829 (0.778--0.877) & \textbf{0.759 (0.699--0.811)} \\
Phikon-v2 & 0.812 (0.765--0.859) & 0.769 (0.723--0.811) & 0.669 (0.586--0.748) & 0.829 (0.778--0.877) & 0.687 (0.622--0.745) \\
\hline
\end{tabular}
\par\smallskip
\noindent\parbox{\textwidth}{\SuppTableNotes Values are point estimate (95\% CI). Binary-task CIs use DeLong’s method; multiclass AUC and the other metrics use 2,000-patient nonparametric bootstrap percentile intervals. Bold indicates the highest value for each metric within a task; underlining indicates the second-highest value.}
\end{table}

\begin{table}[p]
\centering
\caption{Performance of various models on internal test set for T19: IDH1/IDH2 wild type (0) vs either gene mutated (1).}
\label{tab:dsmil_macro_metrics_task25}
\SuppTableStandard
\setlength{\tabcolsep}{4pt}
\begin{tabular}{@{}lccccc@{}}
\hline
Model & AUC & Accuracy & Sensitivity & Specificity & F1 score \\
\hline
\textbf{FOCUS} & \textbf{0.955 (0.935--0.975)} & \textbf{0.885 (0.848--0.919)} & \textbf{0.917 (0.872--0.957)} & 0.855 (0.800--0.907) & \textbf{0.885 (0.844--0.920)} \\
H-Optimus-1 & 0.922 (0.892--0.952) & 0.863 (0.823--0.898) & 0.891 (0.839--0.936) & 0.837 (0.778--0.893) & 0.863 (0.818--0.900) \\
H-Optimus-0 & 0.920 (0.889--0.950) & 0.848 (0.807--0.888) & 0.897 (0.849--0.940) & 0.801 (0.737--0.861) & 0.851 (0.807--0.891) \\
GigaPath & \underline{0.939 (0.915--0.963)} & 0.860 (0.820--0.898) & 0.904 (0.856--0.946) & 0.819 (0.757--0.877) & 0.862 (0.818--0.900) \\
CRISP & 0.927 (0.900--0.954) & 0.848 (0.804--0.885) & 0.865 (0.810--0.918) & 0.831 (0.772--0.887) & 0.846 (0.796--0.888) \\
Virchow2 & 0.927 (0.900--0.955) & 0.851 (0.811--0.891) & 0.865 (0.810--0.918) & 0.837 (0.780--0.894) & 0.849 (0.801--0.891) \\
Virchow & 0.916 (0.886--0.947) & 0.854 (0.814--0.891) & \underline{0.910 (0.865--0.952)} & 0.801 (0.737--0.861) & 0.858 (0.814--0.898) \\
UNI & 0.928 (0.901--0.956) & \underline{0.873 (0.835--0.910)} & 0.872 (0.817--0.921) & \textbf{0.873 (0.823--0.920)} & \underline{0.869 (0.826--0.907)} \\
Phikon-v2 & 0.901 (0.868--0.934) & 0.801 (0.755--0.845) & 0.737 (0.667--0.807) & \underline{0.861 (0.806--0.911)} & 0.782 (0.727--0.833) \\
\hline
\end{tabular}
\par\smallskip
\noindent\parbox{\textwidth}{\SuppTableNotes Values are point estimate (95\% CI). Binary-task CIs use DeLong’s method; multiclass AUC and the other metrics use 2,000-patient nonparametric bootstrap percentile intervals. Bold indicates the highest value for each metric within a task; underlining indicates the second-highest value.}
\end{table}

\begin{table}[p]
\centering
\caption{Performance of various models on internal test set for T20: MGMT unmethylated (0) vs methylated (1).}
\label{tab:dsmil_macro_metrics_task26}
\SuppTableStandard
\setlength{\tabcolsep}{4pt}
\begin{tabular}{@{}lccccc@{}}
\hline
Model & AUC & Accuracy & Sensitivity & Specificity & F1 score \\
\hline
\textbf{FOCUS} & \textbf{0.736 (0.677--0.794)} & \textbf{0.680 (0.625--0.735)} & 0.779 (0.714--0.843) & 0.554 (0.472--0.644) & \underline{0.732 (0.673--0.782)} \\
H-Optimus-1 & \underline{0.734 (0.676--0.792)} & 0.655 (0.600--0.709) & 0.721 (0.653--0.791) & 0.570 (0.483--0.658) & 0.700 (0.642--0.756) \\
H-Optimus-0 & 0.705 (0.644--0.767) & 0.655 (0.600--0.709) & 0.701 (0.631--0.773) & \textbf{0.595 (0.513--0.680)} & 0.695 (0.632--0.748) \\
GigaPath & 0.727 (0.668--0.787) & 0.658 (0.604--0.713) & 0.721 (0.652--0.789) & \underline{0.579 (0.492--0.667)} & 0.703 (0.642--0.757) \\
CRISP & 0.707 (0.646--0.768) & 0.644 (0.589--0.698) & 0.773 (0.708--0.837) & 0.479 (0.392--0.565) & 0.708 (0.650--0.760) \\
Virchow2 & 0.718 (0.658--0.778) & 0.673 (0.618--0.727) & 0.786 (0.721--0.848) & 0.529 (0.440--0.618) & 0.729 (0.674--0.781) \\
Virchow & 0.703 (0.642--0.764) & 0.629 (0.571--0.684) & 0.753 (0.684--0.819) & 0.471 (0.386--0.557) & 0.695 (0.633--0.747) \\
UNI & 0.711 (0.650--0.772) & \underline{0.676 (0.622--0.731)} & \textbf{0.838 (0.779--0.894)} & 0.471 (0.381--0.560) & \textbf{0.744 (0.690--0.792)} \\
Phikon-v2 & 0.672 (0.607--0.738) & 0.662 (0.607--0.716) & \underline{0.799 (0.733--0.861)} & 0.488 (0.398--0.576) & 0.726 (0.669--0.777) \\
\hline
\end{tabular}
\par\smallskip
\noindent\parbox{\textwidth}{\SuppTableNotes Values are point estimate (95\% CI). Binary-task CIs use DeLong’s method; multiclass AUC and the other metrics use 2,000-patient nonparametric bootstrap percentile intervals. Bold indicates the highest value for each metric within a task; underlining indicates the second-highest value.}
\end{table}

\begin{table}[p]
\centering
\caption{Performance of various models on internal test set for T21: No concurrent chromosome 7 gain and chromosome 10 loss (0) vs concurrent gain/loss (1).}
\label{tab:dsmil_macro_metrics_task28}
\SuppTableStandard
\setlength{\tabcolsep}{4pt}
\begin{tabular}{@{}lccccc@{}}
\hline
Model & AUC & Accuracy & Sensitivity & Specificity & F1 score \\
\hline
\textbf{FOCUS} & \textbf{0.884 (0.846--0.923)} & \textbf{0.838 (0.798--0.879)} & \textbf{0.273 (0.157--0.395)} & 0.955 (0.927--0.978) & \textbf{0.366 (0.229--0.494)} \\
H-Optimus-1 & 0.836 (0.791--0.881) & 0.807 (0.763--0.850) & \underline{0.127 (0.043--0.217)} & 0.947 (0.919--0.973) & \underline{0.184 (0.063--0.299)} \\
H-Optimus-0 & 0.830 (0.782--0.878) & 0.810 (0.766--0.850) & 0.109 (0.035--0.196) & 0.955 (0.928--0.977) & 0.164 (0.056--0.277) \\
GigaPath & 0.837 (0.790--0.884) & 0.816 (0.773--0.860) & 0.091 (0.019--0.173) & 0.966 (0.943--0.985) & 0.145 (0.032--0.263) \\
CRISP & 0.831 (0.779--0.884) & 0.822 (0.782--0.863) & 0.091 (0.019--0.173) & 0.974 (0.953--0.992) & 0.149 (0.032--0.267) \\
Virchow2 & 0.825 (0.770--0.880) & \underline{0.826 (0.782--0.866)} & 0.109 (0.035--0.200) & 0.974 (0.954--0.989) & 0.176 (0.060--0.301) \\
Virchow & 0.824 (0.770--0.877) & 0.822 (0.782--0.863) & 0.055 (0.000--0.125) & \underline{0.981 (0.963--0.996)} & 0.095 (0.000--0.206) \\
UNI & \underline{0.847 (0.797--0.897)} & \underline{0.826 (0.785--0.866)} & 0.036 (0.000--0.094) & \textbf{0.989 (0.974--1.000)} & 0.067 (0.000--0.164) \\
Phikon-v2 & 0.804 (0.755--0.854) & 0.813 (0.769--0.854) & 0.000 (0.000--0.000) & \underline{0.981 (0.964--0.996)} & 0.000 (0.000--0.000) \\
\hline
\end{tabular}
\par\smallskip
\noindent\parbox{\textwidth}{\SuppTableNotes Values are point estimate (95\% CI). Binary-task CIs use DeLong’s method; multiclass AUC and the other metrics use 2,000-patient nonparametric bootstrap percentile intervals. Bold indicates the highest value for each metric within a task; underlining indicates the second-highest value.}
\end{table}

\begin{table}[p]
\centering
\caption{Performance of various models on internal test set for T22: EGFR non-amplified (0) vs amplified (1).}
\label{tab:dsmil_macro_metrics_task31}
\SuppTableStandard
\setlength{\tabcolsep}{4pt}
\begin{tabular}{@{}lccccc@{}}
\hline
Model & AUC & Accuracy & Sensitivity & Specificity & F1 score \\
\hline
\textbf{FOCUS} & \textbf{0.855 (0.790--0.920)} & \underline{0.888 (0.854--0.919)} & \textbf{0.574 (0.425--0.712)} & 0.942 (0.913--0.967) & \textbf{0.600 (0.472--0.706)} \\
H-Optimus-1 & 0.828 (0.758--0.899) & 0.866 (0.826--0.900) & \underline{0.489 (0.341--0.643)} & 0.931 (0.899--0.957) & 0.517 (0.381--0.634) \\
H-Optimus-0 & 0.813 (0.738--0.889) & 0.866 (0.829--0.900) & 0.447 (0.302--0.591) & 0.938 (0.910--0.964) & 0.494 (0.354--0.614) \\
GigaPath & \underline{0.843 (0.773--0.913)} & 0.872 (0.835--0.907) & 0.468 (0.333--0.618) & 0.942 (0.912--0.968) & \underline{0.518 (0.382--0.636)} \\
CRISP & 0.797 (0.719--0.874) & 0.872 (0.835--0.907) & 0.319 (0.189--0.455) & 0.967 (0.944--0.986) & 0.423 (0.271--0.558) \\
Virchow2 & 0.797 (0.718--0.876) & 0.875 (0.838--0.910) & 0.277 (0.151--0.420) & 0.978 (0.960--0.993) & 0.394 (0.233--0.537) \\
Virchow & 0.804 (0.731--0.877) & 0.882 (0.844--0.913) & 0.277 (0.151--0.412) & 0.985 (0.970--0.996) & 0.406 (0.241--0.545) \\
UNI & 0.829 (0.761--0.896) & \textbf{0.891 (0.857--0.925)} & 0.319 (0.191--0.457) & \underline{0.989 (0.975--1.000)} & 0.462 (0.301--0.606) \\
Phikon-v2 & 0.779 (0.704--0.854) & 0.866 (0.829--0.900) & 0.085 (0.020--0.170) & \textbf{1.000 (1.000--1.000)} & 0.157 (0.038--0.291) \\
\hline
\end{tabular}
\par\smallskip
\noindent\parbox{\textwidth}{\SuppTableNotes Values are point estimate (95\% CI). Binary-task CIs use DeLong’s method; multiclass AUC and the other metrics use 2,000-patient nonparametric bootstrap percentile intervals. Bold indicates the highest value for each metric within a task; underlining indicates the second-highest value.}
\end{table}

\begin{table}[p]
\centering
\caption{Performance of various models on internal test set for T23: TERT c.-124/-146 wild type (0) vs either promoter mutation (1).}
\label{tab:dsmil_macro_metrics_task32}
\SuppTableStandard
\setlength{\tabcolsep}{4pt}
\begin{tabular}{@{}lccccc@{}}
\hline
Model & AUC & Accuracy & Sensitivity & Specificity & F1 score \\
\hline
\textbf{FOCUS} & \textbf{0.877 (0.840--0.915)} & \textbf{0.823 (0.780--0.866)} & 0.805 (0.741--0.867) & \textbf{0.840 (0.782--0.897)} & \textbf{0.818 (0.769--0.863)} \\
H-Optimus-1 & 0.856 (0.816--0.897) & 0.764 (0.720--0.807) & \textbf{0.818 (0.756--0.875)} & 0.712 (0.640--0.778) & 0.774 (0.724--0.821) \\
H-Optimus-0 & 0.828 (0.784--0.873) & 0.752 (0.705--0.798) & \underline{0.811 (0.750--0.871)} & 0.693 (0.622--0.765) & 0.763 (0.709--0.811) \\
GigaPath & \underline{0.865 (0.826--0.904)} & \underline{0.786 (0.739--0.829)} & 0.767 (0.701--0.830) & \underline{0.804 (0.742--0.863)} & \underline{0.780 (0.725--0.828)} \\
CRISP & 0.852 (0.811--0.892) & 0.755 (0.705--0.801) & 0.774 (0.707--0.838) & 0.736 (0.667--0.804) & 0.757 (0.703--0.807) \\
Virchow2 & 0.829 (0.785--0.873) & 0.742 (0.696--0.789) & 0.799 (0.736--0.860) & 0.687 (0.615--0.762) & 0.754 (0.701--0.802) \\
Virchow & 0.841 (0.798--0.884) & 0.767 (0.717--0.814) & 0.792 (0.729--0.857) & 0.742 (0.673--0.811) & 0.771 (0.716--0.821) \\
UNI & 0.848 (0.806--0.890) & 0.767 (0.717--0.811) & 0.742 (0.673--0.809) & 0.791 (0.728--0.854) & 0.759 (0.702--0.808) \\
Phikon-v2 & 0.806 (0.759--0.852) & 0.714 (0.665--0.764) & 0.805 (0.744--0.865) & 0.626 (0.551--0.702) & 0.736 (0.681--0.784) \\
\hline
\end{tabular}
\par\smallskip
\noindent\parbox{\textwidth}{\SuppTableNotes Values are point estimate (95\% CI). Binary-task CIs use DeLong’s method; multiclass AUC and the other metrics use 2,000-patient nonparametric bootstrap percentile intervals. Bold indicates the highest value for each metric within a task; underlining indicates the second-highest value.}
\end{table}

\endgroup


\begingroup
\renewcommand{\arraystretch}{1.18}
\begin{landscape}
\begin{table}[p]
\centering
\caption{Performance across foundation models on diagnostic tasks (T1--T11) in the temporal cohort.}
\label{tab:external1_diagnosis_auc_acc}
\SuppTableStandard
\setlength{\tabcolsep}{2pt}
\begin{tabular}{@{}lccccccccc@{}}
\hline
Task & FOCUS & H-Optimus-1 & H-Optimus-0 & GigaPath & CRISP & Virchow2 & Virchow & UNI & Phikon-v2 \\
\hline
T1 & \makecell[c]{\textbf{0.954}\\\textbf{(0.920--0.987)}} & \makecell[c]{0.950\\(0.920--0.980)} & \makecell[c]{0.949\\(0.915--0.982)} & \makecell[c]{0.946\\(0.911--0.980)} & \makecell[c]{0.942\\(0.905--0.979)} & \makecell[c]{\underline{0.953}\\\underline{(0.923--0.983)}} & \makecell[c]{0.941\\(0.906--0.976)} & \makecell[c]{0.945\\(0.911--0.978)} & \makecell[c]{0.935\\(0.899--0.971)} \\
T2 & \makecell[c]{\textbf{0.961}\\\textbf{(0.923--0.998)}} & \makecell[c]{0.924\\(0.835--1.000)} & \makecell[c]{0.916\\(0.813--1.000)} & \makecell[c]{0.917\\(0.820--1.000)} & \makecell[c]{0.908\\(0.837--0.978)} & \makecell[c]{0.895\\(0.769--1.000)} & \makecell[c]{\underline{0.947}\\\underline{(0.901--0.992)}} & \makecell[c]{0.907\\(0.779--1.000)} & \makecell[c]{0.890\\(0.795--0.985)} \\
T3 & \makecell[c]{\textbf{0.967}\\\textbf{(0.928--1.000)}} & \makecell[c]{0.952\\(0.904--0.999)} & \makecell[c]{0.942\\(0.888--0.997)} & \makecell[c]{0.957\\(0.915--1.000)} & \makecell[c]{0.959\\(0.919--0.999)} & \makecell[c]{0.938\\(0.883--0.992)} & \makecell[c]{0.922\\(0.863--0.982)} & \makecell[c]{0.954\\(0.909--0.999)} & \makecell[c]{\underline{0.967}\\\underline{(0.927--1.000)}} \\
T4 & \makecell[c]{\textbf{0.953}\\\textbf{(0.898--0.988)}} & \makecell[c]{0.943\\(0.868--0.987)} & \makecell[c]{0.950\\(0.894--0.988)} & \makecell[c]{0.950\\(0.899--0.984)} & \makecell[c]{\underline{0.952}\\\underline{(0.893--0.989)}} & \makecell[c]{0.944\\(0.895--0.980)} & \makecell[c]{0.952\\(0.887--0.989)} & \makecell[c]{0.943\\(0.875--0.984)} & \makecell[c]{0.949\\(0.904--0.980)} \\
T5 & \makecell[c]{\underline{0.956}\\\underline{(0.916--0.996)}} & \makecell[c]{0.952\\(0.907--0.998)} & \makecell[c]{0.952\\(0.916--0.988)} & \makecell[c]{\textbf{0.957}\\\textbf{(0.919--0.995)}} & \makecell[c]{0.955\\(0.906--1.000)} & \makecell[c]{0.951\\(0.906--0.995)} & \makecell[c]{0.948\\(0.909--0.987)} & \makecell[c]{0.951\\(0.912--0.990)} & \makecell[c]{0.909\\(0.852--0.966)} \\
T6 & \makecell[c]{0.948\\(0.885--1.000)} & \makecell[c]{\underline{0.951}\\\underline{(0.892--1.000)}} & \makecell[c]{0.941\\(0.875--1.000)} & \makecell[c]{\textbf{0.952}\\\textbf{(0.900--1.000)}} & \makecell[c]{0.949\\(0.888--1.000)} & \makecell[c]{0.937\\(0.864--1.000)} & \makecell[c]{0.921\\(0.843--0.998)} & \makecell[c]{0.938\\(0.871--1.000)} & \makecell[c]{0.897\\(0.813--0.981)} \\
T7 & \makecell[c]{\textbf{0.896}\\\textbf{(0.738--0.981)}} & \makecell[c]{0.824\\(0.653--0.945)} & \makecell[c]{0.787\\(0.602--0.928)} & \makecell[c]{0.840\\(0.649--0.958)} & \makecell[c]{0.838\\(0.666--0.959)} & \makecell[c]{0.841\\(0.691--0.949)} & \makecell[c]{\underline{0.849}\\\underline{(0.692--0.955)}} & \makecell[c]{0.793\\(0.589--0.932)} & \makecell[c]{0.848\\(0.689--0.960)} \\
T8 & \makecell[c]{\textbf{0.961}\\\textbf{(0.928--0.995)}} & \makecell[c]{0.907\\(0.836--0.979)} & \makecell[c]{0.926\\(0.866--0.986)} & \makecell[c]{\underline{0.950}\\\underline{(0.914--0.986)}} & \makecell[c]{0.936\\(0.878--0.994)} & \makecell[c]{0.889\\(0.799--0.979)} & \makecell[c]{0.885\\(0.798--0.972)} & \makecell[c]{0.923\\(0.871--0.975)} & \makecell[c]{0.847\\(0.727--0.967)} \\
T9 & \makecell[c]{\textbf{0.939}\\\textbf{(0.905--0.966)}} & \makecell[c]{0.882\\(0.833--0.923)} & \makecell[c]{0.871\\(0.821--0.915)} & \makecell[c]{0.898\\(0.858--0.934)} & \makecell[c]{0.891\\(0.849--0.929)} & \makecell[c]{0.898\\(0.855--0.937)} & \makecell[c]{0.879\\(0.829--0.922)} & \makecell[c]{\underline{0.909}\\\underline{(0.866--0.945)}} & \makecell[c]{0.849\\(0.795--0.899)} \\
T10 & \makecell[c]{\underline{0.969}\\\underline{(0.882--1.000)}} & \makecell[c]{\underline{0.969}\\\underline{(0.882--1.000)}} & \makecell[c]{\underline{0.969}\\\underline{(0.882--1.000)}} & \makecell[c]{0.938\\(0.791--1.000)} & \makecell[c]{\underline{0.969}\\\underline{(0.882--1.000)}} & \makecell[c]{\underline{0.969}\\\underline{(0.882--1.000)}} & \makecell[c]{0.938\\(0.796--1.000)} & \makecell[c]{\underline{0.969}\\\underline{(0.882--1.000)}} & \makecell[c]{\textbf{1.000}\\\textbf{(1.000--1.000)}} \\
T11 & \makecell[c]{\textbf{0.997}\\\textbf{(0.990--1.000)}} & \makecell[c]{0.976\\(0.927--1.000)} & \makecell[c]{0.921\\(0.764--1.000)} & \makecell[c]{\underline{0.987}\\\underline{(0.959--1.000)}} & \makecell[c]{0.974\\(0.920--1.000)} & \makecell[c]{0.982\\(0.943--1.000)} & \makecell[c]{0.971\\(0.912--1.000)} & \makecell[c]{0.984\\(0.951--1.000)} & \makecell[c]{0.800\\(0.417--1.000)} \\
\hline
\end{tabular}
\par\smallskip
\noindent\parbox{\linewidth}{\SuppTableNotes Values are AUC (95\% CI). Binary-task AUC intervals use DeLong's method and multiclass AUC intervals use 2,000-resample percentile bootstrap; when DeLong's interval is undefined because a class has fewer than two samples, percentile bootstrap is used. ACC intervals use Wilson's method. Bold indicates the highest value within a task; underlining indicates the second-highest value.}
\end{table}
\end{landscape}
\begin{landscape}
\begin{table}[p]
\centering
\caption{Performance across foundation models on diagnostic tasks (T1--T11) in the external cohort.}
\label{tab:external2_diagnosis_auc_acc}
\SuppTableStandard
\setlength{\tabcolsep}{2pt}
\begin{tabular}{@{}lccccccccc@{}}
\hline
Task & FOCUS & H-Optimus-1 & H-Optimus-0 & GigaPath & CRISP & Virchow2 & Virchow & UNI & Phikon-v2 \\
\hline
T1 & \makecell[c]{\textbf{0.969}\\\textbf{(0.932--1.000)}} & \makecell[c]{\underline{0.967}\\\underline{(0.937--0.997)}} & \makecell[c]{0.964\\(0.936--0.993)} & \makecell[c]{0.941\\(0.891--0.991)} & \makecell[c]{0.954\\(0.910--0.998)} & \makecell[c]{0.956\\(0.915--0.998)} & \makecell[c]{0.950\\(0.901--0.999)} & \makecell[c]{0.943\\(0.889--0.997)} & \makecell[c]{0.895\\(0.786--1.000)} \\
T2 & \makecell[c]{\textbf{0.942}\\\textbf{(0.895--0.988)}} & \makecell[c]{0.869\\(0.760--0.978)} & \makecell[c]{0.858\\(0.739--0.977)} & \makecell[c]{0.836\\(0.750--0.921)} & \makecell[c]{\underline{0.939}\\\underline{(0.900--0.978)}} & \makecell[c]{0.842\\(0.725--0.958)} & \makecell[c]{0.805\\(0.697--0.913)} & \makecell[c]{0.922\\(0.874--0.969)} & \makecell[c]{0.814\\(0.675--0.954)} \\
T3 & \makecell[c]{\underline{0.969}\\\underline{(0.950--0.989)}} & \makecell[c]{0.965\\(0.944--0.987)} & \makecell[c]{0.961\\(0.939--0.983)} & \makecell[c]{0.904\\(0.866--0.942)} & \makecell[c]{0.935\\(0.903--0.967)} & \makecell[c]{\textbf{0.971}\\\textbf{(0.952--0.990)}} & \makecell[c]{0.849\\(0.799--0.899)} & \makecell[c]{0.921\\(0.886--0.956)} & \makecell[c]{0.833\\(0.778--0.888)} \\
T4 & \makecell[c]{\textbf{0.929}\\\textbf{(0.842--0.991)}} & \makecell[c]{\underline{0.913}\\\underline{(0.790--1.000)}} & \makecell[c]{0.904\\(0.818--0.972)} & \makecell[c]{0.567\\(0.236--0.921)} & \makecell[c]{0.718\\(0.495--0.913)} & \makecell[c]{0.603\\(0.375--0.840)} & \makecell[c]{0.878\\(0.775--0.965)} & \makecell[c]{0.644\\(0.457--0.838)} & \makecell[c]{0.638\\(0.348--0.895)} \\
T5 & \makecell[c]{\textbf{0.930}\\\textbf{(0.784--1.000)}} & \makecell[c]{\textbf{0.930}\\\textbf{(0.844--1.000)}} & \makecell[c]{0.870\\(0.764--0.976)} & \makecell[c]{0.840\\(0.593--1.000)} & \makecell[c]{0.910\\(0.726--1.000)} & \makecell[c]{\underline{0.920}\\\underline{(0.789--1.000)}} & \makecell[c]{\textbf{0.930}\\\textbf{(0.844--1.000)}} & \makecell[c]{0.910\\(0.817--1.000)} & \makecell[c]{0.900\\(0.812--0.988)} \\
T6 & \makecell[c]{0.867\\(0.812--0.922)} & \makecell[c]{0.786\\(0.715--0.856)} & \makecell[c]{0.819\\(0.753--0.885)} & \makecell[c]{\underline{0.876}\\\underline{(0.824--0.929)}} & \makecell[c]{\textbf{0.955}\\\textbf{(0.928--0.983)}} & \makecell[c]{0.754\\(0.681--0.827)} & \makecell[c]{0.785\\(0.716--0.853)} & \makecell[c]{0.822\\(0.759--0.884)} & \makecell[c]{0.670\\(0.587--0.753)} \\
T7 & \makecell[c]{0.934\\(0.883--0.973)} & \makecell[c]{0.924\\(0.864--0.969)} & \makecell[c]{0.935\\(0.884--0.972)} & \makecell[c]{0.924\\(0.878--0.962)} & \makecell[c]{\textbf{0.984}\\\textbf{(0.961--0.998)}} & \makecell[c]{0.937\\(0.894--0.969)} & \makecell[c]{0.955\\(0.915--0.982)} & \makecell[c]{\underline{0.972}\\\underline{(0.945--0.990)}} & \makecell[c]{0.822\\(0.738--0.894)} \\
T8 & \makecell[c]{0.727\\(0.520--0.935)} & \makecell[c]{0.663\\(0.433--0.893)} & \makecell[c]{0.697\\(0.494--0.900)} & \makecell[c]{\underline{0.754}\\\underline{(0.551--0.957)}} & \makecell[c]{\textbf{0.898}\\\textbf{(0.809--0.987)}} & \makecell[c]{0.742\\(0.570--0.915)} & \makecell[c]{0.708\\(0.461--0.956)} & \makecell[c]{0.606\\(0.347--0.865)} & \makecell[c]{0.701\\(0.402--0.999)} \\
T9 & \makecell[c]{\textbf{0.844}\\\textbf{(0.712--0.950)}} & \makecell[c]{\underline{0.835}\\\underline{(0.724--0.928)}} & \makecell[c]{0.792\\(0.637--0.939)} & \makecell[c]{0.792\\(0.651--0.917)} & \makecell[c]{0.719\\(0.567--0.860)} & \makecell[c]{0.709\\(0.545--0.849)} & \makecell[c]{0.673\\(0.446--0.901)} & \makecell[c]{0.791\\(0.626--0.923)} & \makecell[c]{0.760\\(0.588--0.909)} \\
T10 & \makecell[c]{\textbf{1.000}\\\textbf{(1.000--1.000)}} & \makecell[c]{\textbf{1.000}\\\textbf{(1.000--1.000)}} & \makecell[c]{\textbf{1.000}\\\textbf{(1.000--1.000)}} & \makecell[c]{\textbf{1.000}\\\textbf{(1.000--1.000)}} & \makecell[c]{\textbf{1.000}\\\textbf{(1.000--1.000)}} & \makecell[c]{\underline{0.800}\\\underline{(0.400--1.000)}} & \makecell[c]{\textbf{1.000}\\\textbf{(1.000--1.000)}} & \makecell[c]{\textbf{1.000}\\\textbf{(1.000--1.000)}} & \makecell[c]{\underline{0.800}\\\underline{(0.400--1.000)}} \\
T11 & \makecell[c]{\textbf{0.951}\\\textbf{(0.855--1.000)}} & \makecell[c]{0.858\\(0.780--0.936)} & \makecell[c]{0.515\\(0.200--0.830)} & \makecell[c]{0.873\\(0.769--0.976)} & \makecell[c]{0.925\\(0.863--0.987)} & \makecell[c]{0.938\\(0.878--0.998)} & \makecell[c]{0.725\\(0.516--0.933)} & \makecell[c]{\underline{0.948}\\\underline{(0.896--1.000)}} & \makecell[c]{0.822\\(0.691--0.953)} \\
\hline
\end{tabular}
\par\smallskip
\noindent\parbox{\linewidth}{\SuppTableNotes Values are AUC (95\% CI). Binary-task AUC intervals use DeLong's method and multiclass AUC intervals use 2,000-resample percentile bootstrap; when DeLong's interval is undefined because a class has fewer than two samples, percentile bootstrap is used. ACC intervals use Wilson's method. Bold indicates the highest value within a task; underlining indicates the second-highest value.}
\end{table}
\end{landscape}
\begin{landscape}
\begin{table}[p]
\centering
\caption{Performance across foundation models on biomarker tasks (T12--T23) in the temporal cohort.}
\label{tab:external1_biomarker_auc_acc}
\SuppTableStandard
\setlength{\tabcolsep}{2pt}
\begin{tabular}{@{}lccccccccc@{}}
\hline
Task & FOCUS & H-Optimus-1 & H-Optimus-0 & GigaPath & CRISP & Virchow2 & Virchow & UNI & Phikon-v2 \\
\hline
T12 & \makecell[c]{\textbf{0.841}\\\textbf{(0.770--0.913)}} & \makecell[c]{0.784\\(0.710--0.857)} & \makecell[c]{0.765\\(0.684--0.845)} & \makecell[c]{0.763\\(0.673--0.852)} & \makecell[c]{0.811\\(0.732--0.891)} & \makecell[c]{0.792\\(0.709--0.876)} & \makecell[c]{0.739\\(0.661--0.816)} & \makecell[c]{\underline{0.825}\\\underline{(0.749--0.901)}} & \makecell[c]{0.749\\(0.669--0.829)} \\
T13 & \makecell[c]{\textbf{0.908}\\\textbf{(0.858--0.959)}} & \makecell[c]{0.836\\(0.762--0.909)} & \makecell[c]{0.809\\(0.734--0.885)} & \makecell[c]{\underline{0.884}\\\underline{(0.824--0.944)}} & \makecell[c]{0.847\\(0.780--0.914)} & \makecell[c]{0.844\\(0.775--0.913)} & \makecell[c]{0.829\\(0.757--0.900)} & \makecell[c]{0.865\\(0.802--0.928)} & \makecell[c]{0.808\\(0.733--0.883)} \\
T14 & \makecell[c]{\textbf{0.847}\\\textbf{(0.710--0.985)}} & \makecell[c]{0.722\\(0.546--0.898)} & \makecell[c]{0.716\\(0.517--0.915)} & \makecell[c]{\underline{0.808}\\\underline{(0.675--0.941)}} & \makecell[c]{0.801\\(0.655--0.947)} & \makecell[c]{0.800\\(0.683--0.918)} & \makecell[c]{0.762\\(0.564--0.960)} & \makecell[c]{0.764\\(0.600--0.928)} & \makecell[c]{0.684\\(0.520--0.848)} \\
T15 & \makecell[c]{0.923\\(0.798--1.000)} & \makecell[c]{0.933\\(0.838--1.000)} & \makecell[c]{\textbf{0.961}\\\textbf{(0.911--1.000)}} & \makecell[c]{0.946\\(0.875--1.000)} & \makecell[c]{\underline{0.954}\\\underline{(0.879--1.000)}} & \makecell[c]{0.923\\(0.817--1.000)} & \makecell[c]{0.928\\(0.794--1.000)} & \makecell[c]{0.951\\(0.881--1.000)} & \makecell[c]{0.936\\(0.880--0.991)} \\
T16 & \makecell[c]{\textbf{0.886}\\\textbf{(0.823--0.948)}} & \makecell[c]{0.836\\(0.755--0.916)} & \makecell[c]{0.819\\(0.689--0.949)} & \makecell[c]{\underline{0.850}\\\underline{(0.754--0.946)}} & \makecell[c]{0.847\\(0.725--0.969)} & \makecell[c]{0.803\\(0.675--0.930)} & \makecell[c]{0.785\\(0.690--0.879)} & \makecell[c]{0.803\\(0.692--0.913)} & \makecell[c]{0.763\\(0.638--0.889)} \\
T17 & \makecell[c]{\textbf{0.872}\\\textbf{(0.811--0.933)}} & \makecell[c]{0.833\\(0.768--0.899)} & \makecell[c]{0.826\\(0.755--0.897)} & \makecell[c]{0.830\\(0.759--0.901)} & \makecell[c]{\underline{0.836}\\\underline{(0.769--0.903)}} & \makecell[c]{0.801\\(0.725--0.877)} & \makecell[c]{0.827\\(0.759--0.895)} & \makecell[c]{0.789\\(0.713--0.865)} & \makecell[c]{0.780\\(0.704--0.855)} \\
T18 & \makecell[c]{\textbf{0.859}\\\textbf{(0.806--0.912)}} & \makecell[c]{0.794\\(0.728--0.860)} & \makecell[c]{0.766\\(0.695--0.836)} & \makecell[c]{\underline{0.803}\\\underline{(0.740--0.866)}} & \makecell[c]{0.788\\(0.722--0.854)} & \makecell[c]{0.747\\(0.679--0.816)} & \makecell[c]{0.754\\(0.684--0.824)} & \makecell[c]{0.797\\(0.729--0.865)} & \makecell[c]{0.710\\(0.637--0.783)} \\
T19 & \makecell[c]{\textbf{0.943}\\\textbf{(0.909--0.977)}} & \makecell[c]{0.873\\(0.825--0.922)} & \makecell[c]{0.872\\(0.822--0.922)} & \makecell[c]{0.903\\(0.858--0.947)} & \makecell[c]{\underline{0.913}\\\underline{(0.871--0.954)}} & \makecell[c]{0.906\\(0.864--0.948)} & \makecell[c]{0.858\\(0.806--0.911)} & \makecell[c]{0.901\\(0.857--0.944)} & \makecell[c]{0.817\\(0.759--0.875)} \\
T20 & \makecell[c]{\textbf{0.665}\\\textbf{(0.566--0.765)}} & \makecell[c]{\underline{0.628}\\\underline{(0.522--0.734)}} & \makecell[c]{0.565\\(0.456--0.674)} & \makecell[c]{0.603\\(0.498--0.708)} & \makecell[c]{0.610\\(0.504--0.716)} & \makecell[c]{0.609\\(0.503--0.714)} & \makecell[c]{0.607\\(0.503--0.711)} & \makecell[c]{0.609\\(0.503--0.715)} & \makecell[c]{0.576\\(0.466--0.687)} \\
T21 & \makecell[c]{\textbf{0.898}\\\textbf{(0.845--0.952)}} & \makecell[c]{0.844\\(0.788--0.900)} & \makecell[c]{0.837\\(0.768--0.905)} & \makecell[c]{\underline{0.897}\\\underline{(0.850--0.944)}} & \makecell[c]{0.870\\(0.814--0.927)} & \makecell[c]{0.852\\(0.790--0.915)} & \makecell[c]{0.865\\(0.808--0.923)} & \makecell[c]{0.896\\(0.851--0.941)} & \makecell[c]{0.823\\(0.760--0.886)} \\
T22 & \makecell[c]{\textbf{0.889}\\\textbf{(0.831--0.946)}} & \makecell[c]{0.865\\(0.806--0.924)} & \makecell[c]{0.874\\(0.816--0.931)} & \makecell[c]{\underline{0.875}\\\underline{(0.810--0.941)}} & \makecell[c]{0.864\\(0.802--0.926)} & \makecell[c]{0.843\\(0.781--0.905)} & \makecell[c]{0.847\\(0.774--0.919)} & \makecell[c]{0.842\\(0.775--0.909)} & \makecell[c]{0.806\\(0.735--0.877)} \\
T23 & \makecell[c]{\textbf{0.855}\\\textbf{(0.802--0.909)}} & \makecell[c]{0.803\\(0.741--0.865)} & \makecell[c]{0.777\\(0.711--0.843)} & \makecell[c]{0.817\\(0.757--0.877)} & \makecell[c]{\underline{0.833}\\\underline{(0.775--0.890)}} & \makecell[c]{0.812\\(0.752--0.873)} & \makecell[c]{0.827\\(0.767--0.886)} & \makecell[c]{0.802\\(0.740--0.865)} & \makecell[c]{0.779\\(0.714--0.843)} \\
\hline
\end{tabular}
\par\smallskip
\noindent\parbox{\linewidth}{\SuppTableNotes Values are AUC (95\% CI). Binary-task AUC intervals use DeLong's method and multiclass AUC intervals use 2,000-resample percentile bootstrap; when DeLong's interval is undefined because a class has fewer than two samples, percentile bootstrap is used. ACC intervals use Wilson's method. Bold indicates the highest value within a task; underlining indicates the second-highest value.}
\end{table}
\end{landscape}
\begin{landscape}
\begin{table}[p]
\centering
\caption{Performance across foundation models on biomarker tasks (T12--T23) in the external cohort.}
\label{tab:external2_biomarker_auc_acc}
\SuppTableStandard
\setlength{\tabcolsep}{2pt}
\begin{tabular}{@{}lccccccccc@{}}
\hline
Task & FOCUS & H-Optimus-1 & H-Optimus-0 & GigaPath & CRISP & Virchow2 & Virchow & UNI & Phikon-v2 \\
\hline
T12 & \makecell[c]{\textbf{1.000}\\\textbf{(1.000--1.000)}} & \makecell[c]{0.917\\(0.773--1.000)} & \makecell[c]{\underline{0.979}\\\underline{(0.921--1.000)}} & \makecell[c]{0.583\\(0.000--1.000)} & \makecell[c]{0.771\\(0.467--1.000)} & \makecell[c]{0.896\\(0.743--1.000)} & \makecell[c]{0.938\\(0.815--1.000)} & \makecell[c]{0.854\\(0.556--1.000)} & \makecell[c]{\underline{0.979}\\\underline{(0.921--1.000)}} \\
T13 & \makecell[c]{\textbf{1.000}\\\textbf{(1.000--1.000)}} & \makecell[c]{\underline{0.926}\\\underline{(0.798--1.000)}} & \makecell[c]{\underline{0.926}\\\underline{(0.803--1.000)}} & \makecell[c]{0.907\\(0.775--1.000)} & \makecell[c]{0.704\\(0.219--1.000)} & \makecell[c]{0.722\\(0.479--0.965)} & \makecell[c]{\textbf{1.000}\\\textbf{(1.000--1.000)}} & \makecell[c]{0.611\\(0.319--0.904)} & \makecell[c]{0.593\\(0.195--0.991)} \\
T14 & \makecell[c]{\underline{0.316}\\\underline{(0.154--0.540)}} & \makecell[c]{0.105\\(0.029--0.314)} & \makecell[c]{0.158\\(0.055--0.376)} & \makecell[c]{0.000\\(0.000--0.168)} & \makecell[c]{0.053\\(0.009--0.246)} & \makecell[c]{0.000\\(0.000--0.168)} & \makecell[c]{0.053\\(0.009--0.246)} & \makecell[c]{0.105\\(0.029--0.314)} & \makecell[c]{\textbf{1.000}\\\textbf{(0.832--1.000)}} \\
T15 & \makecell[c]{\textbf{1.000}\\\textbf{(0.832--1.000)}} & \makecell[c]{\textbf{1.000}\\\textbf{(0.832--1.000)}} & \makecell[c]{\textbf{1.000}\\\textbf{(0.832--1.000)}} & \makecell[c]{\textbf{1.000}\\\textbf{(0.832--1.000)}} & \makecell[c]{\textbf{1.000}\\\textbf{(0.832--1.000)}} & \makecell[c]{\textbf{1.000}\\\textbf{(0.832--1.000)}} & \makecell[c]{\textbf{1.000}\\\textbf{(0.832--1.000)}} & \makecell[c]{\textbf{1.000}\\\textbf{(0.832--1.000)}} & \makecell[c]{\textbf{1.000}\\\textbf{(0.832--1.000)}} \\
T16 & \makecell[c]{\textbf{1.000}\\\textbf{(1.000--1.000)}} & \makecell[c]{\underline{0.950}\\\underline{(0.842--1.000)}} & \makecell[c]{0.850\\(0.684--1.000)} & \makecell[c]{0.700\\(0.500--0.900)} & \makecell[c]{\textbf{1.000}\\\textbf{(1.000--1.000)}} & \makecell[c]{\underline{0.950}\\\underline{(0.842--1.000)}} & \makecell[c]{0.600\\(0.368--0.800)} & \makecell[c]{0.600\\(0.400--0.800)} & \makecell[c]{0.500\\(0.299--0.737)} \\
T17 & \makecell[c]{\textbf{1.000}\\\textbf{(1.000--1.000)}} & \makecell[c]{0.958\\(0.867--1.000)} & \makecell[c]{\underline{0.979}\\\underline{(0.921--1.000)}} & \makecell[c]{0.917\\(0.773--1.000)} & \makecell[c]{\underline{0.979}\\\underline{(0.921--1.000)}} & \makecell[c]{0.896\\(0.677--1.000)} & \makecell[c]{0.896\\(0.677--1.000)} & \makecell[c]{0.875\\(0.617--1.000)} & \makecell[c]{0.917\\(0.738--1.000)} \\
T18 & \makecell[c]{\textbf{0.920}\\\textbf{(0.793--1.000)}} & \makecell[c]{0.852\\(0.671--1.000)} & \makecell[c]{\underline{0.909}\\\underline{(0.762--1.000)}} & \makecell[c]{0.523\\(0.234--0.812)} & \makecell[c]{0.716\\(0.428--1.000)} & \makecell[c]{0.705\\(0.462--0.947)} & \makecell[c]{0.864\\(0.695--1.000)} & \makecell[c]{0.841\\(0.653--1.000)} & \makecell[c]{0.818\\(0.600--1.000)} \\
T19 & \makecell[c]{\textbf{0.990}\\\textbf{(0.962--1.000)}} & \makecell[c]{0.878\\(0.723--1.000)} & \makecell[c]{\underline{0.908}\\\underline{(0.780--1.000)}} & \makecell[c]{0.633\\(0.368--0.897)} & \makecell[c]{0.806\\(0.560--1.000)} & \makecell[c]{0.806\\(0.545--1.000)} & \makecell[c]{0.878\\(0.705--1.000)} & \makecell[c]{0.878\\(0.723--1.000)} & \makecell[c]{0.806\\(0.545--1.000)} \\
T20 & \makecell[c]{\underline{0.702}\\\underline{(0.461--0.943)}} & \makecell[c]{0.606\\(0.328--0.884)} & \makecell[c]{\textbf{0.788}\\\textbf{(0.587--0.990)}} & \makecell[c]{0.577\\(0.299--0.855)} & \makecell[c]{0.510\\(0.216--0.803)} & \makecell[c]{0.423\\(0.122--0.724)} & \makecell[c]{0.663\\(0.418--0.909)} & \makecell[c]{0.433\\(0.162--0.704)} & \makecell[c]{0.615\\(0.348--0.883)} \\
T21 & \makecell[c]{0.657\\(0.388--0.926)} & \makecell[c]{0.700\\(0.415--0.985)} & \makecell[c]{\textbf{0.886}\\\textbf{(0.730--1.000)}} & \makecell[c]{0.686\\(0.426--0.946)} & \makecell[c]{0.743\\(0.449--1.000)} & \makecell[c]{0.771\\(0.483--1.000)} & \makecell[c]{0.729\\(0.458--0.999)} & \makecell[c]{\underline{0.786}\\\underline{(0.526--1.000)}} & \makecell[c]{0.557\\(0.188--0.926)} \\
T22 & \makecell[c]{\underline{0.882}\\\underline{(0.706--1.000)}} & \makecell[c]{0.471\\(0.000--1.000)} & \makecell[c]{0.647\\(0.054--1.000)} & \makecell[c]{0.529\\(0.000--1.000)} & \makecell[c]{0.765\\(0.478--1.000)} & \makecell[c]{0.794\\(0.597--0.992)} & \makecell[c]{0.824\\(0.552--1.000)} & \makecell[c]{\textbf{0.912}\\\textbf{(0.715--1.000)}} & \makecell[c]{\underline{0.882}\\\underline{(0.724--1.000)}} \\
T23 & \makecell[c]{\textbf{0.873}\\\textbf{(0.697--1.000)}} & \makecell[c]{0.727\\(0.502--0.952)} & \makecell[c]{\underline{0.836}\\\underline{(0.661--1.000)}} & \makecell[c]{0.718\\(0.468--0.969)} & \makecell[c]{0.809\\(0.599--1.000)} & \makecell[c]{0.727\\(0.499--0.956)} & \makecell[c]{0.700\\(0.433--0.967)} & \makecell[c]{0.755\\(0.523--0.986)} & \makecell[c]{0.736\\(0.507--0.966)} \\
\hline
\end{tabular}
\par\smallskip
\noindent\parbox{\linewidth}{\SuppTableNotes Values are AUC (95\% CI). Values for T14, T15 are ACC because only one class was observed in this cohort. Binary-task AUC intervals use DeLong's method and multiclass AUC intervals use 2,000-resample percentile bootstrap; when DeLong's interval is undefined because a class has fewer than two samples, percentile bootstrap is used. ACC intervals use Wilson's method. Bold indicates the highest value within a task; underlining indicates the second-highest value.}
\end{table}
\end{landscape}


\endgroup 
\begingroup
\renewcommand{\arraystretch}{1.16}
\begin{table}[t]
\centering
\caption{Histomorphological characteristics of the ten shared patch prototypes.}
\label{tab:prototype-morphology}
\small
\begin{tabular}{@{}p{0.12\textwidth}p{0.80\textwidth}@{}}
\toprule
Prototype & Predominant visual characteristics \\
\midrule
P00 & Preparation-related artefact. \\
P01 & Predominantly blank field with minimal tissue content. \\
P02 & Relatively monomorphic tumour cells accompanied by conspicuous branching vessels. \\
P03 & Dispersed tumour cells with marked cytological atypia. \\
P04 & Blurring and artefactual distortion that limit assessment of cellular detail. \\
P05 & Pronounced cytological atypia and conspicuous mitotic figures, with endothelial proliferation. \\
P06 & Relatively monomorphic tumour cells and branching vessels, similar to P02. \\
P07 & Dispersed tumour cells with marked cytological atypia and oval vascular profiles, resembling the pattern in P03. \\
P08 & Low tumour cellularity with a perineuronal distribution in a cortex-adjacent tissue pattern. \\
P09 & Low-cellularity, perineuronal tumour distribution in a more cortical-appearing background than P08. \\
\bottomrule
\end{tabular}
\par\smallskip
\noindent\parbox{0.92\textwidth}{\footnotesize Descriptions summarize visual features of representative patches. P00, P01 and P04 primarily capture artefact or limited tissue content and should not be interpreted as specific tumour phenotypes. Prototype labels do not constitute independent histopathological diagnoses.}
\end{table}
\endgroup


\begingroup 
\renewcommand{\arraystretch}{1.18}
\begin{landscape}
\begin{table}[p]
\centering
\caption{Performance of nine foundation models in the lymphoma differentiation cohort ($n=102$ slides; 81 lymphoma and 21 non-lymphoma).}
\label{tab:lymphoma_differentiation_metrics}
\SuppTableStandard
\setlength{\tabcolsep}{2pt}
\begin{tabular}{@{}lccccccccc@{}}
\hline
Metric & FOCUS & H-Optimus-1 & H-Optimus-0 & GigaPath & CRISP & Virchow2 & Virchow & UNI & Phikon-v2 \\
\hline
AUC & \makecell[c]{\textbf{0.967}\\\textbf{(0.936--0.998)}} & \makecell[c]{0.943\\(0.900--0.986)} & \makecell[c]{0.899\\(0.832--0.965)} & \makecell[c]{0.924\\(0.869--0.978)} & \makecell[c]{0.935\\(0.887--0.982)} & \makecell[c]{0.907\\(0.846--0.967)} & \makecell[c]{0.904\\(0.832--0.975)} & \makecell[c]{\underline{0.956}\\\underline{(0.920--0.992)}} & \makecell[c]{0.778\\(0.689--0.868)} \\
Accuracy & \makecell[c]{\textbf{0.902}\\\textbf{(0.829--0.946)}} & \makecell[c]{\underline{0.892}\\\underline{(0.817--0.939)}} & \makecell[c]{0.873\\(0.794--0.924)} & \makecell[c]{0.833\\(0.749--0.893)} & \makecell[c]{\underline{0.892}\\\underline{(0.817--0.939)}} & \makecell[c]{0.833\\(0.749--0.893)} & \makecell[c]{0.814\\(0.727--0.877)} & \makecell[c]{0.843\\(0.760--0.901)} & \makecell[c]{0.794\\(0.706--0.861)} \\
F1 score & \makecell[c]{\textbf{0.940}\\\textbf{(0.897--0.972)}} & \makecell[c]{0.930\\(0.884--0.968)} & \makecell[c]{0.923\\(0.876--0.961)} & \makecell[c]{0.904\\(0.854--0.946)} & \makecell[c]{\underline{0.933}\\\underline{(0.889--0.970)}} & \makecell[c]{0.902\\(0.848--0.944)} & \makecell[c]{0.869\\(0.806--0.921)} & \makecell[c]{0.908\\(0.855--0.949)} & \makecell[c]{0.885\\(0.834--0.932)} \\
\hline
\end{tabular}
\par\smallskip
\noindent\parbox{\linewidth}{\SuppTableNotes Values are estimates (95\% CI).  Each model's classification threshold maximizes Youden's J on validation predictions. The thresholds are fixed when applied to this cohort. AUC intervals use DeLong's method; accuracy intervals use Wilson's method; F1 intervals use 2,000 slide-level bootstrap resamples. Bold indicates the highest value for each metric; underlining indicates the second-highest value.}
\end{table}
\end{landscape}


\renewcommand{\arraystretch}{1.18}
\begin{table}[htbp]
\centering
\caption{Specificity at 95\% sensitivity for T1 across three cohorts and T11 in pooled cohorts.}
\label{tab:task1_three_cohorts_task12_pooled}
\SuppTableCompact
\setlength{\tabcolsep}{3pt}
\begin{tabular}{@{}lcccc@{}}
\hline
Model & T1 internal & T1 temporal & T1 external & T11 pooled \\
\hline
FOCUS & \textbf{0.880 (0.800--0.980)} & \textbf{0.862 (0.448--0.966)} & \textbf{0.778 (0.556--1.000)} & \textbf{0.813 (0.689--0.957)} \\
H-Optimus-1 & \underline{0.833 (0.741--0.944)} & \underline{0.759 (0.552--0.897)} & \textbf{0.778 (0.444--1.000)} & 0.680 (0.571--0.861) \\
H-Optimus-0 & \underline{0.833 (0.741--0.944)} & \underline{0.759 (0.586--0.931)} & \underline{0.667 (0.333--1.000)} & 0.688 (0.599--0.860) \\
GigaPath & 0.815 (0.704--0.907) & 0.724 (0.517--0.897) & \underline{0.667 (0.333--1.000)} & \underline{0.803 (0.538--0.942)} \\
CRISP & 0.796 (0.667--0.907) & 0.621 (0.448--0.862) & \textbf{0.778 (0.444--1.000)} & 0.724 (0.536--0.930) \\
Virchow2 & \underline{0.833 (0.722--0.926)} & 0.724 (0.517--0.931) & \underline{0.667 (0.333--1.000)} & 0.786 (0.638--0.917) \\
Virchow & 0.815 (0.704--0.908) & 0.690 (0.448--0.862) & \textbf{0.778 (0.444--1.000)} & 0.651 (0.471--0.820) \\
UNI & 0.815 (0.685--0.944) & 0.724 (0.483--0.862) & \underline{0.667 (0.333--1.000)} & 0.665 (0.370--0.904) \\
Phikon-v2 & 0.815 (0.703--0.926) & 0.690 (0.345--0.862) & \underline{0.667 (0.333--1.000)} & 0.584 (0.467--0.750) \\
\hline
\end{tabular}
\par\smallskip
\noindent\parbox{\textwidth}{\SuppTableNotes Values are specificity (95\% CI) at an empirical ROC operating point with at least 95\% sensitivity for T1 and at least 95\% lymphoma sensitivity for T11. Intervals use 2,000 percentile bootstrap resamples, stratified by class for T1 and by cohort and class for T11. Operating points are selected on the evaluation data. Bold indicates the highest value and underlining the second-highest value in each column.}
\end{table}


\renewcommand{\arraystretch}{1.16}
\begin{table}[htbp]
\centering
\caption{Level-wise H-F1 for the Test dataset.}
\label{tab:dsmil_levelwise_h_f1}
\SuppTableStandard
\setlength{\tabcolsep}{4pt}
\begin{tabular}{@{}lcccc@{}}
\hline
Model & Level 1 & Level 2 & Level 3 & Level 4 \\
\hline
\textbf{FOCUS} & \underline{0.933 (0.911--0.951)} & 0.931 (0.912--0.947) & \textbf{0.905 (0.884--0.921)} & \textbf{0.865 (0.845--0.882)} \\
H-Optimus-1 & 0.922 (0.901--0.942) & 0.923 (0.902--0.940) & 0.891 (0.871--0.909) & 0.848 (0.828--0.866) \\
H-Optimus-0 & 0.922 (0.898--0.942) & 0.921 (0.900--0.940) & 0.887 (0.867--0.907) & 0.844 (0.823--0.864) \\
GigaPath & 0.911 (0.887--0.933) & 0.913 (0.890--0.930) & 0.881 (0.859--0.899) & 0.842 (0.820--0.861) \\
CRISP & 0.899 (0.878--0.924) & 0.907 (0.887--0.927) & 0.871 (0.850--0.891) & 0.830 (0.810--0.850) \\
Virchow2 & \textbf{0.937 (0.916--0.956)} & \textbf{0.934 (0.915--0.950)} & \underline{0.901 (0.881--0.918)} & \underline{0.858 (0.839--0.876)} \\
Virchow & \textbf{0.937 (0.917--0.957)} & \underline{0.932 (0.915--0.948)} & 0.894 (0.875--0.911) & 0.849 (0.830--0.866) \\
UNI & 0.925 (0.904--0.945) & 0.923 (0.903--0.939) & 0.889 (0.867--0.905) & 0.849 (0.828--0.867) \\
Phikon-v2 & 0.922 (0.899--0.942) & 0.917 (0.897--0.934) & 0.873 (0.853--0.892) & 0.819 (0.799--0.839) \\
\hline
\end{tabular}
\par\smallskip
\noindent\parbox{\textwidth}{\SuppTableNotes Values are point estimate (95\% CI). Confidence intervals were estimated by bootstrap resampling. Bold indicates the highest value for each level; underlining indicates the second-highest value.}
\end{table}
\begin{table}[htbp]
\centering
\caption{Level-wise depth-weighted H-F1 (DH-F1) for the Test dataset.}
\label{tab:dsmil_levelwise_dh_f1}
\SuppTableStandard
\setlength{\tabcolsep}{4pt}
\begin{tabular}{@{}lcccc@{}}
\hline
Model & Level 1 & Level 2 & Level 3 & Level 4 \\
\hline
\textbf{FOCUS} & \underline{0.933 (0.911--0.951)} & \underline{0.931 (0.911--0.947)} & \textbf{0.890 (0.869--0.908)} & \textbf{0.825 (0.801--0.847)} \\
H-Optimus-1 & 0.922 (0.901--0.942) & 0.923 (0.903--0.940) & 0.873 (0.852--0.893) & 0.804 (0.781--0.826) \\
H-Optimus-0 & 0.922 (0.898--0.942) & 0.921 (0.900--0.939) & 0.869 (0.846--0.891) & 0.800 (0.776--0.824) \\
GigaPath & 0.911 (0.887--0.933) & 0.913 (0.891--0.931) & 0.864 (0.840--0.884) & 0.803 (0.778--0.823) \\
CRISP & 0.899 (0.878--0.924) & 0.909 (0.890--0.929) & 0.853 (0.830--0.875) & 0.788 (0.765--0.810) \\
Virchow2 & \textbf{0.937 (0.916--0.956)} & \textbf{0.932 (0.913--0.949)} & \underline{0.882 (0.860--0.901)} & \underline{0.814 (0.791--0.836)} \\
Virchow & \textbf{0.937 (0.917--0.957)} & 0.931 (0.913--0.947) & 0.873 (0.851--0.893) & 0.801 (0.778--0.822) \\
UNI & 0.925 (0.904--0.945) & 0.922 (0.902--0.938) & 0.870 (0.846--0.889) & 0.808 (0.783--0.829) \\
Phikon-v2 & 0.922 (0.899--0.942) & 0.915 (0.895--0.932) & 0.848 (0.824--0.870) & 0.765 (0.739--0.788) \\
\hline
\end{tabular}
\par\smallskip
\noindent\parbox{\textwidth}{\SuppTableNotes Values are point estimate (95\% CI). Confidence intervals were estimated by bootstrap resampling. Bold indicates the highest value for each level; underlining indicates the second-highest value.}
\end{table}
\begin{table}[htbp]
\centering
\caption{Level-wise Accuracy for the Test dataset.}
\label{tab:dsmil_levelwise_accuracy}
\SuppTableStandard
\setlength{\tabcolsep}{4pt}
\begin{tabular}{@{}lcccc@{}}
\hline
Model & Level 1 & Level 2 & Level 3 & Level 4 \\
\hline
\textbf{FOCUS} & \underline{0.933 (0.911--0.951)} & \underline{0.905 (0.878--0.927)} & \textbf{0.830 (0.800--0.856)} & \textbf{0.702 (0.667--0.736)} \\
H-Optimus-1 & 0.922 (0.901--0.942) & 0.895 (0.867--0.917) & 0.804 (0.772--0.832) & 0.668 (0.632--0.705) \\
H-Optimus-0 & 0.922 (0.898--0.942) & 0.891 (0.864--0.916) & 0.798 (0.765--0.830) & 0.659 (0.619--0.697) \\
GigaPath & 0.911 (0.887--0.933) & 0.881 (0.852--0.905) & 0.789 (0.757--0.818) & 0.665 (0.627--0.699) \\
CRISP & 0.899 (0.878--0.924) & 0.870 (0.844--0.898) & 0.757 (0.722--0.789) & 0.610 (0.570--0.647) \\
Virchow2 & \textbf{0.937 (0.916--0.956)} & \textbf{0.907 (0.881--0.930)} & \underline{0.817 (0.784--0.846)} & \underline{0.679 (0.642--0.717)} \\
Virchow & \textbf{0.937 (0.917--0.957)} & 0.904 (0.879--0.927) & 0.801 (0.771--0.832) & 0.651 (0.613--0.687) \\
UNI & 0.925 (0.904--0.945) & 0.891 (0.864--0.913) & 0.798 (0.765--0.826) & 0.673 (0.635--0.710) \\
Phikon-v2 & 0.922 (0.899--0.942) & 0.882 (0.856--0.905) & 0.763 (0.728--0.795) & 0.586 (0.546--0.622) \\
\hline
\end{tabular}
\par\smallskip
\noindent\parbox{\textwidth}{\SuppTableNotes Values are point estimate (95\% CI). Confidence intervals were estimated by bootstrap resampling. Bold indicates the highest value for each level; underlining indicates the second-highest value.}
\end{table}

\clearpage

\renewcommand{\arraystretch}{1.16}
\begin{table}[htbp]
\centering
\caption{Level-wise H-F1 for the Temporal Validation Cohort.}
\label{tab:dsmil_h_f1_result_external1_2e-5_adam_1e-5_binary}
\SuppTableStandard
\setlength{\tabcolsep}{4pt}
\begin{tabular}{@{}lcccc@{}}
\hline
Model & Level 1 & Level 2 & Level 3 & Level 4 \\
\hline
\textbf{FOCUS} & \textbf{0.926 (0.893--0.955)} & \textbf{0.904 (0.874--0.930)} & \textbf{0.863 (0.831--0.893)} & \textbf{0.816 (0.785--0.845)} \\
H-Optimus-1 & 0.893 (0.858--0.926) & 0.864 (0.832--0.894) & 0.820 (0.785--0.853) & 0.764 (0.731--0.797) \\
H-Optimus-0 & \underline{0.909 (0.877--0.942)} & \underline{0.888 (0.858--0.918)} & \underline{0.850 (0.817--0.882)} & 0.787 (0.754--0.818) \\
GigaPath & 0.867 (0.829--0.903) & 0.853 (0.818--0.885) & 0.814 (0.779--0.847) & 0.757 (0.721--0.790) \\
CRISP & 0.884 (0.848--0.916) & 0.880 (0.849--0.909) & 0.842 (0.811--0.871) & \underline{0.787 (0.759--0.816)} \\
Virchow2 & 0.906 (0.874--0.935) & 0.862 (0.832--0.893) & 0.795 (0.761--0.826) & 0.754 (0.719--0.786) \\
Virchow & 0.896 (0.861--0.929) & 0.795 (0.758--0.830) & 0.729 (0.690--0.769) & 0.685 (0.647--0.724) \\
UNI & 0.893 (0.858--0.926) & 0.880 (0.848--0.908) & 0.832 (0.798--0.861) & 0.784 (0.748--0.816) \\
Phikon-v2 & 0.896 (0.861--0.929) & 0.880 (0.849--0.910) & 0.823 (0.790--0.854) & 0.751 (0.722--0.780) \\
\hline
\end{tabular}
\par\smallskip
\noindent\parbox{\textwidth}{\SuppTableNotes Values are point estimate (95\% CI). Confidence intervals were estimated by bootstrap resampling. Bold indicates the highest value for each level; underlining indicates the second-highest value.}
\end{table}
\begin{table}[htbp]
\centering
\caption{Level-wise depth-weighted DH-F1 for the Temporal Validation Cohort.}
\label{tab:dsmil_dh_f1_result_external1_2e-5_adam_1e-5_binary}
\SuppTableStandard
\setlength{\tabcolsep}{4pt}
\begin{tabular}{@{}lcccc@{}}
\hline
Model & Level 1 & Level 2 & Level 3 & Level 4 \\
\hline
\textbf{FOCUS} & \textbf{0.926 (0.893--0.955)} & \textbf{0.896 (0.864--0.924)} & \textbf{0.836 (0.799--0.870)} & \textbf{0.761 (0.724--0.796)} \\
H-Optimus-1 & 0.893 (0.858--0.926) & 0.853 (0.820--0.886) & 0.790 (0.751--0.828) & 0.699 (0.662--0.738) \\
H-Optimus-0 & \underline{0.909 (0.877--0.942)} & \underline{0.881 (0.849--0.912)} & \underline{0.825 (0.786--0.861)} & 0.721 (0.682--0.758) \\
GigaPath & 0.867 (0.829--0.903) & 0.848 (0.812--0.882) & 0.789 (0.749--0.827) & 0.694 (0.653--0.731) \\
CRISP & 0.884 (0.848--0.916) & 0.879 (0.847--0.909) & 0.822 (0.787--0.855) & \underline{0.731 (0.696--0.766)} \\
Virchow2 & 0.906 (0.874--0.935) & 0.846 (0.813--0.879) & 0.749 (0.707--0.787) & 0.688 (0.647--0.728) \\
Virchow & 0.896 (0.861--0.929) & 0.758 (0.717--0.799) & 0.673 (0.628--0.720) & 0.608 (0.566--0.653) \\
UNI & 0.893 (0.858--0.926) & 0.875 (0.841--0.906) & 0.802 (0.764--0.836) & 0.725 (0.685--0.763) \\
Phikon-v2 & 0.896 (0.861--0.929) & 0.873 (0.842--0.904) & 0.789 (0.753--0.824) & 0.673 (0.638--0.707) \\
\hline
\end{tabular}
\par\smallskip
\noindent\parbox{\textwidth}{\SuppTableNotes Values are point estimate (95\% CI). Confidence intervals were estimated by bootstrap resampling. Bold indicates the highest value for each level; underlining indicates the second-highest value.}
\end{table}
\begin{table}[htbp]
\centering
\caption{Level-wise Accuracy for the Temporal Validation Cohort.}
\label{tab:dsmil_accuracy_result_external1_2e-5_adam_1e-5_binary}
\SuppTableStandard
\setlength{\tabcolsep}{4pt}
\begin{tabular}{@{}lcccc@{}}
\hline
Model & Level 1 & Level 2 & Level 3 & Level 4 \\
\hline
\textbf{FOCUS} & \textbf{0.926 (0.893--0.955)} & \textbf{0.854 (0.809--0.893)} & \textbf{0.738 (0.689--0.786)} & \textbf{0.576 (0.521--0.631)} \\
H-Optimus-1 & 0.893 (0.858--0.926) & 0.799 (0.754--0.845) & 0.683 (0.631--0.731) & 0.498 (0.443--0.553) \\
H-Optimus-0 & \underline{0.909 (0.877--0.942)} & \underline{0.835 (0.790--0.877)} & \underline{0.725 (0.670--0.777)} & 0.518 (0.460--0.576) \\
GigaPath & 0.867 (0.829--0.903) & 0.793 (0.748--0.838) & 0.676 (0.625--0.728) & 0.492 (0.434--0.547) \\
CRISP & 0.884 (0.848--0.916) & 0.825 (0.780--0.867) & 0.696 (0.647--0.748) & 0.485 (0.430--0.541) \\
Virchow2 & 0.906 (0.874--0.935) & 0.790 (0.744--0.832) & 0.625 (0.570--0.676) & 0.518 (0.463--0.573) \\
Virchow & 0.896 (0.861--0.929) & 0.670 (0.615--0.725) & 0.570 (0.515--0.621) & 0.443 (0.392--0.498) \\
UNI & 0.893 (0.858--0.926) & 0.829 (0.783--0.867) & 0.686 (0.634--0.732) & \underline{0.550 (0.492--0.605)} \\
Phikon-v2 & 0.896 (0.861--0.929) & 0.822 (0.780--0.864) & 0.657 (0.605--0.709) & 0.421 (0.369--0.476) \\
\hline
\end{tabular}
\par\smallskip
\noindent\parbox{\textwidth}{\SuppTableNotes Values are point estimate (95\% CI). Confidence intervals were estimated by bootstrap resampling. Bold indicates the highest value for each level; underlining indicates the second-highest value.}
\end{table}

\renewcommand{\arraystretch}{1.16}
\begin{table}[htbp]
\centering
\caption{Level-wise H-F1 for the External Validation Cohort.}
\label{tab:dsmil_h_f1_result_external2_2e-5_adam_1e-5_binary}
\SuppTableStandard
\setlength{\tabcolsep}{4pt}
\begin{tabular}{@{}lcccc@{}}
\hline
Model & Level 1 & Level 2 & Level 3 & Level 4 \\
\hline
\textbf{FOCUS} & \underline{0.911 (0.874--0.943)} & \textbf{0.895 (0.862--0.926)} & \textbf{0.825 (0.791--0.857)} & \textbf{0.772 (0.737--0.805)} \\
H-Optimus-1 & \textbf{0.915 (0.878--0.947)} & 0.826 (0.791--0.859) & 0.706 (0.669--0.741) & 0.680 (0.640--0.717) \\
H-Optimus-0 & \underline{0.911 (0.874--0.943)} & \underline{0.877 (0.847--0.913)} & 0.781 (0.747--0.813) & 0.729 (0.696--0.763) \\
GigaPath & \underline{0.911 (0.874--0.943)} & 0.855 (0.821--0.887) & 0.733 (0.696--0.766) & 0.687 (0.651--0.719) \\
CRISP & \underline{0.911 (0.874--0.943)} & 0.865 (0.831--0.897) & \underline{0.787 (0.752--0.821)} & \underline{0.743 (0.708--0.777)} \\
Virchow2 & \underline{0.911 (0.874--0.943)} & 0.869 (0.837--0.901) & 0.717 (0.684--0.749) & 0.679 (0.646--0.711) \\
Virchow & \underline{0.911 (0.874--0.943)} & 0.817 (0.781--0.852) & 0.677 (0.640--0.713) & 0.652 (0.613--0.688) \\
UNI & \textbf{0.915 (0.878--0.947)} & 0.808 (0.772--0.842) & 0.729 (0.692--0.766) & 0.678 (0.641--0.714) \\
Phikon-v2 & \underline{0.911 (0.874--0.943)} & 0.830 (0.795--0.864) & 0.651 (0.617--0.684) & 0.619 (0.584--0.650) \\
\hline
\end{tabular}
\par\smallskip
\noindent\parbox{\textwidth}{\SuppTableNotes Values are point estimate (95\% CI). Confidence intervals were estimated by bootstrap resampling. Bold indicates the highest value for each level; underlining indicates the second-highest value.}
\end{table}
\begin{table}[htbp]
\centering
\caption{Level-wise DH-F1 for the External Validation Cohort.}
\label{tab:dsmil_dh_f1_result_external2_2e-5_adam_1e-5_binary}
\SuppTableStandard
\setlength{\tabcolsep}{4pt}
\begin{tabular}{@{}lcccc@{}}
\hline
Model & Level 1 & Level 2 & Level 3 & Level 4 \\
\hline
\textbf{FOCUS} & \underline{0.911 (0.874--0.943)} & \textbf{0.889 (0.856--0.922)} & \textbf{0.784 (0.745--0.819)} & \textbf{0.690 (0.650--0.729)} \\
H-Optimus-1 & \textbf{0.915 (0.878--0.947)} & 0.796 (0.756--0.834) & 0.624 (0.579--0.666) & 0.578 (0.531--0.628) \\
H-Optimus-0 & \underline{0.911 (0.874--0.943)} & \underline{0.866 (0.834--0.903)} & 0.722 (0.681--0.760) & 0.632 (0.593--0.670) \\
GigaPath & \underline{0.911 (0.874--0.943)} & 0.835 (0.798--0.872) & 0.656 (0.613--0.695) & 0.580 (0.538--0.618) \\
CRISP & \underline{0.911 (0.874--0.943)} & 0.849 (0.812--0.884) & \underline{0.736 (0.694--0.777)} & \underline{0.659 (0.617--0.699)} \\
Virchow2 & \underline{0.911 (0.874--0.943)} & 0.855 (0.820--0.890) & 0.626 (0.587--0.664) & 0.563 (0.524--0.599) \\
Virchow & \underline{0.911 (0.874--0.943)} & 0.785 (0.743--0.823) & 0.584 (0.541--0.627) & 0.542 (0.496--0.587) \\
UNI & \textbf{0.915 (0.878--0.947)} & 0.771 (0.728--0.810) & 0.667 (0.623--0.711) & 0.583 (0.539--0.626) \\
Phikon-v2 & \underline{0.911 (0.874--0.943)} & 0.802 (0.763--0.841) & 0.539 (0.498--0.577) & 0.488 (0.449--0.524) \\
\hline
\end{tabular}
\par\smallskip
\noindent\parbox{\textwidth}{\SuppTableNotes Values are point estimate (95\% CI). Confidence intervals were estimated by bootstrap resampling. Bold indicates the highest value for each level; underlining indicates the second-highest value.}
\end{table}
\begin{table}[htbp]
\centering
\caption{Level-wise Accuracy for the External Validation Cohort.}
\label{tab:dsmil_accuracy_result_external2_2e-5_adam_1e-5_binary}
\SuppTableStandard
\setlength{\tabcolsep}{4pt}
\begin{tabular}{@{}lcccc@{}}
\hline
Model & Level 1 & Level 2 & Level 3 & Level 4 \\
\hline
\textbf{FOCUS} & \underline{0.911 (0.874--0.943)} & \textbf{0.837 (0.789--0.886)} & \textbf{0.553 (0.484--0.610)} & \textbf{0.411 (0.346--0.468)} \\
H-Optimus-1 & \textbf{0.915 (0.878--0.947)} & 0.703 (0.646--0.756) & 0.358 (0.297--0.419) & 0.346 (0.289--0.402) \\
H-Optimus-0 & \underline{0.911 (0.874--0.943)} & \underline{0.805 (0.760--0.858)} & 0.463 (0.402--0.524) & 0.325 (0.268--0.382) \\
GigaPath & \underline{0.911 (0.874--0.943)} & 0.760 (0.707--0.813) & 0.374 (0.313--0.431) & 0.264 (0.207--0.317) \\
CRISP & \underline{0.911 (0.874--0.943)} & 0.781 (0.728--0.829) & \underline{0.504 (0.443--0.565)} & \underline{0.382 (0.321--0.439)} \\
Virchow2 & \underline{0.911 (0.874--0.943)} & 0.789 (0.740--0.837) & 0.301 (0.244--0.362) & 0.203 (0.155--0.260) \\
Virchow & \underline{0.911 (0.874--0.943)} & 0.687 (0.626--0.740) & 0.293 (0.236--0.350) & 0.272 (0.215--0.325) \\
UNI & \textbf{0.915 (0.878--0.947)} & 0.667 (0.606--0.724) & 0.459 (0.398--0.520) & 0.342 (0.281--0.398) \\
Phikon-v2 & \underline{0.911 (0.874--0.943)} & 0.711 (0.658--0.768) & 0.195 (0.142--0.244) & 0.142 (0.098--0.187) \\
\hline
\end{tabular}
\par\smallskip
\noindent\parbox{\textwidth}{\SuppTableNotes Values are point estimate (95\% CI). Confidence intervals were estimated by bootstrap resampling. Bold indicates the highest value for each level; underlining indicates the second-highest value.}
\end{table}

\renewcommand{\arraystretch}{1.16}
\begin{table}[htbp]
\centering
\caption{Level-wise H-F1 for the Uncertainty Group.}
\label{tab:dsmil_h_f1_uncertainty_group}
\SuppTableStandard
\setlength{\tabcolsep}{4pt}
\begin{tabular}{@{}lcccc@{}}
\hline
Model & Level 1 & Level 2 & Level 3 & Level 4 \\
\hline
\textbf{FOCUS} & 0.909 (0.844--0.961) & 0.902 (0.844--0.957) & \textbf{0.857 (0.797--0.915)} & \textbf{0.813 (0.753--0.873)} \\
H-Optimus-1 & 0.909 (0.844--0.974) & 0.905 (0.846--0.960) & 0.842 (0.784--0.903) & 0.784 (0.726--0.840) \\
H-Optimus-0 & 0.909 (0.844--0.961) & 0.899 (0.841--0.951) & 0.825 (0.768--0.884) & 0.761 (0.705--0.818) \\
GigaPath & 0.922 (0.857--0.974) & \underline{0.916 (0.860--0.967)} & 0.835 (0.780--0.891) & 0.785 (0.730--0.840) \\
CRISP & \textbf{0.961 (0.909--1.000)} & \textbf{0.944 (0.902--0.981)} & \underline{0.856 (0.807--0.904)} & \underline{0.808 (0.761--0.858)} \\
Virchow2 & \underline{0.935 (0.870--0.987)} & 0.910 (0.856--0.957) & 0.831 (0.772--0.887) & 0.777 (0.715--0.831) \\
Virchow & 0.909 (0.844--0.961) & 0.886 (0.827--0.941) & 0.794 (0.729--0.854) & 0.747 (0.685--0.804) \\
UNI & 0.922 (0.857--0.974) & 0.906 (0.847--0.960) & 0.844 (0.782--0.900) & 0.788 (0.726--0.846) \\
Phikon-v2 & 0.909 (0.844--0.974) & 0.906 (0.851--0.957) & 0.803 (0.744--0.863) & 0.741 (0.685--0.797) \\
\hline
\end{tabular}
\par\smallskip
\noindent\parbox{\textwidth}{\SuppTableNotes Values are point estimate (95\% CI). Confidence intervals were estimated by bootstrap resampling. Bold indicates the highest value for each level; underlining indicates the second-highest value.}
\end{table}

\begin{table}[htbp]
\centering
\caption{Level-wise DH-F1 for the Uncertainty Group.}
\label{tab:dsmil_dh_f1_uncertainty_group}
\SuppTableStandard
\setlength{\tabcolsep}{4pt}
\begin{tabular}{@{}lcccc@{}}
\hline
Model & Level 1 & Level 2 & Level 3 & Level 4 \\
\hline
\textbf{FOCUS} & 0.909 (0.844--0.961) & 0.900 (0.840--0.956) & \textbf{0.831 (0.765--0.899)} & \textbf{0.761 (0.689--0.831)} \\
H-Optimus-1 & 0.909 (0.844--0.974) & 0.904 (0.844--0.960) & \underline{0.808 (0.744--0.878)} & 0.715 (0.647--0.783) \\
H-Optimus-0 & 0.909 (0.844--0.961) & 0.896 (0.837--0.947) & 0.785 (0.717--0.857) & 0.682 (0.615--0.748) \\
GigaPath & 0.922 (0.857--0.974) & \underline{0.914 (0.858--0.965)} & 0.792 (0.728--0.858) & 0.715 (0.646--0.780) \\
CRISP & \textbf{0.961 (0.909--1.000)} & \textbf{0.938 (0.891--0.982)} & 0.808 (0.748--0.869) & \underline{0.736 (0.672--0.801)} \\
Virchow2 & \underline{0.935 (0.870--0.987)} & 0.901 (0.840--0.955) & 0.785 (0.719--0.854) & 0.701 (0.632--0.768) \\
Virchow & 0.909 (0.844--0.961) & 0.878 (0.815--0.937) & 0.742 (0.668--0.812) & 0.672 (0.596--0.745) \\
UNI & 0.922 (0.857--0.974) & 0.901 (0.839--0.956) & 0.808 (0.735--0.879) & 0.720 (0.649--0.788) \\
Phikon-v2 & 0.909 (0.844--0.974) & 0.905 (0.850--0.959) & 0.749 (0.682--0.826) & 0.655 (0.587--0.725) \\
\hline
\end{tabular}
\par\smallskip
\noindent\parbox{\textwidth}{\SuppTableNotes Values are point estimate (95\% CI). Confidence intervals were estimated by bootstrap resampling. Bold indicates the highest value for each level; underlining indicates the second-highest value.}
\end{table}

\begin{table}[htbp]
\centering
\caption{Level-wise Accuracy for the Uncertainty Group.}
\label{tab:dsmil_accuracy_uncertainty_group}
\SuppTableStandard
\setlength{\tabcolsep}{4pt}
\begin{tabular}{@{}lcccc@{}}
\hline
Model & Level 1 & Level 2 & Level 3 & Level 4 \\
\hline
\textbf{FOCUS} & 0.909 (0.844--0.961) & 0.857 (0.779--0.935) & \textbf{0.727 (0.636--0.831)} & \textbf{0.623 (0.520--0.727)} \\
H-Optimus-1 & 0.909 (0.844--0.974) & 0.857 (0.766--0.935) & 0.675 (0.571--0.779) & 0.520 (0.416--0.623) \\
H-Optimus-0 & 0.909 (0.844--0.961) & 0.844 (0.753--0.922) & 0.636 (0.533--0.753) & 0.455 (0.351--0.558) \\
GigaPath & 0.922 (0.857--0.974) & \underline{0.870 (0.792--0.948)} & 0.636 (0.533--0.740) & 0.507 (0.403--0.610) \\
CRISP & \textbf{0.961 (0.909--1.000)} & \textbf{0.909 (0.844--0.974)} & 0.662 (0.558--0.754) & \underline{0.546 (0.442--0.649)} \\
Virchow2 & \underline{0.935 (0.870--0.987)} & 0.857 (0.779--0.935) & 0.649 (0.546--0.753) & 0.520 (0.416--0.623) \\
Virchow & 0.909 (0.844--0.961) & 0.818 (0.727--0.896) & 0.571 (0.455--0.675) & 0.455 (0.338--0.558) \\
UNI & 0.922 (0.857--0.974) & 0.857 (0.779--0.935) & \underline{0.688 (0.584--0.792)} & \underline{0.546 (0.442--0.649)} \\
Phikon-v2 & 0.909 (0.844--0.974) & 0.857 (0.779--0.935) & 0.558 (0.455--0.675) & 0.390 (0.286--0.507) \\
\hline
\end{tabular}
\par\smallskip
\noindent\parbox{\textwidth}{\SuppTableNotes Values are point estimate (95\% CI). Confidence intervals were estimated by bootstrap resampling. Bold indicates the highest value for each level; underlining indicates the second-highest value.}
\end{table}

\renewcommand{\arraystretch}{1.16}
\begin{table}[htbp]
\centering
\caption{Level-wise H-F1 for the Discordance Group.}
\label{tab:dsmil_h_f1_discordance_group}
\SuppTableStandard
\setlength{\tabcolsep}{4pt}
\begin{tabular}{@{}lcccc@{}}
\hline
Model & Level 1 & Level 2 & Level 3 & Level 4 \\
\hline
\textbf{FOCUS} & \textbf{0.924 (0.864--0.985)} & \textbf{0.879 (0.819--0.935)} & \textbf{0.723 (0.661--0.781)} & \textbf{0.637 (0.583--0.691)} \\
H-Optimus-1 & 0.697 (0.576--0.803) & 0.697 (0.587--0.792) & 0.566 (0.482--0.646) & 0.498 (0.421--0.569) \\
H-Optimus-0 & 0.712 (0.591--0.818) & 0.703 (0.603--0.797) & 0.568 (0.486--0.650) & 0.506 (0.433--0.582) \\
GigaPath & 0.697 (0.576--0.818) & 0.700 (0.592--0.795) & 0.571 (0.485--0.652) & 0.510 (0.433--0.584) \\
CRISP & 0.682 (0.561--0.788) & 0.675 (0.564--0.771) & 0.552 (0.467--0.634) & 0.490 (0.411--0.566) \\
Virchow2 & 0.773 (0.667--0.879) & 0.755 (0.661--0.844) & 0.639 (0.551--0.721) & 0.562 (0.486--0.634) \\
Virchow & 0.788 (0.682--0.879) & 0.779 (0.689--0.863) & \underline{0.661 (0.584--0.738)} & \underline{0.586 (0.516--0.652)} \\
UNI & 0.742 (0.636--0.849) & 0.721 (0.619--0.811) & 0.590 (0.508--0.670) & 0.526 (0.447--0.602) \\
Phikon-v2 & \underline{0.803 (0.697--0.909)} & \underline{0.784 (0.694--0.867)} & 0.625 (0.551--0.698) & 0.548 (0.483--0.609) \\
\hline
\end{tabular}
\par\smallskip
\noindent\parbox{\textwidth}{\SuppTableNotes Values are point estimate (95\% CI). Confidence intervals were estimated by bootstrap resampling. Bold indicates the highest value for each level; underlining indicates the second-highest value.}
\end{table}
\begin{table}[htbp]
\centering
\caption{Level-wise DH-F1 for the Discordance Group.}
\label{tab:dsmil_dh_f1_discordance_group}
\SuppTableStandard
\setlength{\tabcolsep}{4pt}
\begin{tabular}{@{}lcccc@{}}
\hline
Model & Level 1 & Level 2 & Level 3 & Level 4 \\
\hline
\textbf{FOCUS} & \textbf{0.924 (0.864--0.985)} & \textbf{0.864 (0.797--0.927)} & \textbf{0.634 (0.561--0.705)} & \textbf{0.504 (0.441--0.568)} \\
H-Optimus-1 & 0.697 (0.576--0.803) & 0.697 (0.587--0.791) & 0.493 (0.414--0.578) & 0.386 (0.321--0.456) \\
H-Optimus-0 & 0.712 (0.591--0.818) & 0.699 (0.598--0.795) & 0.493 (0.411--0.575) & 0.399 (0.329--0.476) \\
GigaPath & 0.697 (0.576--0.818) & 0.701 (0.595--0.794) & 0.501 (0.420--0.583) & 0.408 (0.335--0.484) \\
CRISP & 0.682 (0.561--0.788) & 0.672 (0.560--0.769) & 0.482 (0.399--0.565) & 0.386 (0.315--0.464) \\
Virchow2 & 0.773 (0.667--0.879) & 0.749 (0.653--0.840) & 0.575 (0.486--0.664) & 0.450 (0.382--0.523) \\
Virchow & 0.788 (0.682--0.879) & 0.776 (0.688--0.859) & \underline{0.597 (0.513--0.684)} & \underline{0.476 (0.409--0.547)} \\
UNI & 0.742 (0.636--0.849) & 0.714 (0.615--0.805) & 0.515 (0.431--0.598) & 0.415 (0.339--0.492) \\
Phikon-v2 & \underline{0.803 (0.697--0.909)} & \underline{0.777 (0.686--0.860)} & 0.537 (0.462--0.616) & 0.423 (0.361--0.488) \\
\hline
\end{tabular}
\par\smallskip
\noindent\parbox{\textwidth}{\SuppTableNotes Values are point estimate (95\% CI). Confidence intervals were estimated by bootstrap resampling. Bold indicates the highest value for each level; underlining indicates the second-highest value.}
\end{table}

\begin{table}[htbp]
\centering
\caption{Level-wise Accuracy for the Discordance Group.}
\label{tab:dsmil_accuracy_discordance_group}
\SuppTableStandard
\setlength{\tabcolsep}{4pt}
\begin{tabular}{@{}lcccc@{}}
\hline
Model & Level 1 & Level 2 & Level 3 & Level 4 \\
\hline
\textbf{FOCUS} & \textbf{0.924 (0.864--0.985)} & \textbf{0.803 (0.712--0.894)} & \textbf{0.394 (0.288--0.500)} & \textbf{0.227 (0.136--0.333)} \\
H-Optimus-1 & 0.697 (0.576--0.803) & 0.591 (0.470--0.712) & 0.242 (0.152--0.349) & 0.121 (0.061--0.212) \\
H-Optimus-0 & 0.712 (0.591--0.818) & 0.591 (0.470--0.712) & 0.242 (0.152--0.349) & 0.136 (0.061--0.227) \\
GigaPath & 0.697 (0.576--0.818) & 0.591 (0.469--0.712) & 0.242 (0.152--0.349) & 0.136 (0.061--0.227) \\
CRISP & 0.682 (0.561--0.788) & 0.561 (0.439--0.682) & 0.242 (0.152--0.349) & 0.152 (0.076--0.243) \\
Virchow2 & 0.773 (0.667--0.879) & 0.652 (0.530--0.773) & 0.349 (0.242--0.470) & 0.182 (0.106--0.288) \\
Virchow & 0.788 (0.682--0.879) & \underline{0.682 (0.576--0.788)} & \underline{0.364 (0.258--0.485)} & 0.182 (0.091--0.273) \\
UNI & 0.742 (0.636--0.849) & 0.606 (0.485--0.727) & 0.273 (0.182--0.379) & \underline{0.197 (0.106--0.288)} \\
Phikon-v2 & \underline{0.803 (0.697--0.909)} & \underline{0.682 (0.561--0.788)} & 0.258 (0.152--0.364) & 0.076 (0.015--0.152) \\
\hline
\end{tabular}
\par\smallskip
\noindent\parbox{\textwidth}{\SuppTableNotes Values are point estimate (95\% CI). Confidence intervals were estimated by bootstrap resampling. Bold indicates the highest value for each level; underlining indicates the second-highest value.}
\end{table}


\renewcommand{\arraystretch}{1.16}
\begin{table}[p]
\centering
\caption{Macro AUC before and after incorporating clinical information for the Tiantan test-set diagnostic tasks.}
\label{tab:clinical_information_task_delta}
\SuppTableCompact
\setlength{\tabcolsep}{2.5pt}
\begin{tabular}{@{}l>{\raggedright\arraybackslash}p{0.31\textwidth}rrr@{}}
\hline
Task & Without clinical & With clinical & $\Delta$ \\
\hline

T9 & 0.932 (0.912--0.950) & 0.960 (0.956--0.969) & +0.028 \\

T5& 0.898 (0.861--0.936) & 0.913 (0.882--0.947) & +0.016 \\

T8 & 0.956 (0.931--0.977) & 0.971 (0.953--0.985) & +0.015 \\

T7 & 0.977 (0.944--0.997) & 0.984 (0.954--0.999) & +0.008 \\

T1 & 0.964 (0.941--0.983) & 0.972 (0.954--0.987) & +0.008 \\

T10 & 0.988 (0.958--1.000) & 0.990 (0.965--1.000) & +0.002 \\

T3 & 0.992 (0.985--0.997) & 0.994 (0.987--0.998) & +0.002 \\

T4 & 0.987 (0.977--0.995) & 0.983 (0.967--0.995) & -0.004 \\

T6 & 0.992 (0.976--1.000) & 0.988 (0.965--1.000) & -0.004 \\

T2 & 0.995 (0.990--0.999) & 0.991 (0.979--0.999) & -0.005 \\

\hline
\end{tabular}

\par\smallskip
\noindent\parbox{\textwidth}{
\SuppTableNotes
Values are macro AUCs expressed as point estimates (bootstrap 95\% CIs). 
$\Delta$ is the absolute change after incorporating clinical information. 
Task-specific CIs were estimated using 2,000 stratified bootstrap resamples. 
GNT, glioneuronal tumour; PitNET, pituitary neuroendocrine tumour; 
SFT, solitary fibrous tumour;
PA, pilocytic astrocytoma; PXA, pleomorphic xanthoastrocytoma; CNS, central nervous system.
}
\end{table}

\begin{table}[p]
\centering
\caption{Hierarchical and leaf-level classification metrics before and after incorporating clinical information on the Tiantan test set.}
\label{tab:clinical_information_metric_delta}
\SuppTableCompact
\setlength{\tabcolsep}{3pt}
\begin{tabular}{@{}lrrr@{}}
\hline
Metric & Without clinical & With clinical & $\Delta$ \\
\hline
Hierarchical F1 score & 0.865 (0.845--0.882) & 0.888 (0.871--0.905) & +0.024 \\
Depth-weighted hierarchical F1 score & 0.825 (0.801--0.847) & 0.854 (0.833--0.876) & +0.029 \\
Leaf accuracy & 0.702 (0.667--0.736) & 0.755 (0.720--0.791) & +0.054 \\
Balanced accuracy & 0.578 (0.528--0.627) & 0.628 (0.580--0.678) & +0.050 \\
Leaf macro F1 score & 0.582 (0.530--0.623) & 0.633 (0.578--0.676) & +0.051 \\
Leaf weighted F1 score & 0.691 (0.655--0.728) & 0.746 (0.710--0.781) & +0.055 \\
\hline
Level 1 leaf accuracy & 0.933 (0.911--0.951) & 0.947 (0.927--0.963) & +0.014 \\
Level 2 leaf accuracy & 0.905 (0.878--0.927) & 0.925 (0.904--0.945) & +0.020 \\
Level 3 leaf accuracy & 0.830 (0.800--0.856) & 0.858 (0.829--0.884) & +0.028 \\
Level 4 leaf accuracy & 0.702 (0.667--0.736) & 0.755 (0.720--0.791) & +0.054 \\
\hline
\end{tabular}
\par\smallskip
\noindent\parbox{\textwidth}{\SuppTableNotes Values are point estimate (bootstrap 95\% CI). $\Delta$ is the absolute change after incorporating clinical information. CIs are from the 2,000 bootstrap resamples reported in the source summary files. Hierarchical F1 and depth-weighted hierarchical F1 follow the hierarchical label structure; level-wise leaf accuracy is reported from the coarsest (Level 1) to the finest (Level 4) diagnostic level.}
\end{table}


\endgroup
\begingroup
\renewcommand{\arraystretch}{1.14}
\begin{table}[p]
\centering
\caption{Paired hierarchical performance in the Randomly Selected Group.}
\label{tab:reader-study-performance-g1}
\SuppTableCompact
\setlength{\tabcolsep}{3.2pt}
\begin{tabular}{@{}lrrrrrrrr@{}}
\toprule
Reader & H-F1 & H-F1 + AI & $\Delta$H-F1 & $P$ & DH-F1 & DH-F1 + AI & $\Delta$DH-F1 & $P$ \\
\midrule
Junior 1 & 0.735 & 0.796 & +0.061 & $<0.001$ & 0.645 & 0.716 & +0.070 & 0.003 \\
Junior 2 & 0.670 & 0.817 & +0.147 & $<0.001$ & 0.556 & 0.724 & +0.168 & $<0.001$ \\
Junior 3 & 0.682 & 0.746 & +0.064 & $<0.001$ & 0.580 & 0.649 & +0.069 & $<0.001$ \\
Senior 1 & 0.785 & 0.803 & +0.017 & 0.031 & 0.690 & 0.712 & +0.022 & 0.031 \\
Senior 2 & 0.776 & 0.862 & +0.086 & $<0.001$ & 0.680 & 0.799 & +0.119 & $<0.001$ \\
Senior 3 & 0.876 & 0.876 & +0.000 & 1.000 & 0.810 & 0.810 & +0.000 & 1.000 \\
\midrule
\textbf{FOCUS} & \textbf{0.923} & -- & -- & -- & \textbf{0.897} & -- & -- & -- \\
\bottomrule
\end{tabular}
\par\smallskip
\noindent\parbox{\textwidth}{\SuppTableNotes $\Delta$ is AI-assisted minus unaided performance. $P$ is from a two-sided paired hierarchy-metric randomization test (10,000 permutations); P values are adjusted by Benjamini--Hochberg correction.}
\end{table}
\begin{table}[p]
\centering
\caption{Paired hierarchical performance in the Uncertainty Group.}
\label{tab:reader-study-performance-g2}
\SuppTableCompact
\setlength{\tabcolsep}{3.2pt}
\begin{tabular}{@{}lrrrrrrrr@{}}
\toprule
Reader & H-F1 & H-F1 + AI & $\Delta$H-F1 & $P$ & DH-F1 & DH-F1 + AI & $\Delta$DH-F1 & $P$ \\
\midrule
Junior 1 & 0.711 & 0.784 & +0.073 & 0.001 & 0.593 & 0.680 & +0.087 & 0.002 \\
Junior 2 & 0.654 & 0.707 & +0.053 & 0.039 & 0.529 & 0.593 & +0.064 & 0.024 \\
Junior 3 & 0.767 & 0.783 & +0.016 & 0.299 & 0.677 & 0.698 & +0.021 & 0.172 \\
Senior 1 & 0.719 & 0.767 & +0.048 & 0.004 & 0.629 & 0.686 & +0.057 & 0.003 \\
Senior 2 & 0.702 & 0.807 & +0.105 & $<0.001$ & 0.587 & 0.723 & +0.136 & $<0.001$ \\
Senior 3 & 0.817 & 0.834 & +0.017 & 0.064 & 0.754 & 0.774 & +0.020 & 0.064 \\
\midrule
\textbf{FOCUS} & \textbf{0.813} & -- & -- & -- & \textbf{0.761} & -- & -- & -- \\
\bottomrule
\end{tabular}
\par\smallskip
\noindent\parbox{\textwidth}{\SuppTableNotes $\Delta$ is AI-assisted minus unaided performance. $P$ is from a two-sided paired hierarchy-metric randomization test (10,000 permutations); P values are adjusted by Benjamini--Hochberg correction.}
\end{table}
\begin{table}[p]
\centering
\caption{Paired hierarchical performance in the Discordance Group.}
\label{tab:reader-study-performance-g3}
\SuppTableCompact
\setlength{\tabcolsep}{3.2pt}
\begin{tabular}{@{}lrrrrrrrr@{}}
\toprule
Reader & H-F1 & H-F1 + AI & $\Delta$H-F1 & $P$ & DH-F1 & DH-F1 + AI & $\Delta$DH-F1 & $P$ \\
\midrule
Junior 1 & 0.670 & 0.682 & +0.012 & 0.408 & \textbf{0.553} & 0.566 & +0.013 & 0.433 \\
Junior 2 & 0.600 & 0.591 & -0.009 & 0.932 & 0.463 & 0.451 & -0.012 & 0.931 \\
Junior 3 & 0.419 & 0.476 & +0.057 & 0.010 & 0.326 & 0.387 & +0.061 & 0.010 \\
Senior 1 & 0.496 & 0.508 & +0.012 & 0.125 & 0.393 & 0.404 & +0.011 & 0.125 \\
Senior 2 & 0.466 & 0.568 & +0.102 & 0.013 & 0.360 & 0.468 & +0.108 & 0.013 \\
Senior 3 & 0.529 & 0.550 & +0.021 & 0.126 & 0.462 & 0.477 & +0.015 & 0.126 \\
\midrule
\textbf{FOCUS} & \textbf{0.637} & -- & -- & -- & 0.503 & -- & -- & -- \\
\bottomrule
\end{tabular}
\par\smallskip
\noindent\parbox{\textwidth}{\SuppTableNotes $\Delta$ is AI-assisted minus unaided performance. $P$ is from a two-sided paired hierarchy-metric randomization test (10,000 permutations); P values are adjusted by Benjamini--Hochberg correction.}
\end{table}
\endgroup

\begingroup
\renewcommand{\arraystretch}{1.14}
\begin{table}[p]
\centering
\caption{Paired maximum correct depth in the Randomly Selected Group.}
\label{tab:reader-study-depth-g1}
\SuppTableCompact
\setlength{\tabcolsep}{3.2pt}
\begin{tabular}{@{}lrrrrrr@{}}
\toprule
Reader & $n$ & Unaided & AI-assisted & Mean $\Delta$ & SD & 95\% CI \\
\midrule
Junior 1 & 100 & 2.14 & 2.36 & +0.22 & 0.61 & [0.10, 0.34] \\
Junior 2 & 100 & 2.08 & 2.54 & +0.46 & 0.86 & [0.29, 0.63] \\
Junior 3 & 100 & 2.20 & 2.35 & +0.15 & 0.44 & [0.06, 0.24] \\
Senior 1 & 100 & 2.36 & 2.42 & +0.06 & 0.34 & [-0.01, 0.13] \\
Senior 2 & 100 & 2.27 & 2.71 & +0.44 & 0.69 & [0.31, 0.57] \\
Senior 3 & 100 & 2.61 & 2.61 & +0.00 & 0.00 & [0.00, 0.00] \\
\bottomrule
\end{tabular}
\par\smallskip
\noindent\parbox{\textwidth}{\SuppTableNotes All paired cases are shown. $\Delta$ is AI-assisted minus unaided maximum correct depth; 95\% confidence intervals use the normal approximation to the paired-case mean. SD, standard deviation.}
\end{table}
\begin{table}[p]
\centering
\caption{Paired maximum correct depth in the Uncertainty Group.}
\label{tab:reader-study-depth-g2}
\SuppTableCompact
\setlength{\tabcolsep}{3.2pt}
\begin{tabular}{@{}lrrrrrr@{}}
\toprule
Reader & $n$ & Unaided & AI-assisted & Mean $\Delta$ & SD & 95\% CI \\
\midrule
Junior 1 & 77 & 2.16 & 2.38 & +0.22 & 0.60 & [0.09, 0.35] \\
Junior 2 & 77 & 2.16 & 2.38 & +0.22 & 0.91 & [0.02, 0.42] \\
Junior 3 & 77 & 2.38 & 2.44 & +0.06 & 0.34 & [-0.01, 0.14] \\
Senior 1 & 77 & 2.19 & 2.39 & +0.19 & 0.56 & [0.07, 0.32] \\
Senior 2 & 77 & 2.06 & 2.56 & +0.49 & 0.75 & [0.33, 0.66] \\
Senior 3 & 77 & 2.52 & 2.57 & +0.05 & 0.22 & [0.00, 0.10] \\
\bottomrule
\end{tabular}
\par\smallskip
\noindent\parbox{\textwidth}{\SuppTableNotes All paired cases are shown. $\Delta$ is AI-assisted minus unaided maximum correct depth; 95\% confidence intervals use the normal approximation to the paired-case mean. SD, standard deviation.}
\end{table}
\begin{table}[p]
\centering
\caption{Paired maximum correct depth in the Discordance Group.}
\label{tab:reader-study-depth-g3}
\SuppTableCompact
\setlength{\tabcolsep}{3.2pt}
\begin{tabular}{@{}lrrrrrr@{}}
\toprule
Reader & $n$ & Unaided & AI-assisted & Mean $\Delta$ & SD & 95\% CI \\
\midrule
Junior 1 & 66 & 2.02 & 2.06 & +0.05 & 0.27 & [-0.02, 0.11] \\
Junior 2 & 66 & 1.86 & 2.00 & +0.14 & 0.96 & [-0.09, 0.37] \\
Junior 3 & 66 & 1.14 & 1.29 & +0.15 & 0.44 & [0.05, 0.26] \\
Senior 1 & 66 & 1.44 & 1.47 & +0.03 & 0.25 & [-0.03, 0.09] \\
Senior 2 & 66 & 1.24 & 1.62 & +0.38 & 1.03 & [0.13, 0.63] \\
Senior 3 & 66 & 1.36 & 1.41 & +0.05 & 0.21 & [-0.01, 0.10] \\
\bottomrule
\end{tabular}
\par\smallskip
\noindent\parbox{\textwidth}{\SuppTableNotes All paired cases are shown. $\Delta$ is AI-assisted minus unaided maximum correct depth; 95\% confidence intervals use the normal approximation to the paired-case mean. SD, standard deviation.}
\end{table}
\endgroup

\begingroup
\renewcommand{\arraystretch}{1.14}
\begin{table}[p]
\centering
\caption{Level-wise leaf accuracy in the Randomly Selected Group.}
\label{tab:reader-study-level-accuracy-g1}
\SuppTableCompact
\setlength{\tabcolsep}{3.2pt}
\begin{tabular}{@{}lrrrr@{}}
\toprule
Reader & Level 1 & Level 2 & Level 3 & Level 4 \\
\midrule
Junior 1 & 0.899/0.899 (+0.000) & 0.747/0.848 (+0.101) & 0.545/0.667 (+0.121) & 0.232/0.313 (+0.081) \\
Junior 2 & 0.876/0.980 (+0.104) & 0.784/0.930 (+0.146) & 0.485/0.760 (+0.275) & 0.134/0.490 (+0.356) \\
Junior 3 & 0.879/0.920 (+0.041) & 0.758/0.830 (+0.072) & 0.566/0.680 (+0.114) & 0.222/0.340 (+0.118) \\
Senior 1 & 0.940/0.940 (+0.000) & 0.860/0.890 (+0.030) & 0.690/0.720 (+0.030) & 0.370/0.400 (+0.030) \\
Senior 2 & 0.949/0.980 (+0.031) & 0.838/0.940 (+0.102) & 0.667/0.790 (+0.123) & 0.273/0.610 (+0.337) \\
Senior 3 & 0.980/0.980 (+0.000) & 0.960/0.960 (+0.000) & 0.800/0.800 (+0.000) & 0.550/0.550 (+0.000) \\
\midrule
\textbf{Junior mean} & 0.885/0.933 (+0.048) & 0.763/0.869 (+0.107) & 0.532/0.702 (+0.170) & 0.196/0.381 (+0.185) \\
\textbf{Senior mean} & 0.956/0.967 (+0.010) & 0.886/0.930 (+0.044) & 0.719/0.770 (+0.051) & 0.398/0.520 (+0.122) \\
\bottomrule
\end{tabular}
\par\smallskip
\noindent\parbox{\textwidth}{\SuppTableNotes Values are unaided/AI-assisted accuracy followed by the AI-assisted minus unaided change in parentheses. Level 1: primary site; Level 2: lineage; Level 3: tumour family; Level 4: tumour subtype. Group means are arithmetic means of the three readers.}
\end{table}
\begin{table}[p]
\centering
\caption{Level-wise leaf accuracy in the Uncertainty Group.}
\label{tab:reader-study-level-accuracy-g2}
\SuppTableCompact
\setlength{\tabcolsep}{3.2pt}
\begin{tabular}{@{}lrrrr@{}}
\toprule
Reader & Level 1 & Level 2 & Level 3 & Level 4 \\
\midrule
Junior 1 & 0.948/0.961 (+0.013) & 0.792/0.883 (+0.091) & 0.416/0.558 (+0.143) & 0.143/0.273 (+0.130) \\
Junior 2 & 0.932/0.922 (-0.009) & 0.767/0.831 (+0.064) & 0.521/0.597 (+0.077) & 0.205/0.312 (+0.106) \\
Junior 3 & 0.935/0.935 (+0.000) & 0.831/0.844 (+0.013) & 0.597/0.649 (+0.052) & 0.312/0.351 (+0.039) \\
Senior 1 & 0.921/0.921 (+0.000) & 0.684/0.789 (+0.105) & 0.553/0.645 (+0.092) & 0.316/0.382 (+0.066) \\
Senior 2 & 0.933/0.947 (+0.014) & 0.747/0.895 (+0.148) & 0.467/0.724 (+0.257) & 0.173/0.421 (+0.248) \\
Senior 3 & 0.921/0.921 (+0.000) & 0.842/0.868 (+0.026) & 0.711/0.737 (+0.026) & 0.434/0.461 (+0.026) \\
\midrule
\textbf{Junior mean} & 0.938/0.939 (+0.001) & 0.797/0.853 (+0.056) & 0.511/0.602 (+0.091) & 0.220/0.312 (+0.092) \\
\textbf{Senior mean} & 0.925/0.930 (+0.005) & 0.758/0.851 (+0.093) & 0.577/0.702 (+0.125) & 0.308/0.421 (+0.113) \\
\bottomrule
\end{tabular}
\par\smallskip
\noindent\parbox{\textwidth}{\SuppTableNotes Values are unaided/AI-assisted accuracy followed by the AI-assisted minus unaided change in parentheses. Level 1: primary site; Level 2: lineage; Level 3: tumour family; Level 4: tumour subtype. Group means are arithmetic means of the three readers.}
\end{table}
\begin{table}[p]
\centering
\caption{Level-wise leaf accuracy in the Discordance Group.}
\label{tab:reader-study-level-accuracy-g3}
\SuppTableCompact
\setlength{\tabcolsep}{3.2pt}
\begin{tabular}{@{}lrrrr@{}}
\toprule
Reader & Level 1 & Level 2 & Level 3 & Level 4 \\
\midrule
Junior 1 & 0.909/0.909 (+0.000) & 0.773/0.788 (+0.015) & 0.333/0.364 (+0.030) & 0.121/0.167 (+0.045) \\
Junior 2 & 0.898/0.894 (-0.004) & 0.780/0.758 (-0.022) & 0.356/0.379 (+0.023) & 0.102/0.136 (+0.035) \\
Junior 3 & 0.591/0.621 (+0.030) & 0.439/0.530 (+0.091) & 0.182/0.273 (+0.091) & 0.167/0.227 (+0.061) \\
Senior 1 & 0.677/0.692 (+0.015) & 0.554/0.569 (+0.015) & 0.215/0.215 (+0.000) & 0.092/0.092 (+0.000) \\
Senior 2 & 0.662/0.742 (+0.081) & 0.477/0.591 (+0.114) & 0.123/0.288 (+0.165) & 0.046/0.121 (+0.075) \\
Senior 3 & 0.636/0.667 (+0.030) & 0.470/0.515 (+0.045) & 0.333/0.364 (+0.030) & 0.288/0.318 (+0.030) \\
\midrule
\textbf{Junior mean} & 0.799/0.808 (+0.009) & 0.664/0.692 (+0.028) & 0.290/0.338 (+0.048) & 0.130/0.177 (+0.047) \\
\textbf{Senior mean} & 0.658/0.700 (+0.042) & 0.500/0.558 (+0.058) & 0.224/0.289 (+0.065) & 0.142/0.177 (+0.035) \\
\bottomrule
\end{tabular}
\par\smallskip
\noindent\parbox{\textwidth}{\SuppTableNotes Values are unaided/AI-assisted accuracy followed by the AI-assisted minus unaided change in parentheses. Level 1: primary site; Level 2: lineage; Level 3: tumour family; Level 4: tumour subtype. Group means are arithmetic means of the three readers.}
\end{table}
\endgroup


\begin{table}[p]
\centering
\caption{Paired reader performance in the lymphoma differentiation cohort.}
\label{tab:reader-study-lymphoma}
\SuppTableCompact
\setlength{\tabcolsep}{3.2pt}
\begin{tabular}{@{}lrrrrrrrr@{}}
\toprule
Reader & Accuracy & Accuracy + AI & $\Delta$Accuracy & $P$ & Lymphoma recall & Recall + AI & $\Delta$Recall & $P$ \\
\midrule
Junior 1 & 0.735 & 0.843 & +0.108 & 0.090 & 0.753 & 0.988 & +0.235 & $<0.001$ \\
Junior 2 & 0.660 & 0.718 & +0.058 & 0.164 & 0.780 & 0.866 & +0.085 & 0.062 \\
Junior 3 & 0.843 & 0.941 & +0.098 & 0.062 & 0.914 & 0.988 & +0.074 & 0.062 \\
\textbf{Junior mean} & \textbf{0.746} & \textbf{0.834} & \textbf{+0.088} & \textbf{$<0.001$} & \textbf{0.816} & \textbf{0.947} & \textbf{+0.131} & \textbf{$<0.001$} \\
\addlinespace[2pt]
Senior 1 & 0.863 & 0.892 & +0.029 & 0.333 & 0.864 & 0.889 & +0.025 & 0.500 \\
Senior 2 & 0.725 & 0.784 & +0.059 & 0.338 & 0.753 & 0.877 & +0.123 & 0.062 \\
Senior 3 & 0.902 & 0.941 & +0.039 & 0.338 & 0.901 & 0.975 & +0.074 & 0.062 \\
\textbf{Senior mean} & \textbf{0.830} & \textbf{0.873} & \textbf{+0.042} & \textbf{0.065} & \textbf{0.840} & \textbf{0.914} & \textbf{+0.074} & \textbf{0.002} \\
\midrule
\textbf{FOCUS} & \textbf{0.902} & -- & -- & -- & \textbf{0.963} & -- & -- & -- \\
\bottomrule
\end{tabular}
\par\smallskip
\noindent\parbox{\textwidth}{\SuppTableNotes $\Delta$ is AI-assisted minus unaided performance on the same scale, calculated before rounding. Junior and senior means give equal weight to each of three readers. Each $P$ is from an exact two-sided patient-clustered paired randomization test. Reader $P$ values are Benjamini--Hochberg adjusted across the 12 reader-by-metric tests; group-mean $P$ values are separately adjusted across the four group-by-metric tests.}
\end{table}


\begin{table}[t]
\centering
\caption{Inter-reader agreement before and after AI assistance.}
\label{tab:reader-agreement}
\small
\setlength{\tabcolsep}{4pt}
\begin{tabular}{llrccc}
\toprule
Group & Readers & $N$ & $\kappa_{\rm pre}$ & $\kappa_{\rm AI}$ & $\Delta\kappa$ (95\% CI) \\
\midrule
Randomly Selected Group & All & 100 & 0.256 & 0.346 & +0.089 (+0.042, +0.134) \\
Randomly Selected Group & Junior & 100 & 0.116 & 0.233 & +0.118 (+0.046, +0.191) \\
Randomly Selected Group & Senior & 100 & 0.449 & 0.466 & +0.017 (-0.056, +0.087) \\
\midrule
Uncertainty Group & All & 77 & 0.212 & 0.335 & +0.123 (+0.067, +0.182) \\
Uncertainty Group & Junior & 77 & 0.077 & 0.197 & +0.121 (+0.057, +0.185) \\
Uncertainty Group & Senior & 77 & 0.328 & 0.456 & +0.128 (+0.044, +0.224) \\
\midrule
Discordance Group & All & 66 & 0.155 & 0.159 & +0.004 (-0.037, +0.042) \\
Discordance Group & Junior & 66 & -0.050 & -0.023 & +0.027 (-0.033, +0.083) \\
Discordance Group & Senior & 66 & 0.314 & 0.254 & -0.059 (-0.147, +0.021) \\
\midrule
Pooled & All & 243 & 0.225 & 0.299 & +0.075 (+0.041, +0.105) \\
Pooled & Junior & 243 & 0.075 & 0.165 & +0.090 (+0.045, +0.130) \\
Pooled & Senior & 243 & 0.383 & 0.417 & +0.034 (-0.015, +0.081) \\
\bottomrule
\end{tabular}
\begin{flushleft}
\footnotesize All, Junior and Senior include six, three and three readers, respectively. $N$ is the number of slides; the pooled analysis contains 243 instances. Percentile confidence intervals for $\Delta\kappa$ were obtained from 2,000 paired bootstrap resamples clustered by slide filename.
\end{flushleft}
\end{table}

\begin{table}[t]
\centering
\caption{Inter-reader agreement for lymphoma diagnosis before and after AI assistance.}
\label{tab:lymphoma-reader-agreement}
\small
\setlength{\tabcolsep}{5pt}
\begin{tabular}{lrccc}
\toprule
Readers & $N$ & $\kappa_{\rm pre}$ & $\kappa_{\rm AI}$ & $\Delta\kappa$ (95\% CI) \\
\midrule
All & 102 & 0.360 & 0.375 & +0.015 (-0.078, +0.109) \\
Junior & 102 & 0.175 & 0.197 & +0.022 (-0.136, +0.179) \\
Senior & 102 & 0.452 & 0.508 & +0.055 (-0.107, +0.224) \\
\bottomrule
\end{tabular}
\begin{flushleft}
\footnotesize All, Junior and Senior include six, three and three readers, respectively. Percentile confidence intervals for $\Delta\kappa$ were obtained from 2,000 paired bootstrap resamples of cases.
\end{flushleft}
\end{table}

\begingroup
\renewcommand{\arraystretch}{1.16}
\begin{table}[p]
\centering
\caption{FOCUS processing time across tested hardware configurations.}
\label{tab:focus-processing-time}
\SuppTableStandard
\setlength{\tabcolsep}{4pt}
\begin{tabular}{@{}lrrr@{}}
\toprule
Hardware configuration & Preprocessing (s) & Prediction (s) & Total (s) \\
\midrule
Intel(R) Xeon(R) Platinum 8468V CPU + NVIDIA A100 80GB GPU & 96.47 & 0.97 & 97.44 \\
AMD EPYC 7K62 CPU + NVIDIA A6000 48GB GPU & 110.71 & 0.89 & 111.60 \\
Intel(R) Xeon(R) Platinum 8157 CPU + NVIDIA RTX 3090 GPU & 138.47 & 1.00 & 139.47 \\
\bottomrule
\end{tabular}
\par\smallskip
\noindent\parbox{\textwidth}{\SuppTableNotes Values are mean per-slide processing times. The total post-preparation AI interval was estimated by adding a 60--90\,s slide-scanning interval to the measured total processing time. Conventional slide review typically required 5--10\,min, with a further 5--10\,min when difficult cases required senior consultation.}
\end{table}
\endgroup


\begingroup
\renewcommand{\arraystretch}{1.16}
\begin{table}[p]
\centering
\caption{Test-set metrics for BLCA (Benign vs. malignant).}
\label{tab:dsmil_metrics_blca_benign_malignant}
\fontsize{8.0}{9.5}\selectfont
\setlength{\tabcolsep}{4pt}
\begin{tabular}{@{}lccccc@{}}
\hline
Model & AUC & Accuracy & Sensitivity & Specificity & F1 score \\
\hline
FOCUS & \textbf{1.000 (1.000--1.000)} & \textbf{1.000 (1.000--1.000)} & \textbf{1.000 (1.000--1.000)} & \textbf{1.000 (1.000--1.000)} & \textbf{1.000 (1.000--1.000)} \\
H-Optimus-1 & \textbf{1.000 (1.000--1.000)} & \underline{0.993 (0.979--1.000)} & \textbf{1.000 (1.000--1.000)} & \underline{0.909 (0.700--1.000)} & \underline{0.996 (0.988--1.000)} \\
H-Optimus-0 & \textbf{1.000 (1.000--1.000)} & \textbf{1.000 (1.000--1.000)} & \textbf{1.000 (1.000--1.000)} & \textbf{1.000 (1.000--1.000)} & \textbf{1.000 (1.000--1.000)} \\
GigaPath & \textbf{1.000 (1.000--1.000)} & \underline{0.993 (0.979--1.000)} & \textbf{1.000 (1.000--1.000)} & \underline{0.909 (0.700--1.000)} & \underline{0.996 (0.988--1.000)} \\
Virchow2 & \textbf{1.000 (1.000--1.000)} & 0.986 (0.965--1.000) & \textbf{1.000 (1.000--1.000)} & 0.818 (0.556--1.000) & 0.992 (0.981--1.000) \\
Virchow & \textbf{1.000 (1.000--1.000)} & \underline{0.993 (0.979--1.000)} & \textbf{1.000 (1.000--1.000)} & \underline{0.909 (0.700--1.000)} & \underline{0.996 (0.988--1.000)} \\
UNI & \textbf{1.000 (1.000--1.000)} & \underline{0.993 (0.979--1.000)} & \textbf{1.000 (1.000--1.000)} & \underline{0.909 (0.700--1.000)} & \underline{0.996 (0.988--1.000)} \\
Phikon-v2 & \textbf{1.000 (1.000--1.000)} & \textbf{1.000 (1.000--1.000)} & \textbf{1.000 (1.000--1.000)} & \textbf{1.000 (1.000--1.000)} & \textbf{1.000 (1.000--1.000)} \\
\hline
\end{tabular}
\par\smallskip
\noindent\parbox{\textwidth}{\fontsize{6.5}{7.8}\selectfont Values are estimate (95\% CI). External predictions were used when available, otherwise test predictions. Binary AUC CIs use DeLong's method; all other CIs use 2,000-patient bootstrap percentile intervals. Multiclass Sensitivity, Specificity, PPV, and NPV are macro-averaged. Bold and underlining indicate the highest and second-highest values, respectively.}\end{table}
\begin{table}[p]
\centering
\caption{Test-set metrics for BRCA (Benign vs. malignant).}
\label{tab:dsmil_metrics_brca_benign_malignant}
\fontsize{8.0}{9.5}\selectfont
\setlength{\tabcolsep}{4pt}
\begin{tabular}{@{}lccccc@{}}
\hline
Model & AUC & Accuracy & Sensitivity & Specificity & F1 score \\
\hline
FOCUS & \textbf{1.000 (1.000--1.000)} & \textbf{0.998 (0.995--1.000)} & \textbf{0.998 (0.993--1.000)} & \textbf{1.000 (1.000--1.000)} & \textbf{0.999 (0.997--1.000)} \\
H-Optimus-1 & \textbf{1.000 (1.000--1.000)} & \underline{0.997 (0.992--1.000)} & \underline{0.996 (0.989--1.000)} & \textbf{1.000 (1.000--1.000)} & \underline{0.998 (0.994--1.000)} \\
H-Optimus-0 & \textbf{1.000 (1.000--1.000)} & \underline{0.997 (0.992--1.000)} & \underline{0.996 (0.989--1.000)} & \textbf{1.000 (1.000--1.000)} & \underline{0.998 (0.994--1.000)} \\
GigaPath & \underline{1.000 (1.000--1.000)} & 0.995 (0.988--1.000) & 0.994 (0.985--1.000) & \textbf{1.000 (1.000--1.000)} & 0.997 (0.993--1.000) \\
Virchow2 & \textbf{1.000 (1.000--1.000)} & \underline{0.997 (0.992--1.000)} & \underline{0.996 (0.989--1.000)} & \textbf{1.000 (1.000--1.000)} & \underline{0.998 (0.994--1.000)} \\
Virchow & \textbf{1.000 (1.000--1.000)} & \underline{0.997 (0.992--1.000)} & \underline{0.996 (0.989--1.000)} & \textbf{1.000 (1.000--1.000)} & \underline{0.998 (0.994--1.000)} \\
UNI & \textbf{1.000 (1.000--1.000)} & \textbf{0.998 (0.995--1.000)} & \textbf{0.998 (0.993--1.000)} & \textbf{1.000 (1.000--1.000)} & \textbf{0.999 (0.997--1.000)} \\
Phikon-v2 & \textbf{1.000 (1.000--1.000)} & \underline{0.997 (0.992--1.000)} & \underline{0.996 (0.989--1.000)} & \textbf{1.000 (1.000--1.000)} & \underline{0.998 (0.994--1.000)} \\
\hline
\end{tabular}
\par\smallskip
\noindent\parbox{\textwidth}{\fontsize{6.5}{7.8}\selectfont Values are estimate (95\% CI). External predictions were used when available, otherwise test predictions. Binary AUC CIs use DeLong's method; all other CIs use 2,000-patient bootstrap percentile intervals. Multiclass Sensitivity, Specificity, PPV, and NPV are macro-averaged. Bold and underlining indicate the highest and second-highest values, respectively.}\end{table}
\begin{table}[p]
\centering
\caption{External-set metrics for CESC (Tumour grading).}
\label{tab:dsmil_metrics_cesc_tumor_grading}
\fontsize{8.0}{9.5}\selectfont
\setlength{\tabcolsep}{4pt}
\begin{tabular}{@{}lccccc@{}}
\hline
Model & AUC & Accuracy & Sensitivity & Specificity & F1 score \\
\hline
FOCUS & 0.796 (0.736--0.852) & \underline{0.692 (0.625--0.755)} & \underline{0.482 (0.441--0.522)} & \underline{0.801 (0.763--0.838)} & \underline{0.670 (0.598--0.737)} \\
H-Optimus-1 & 0.760 (0.699--0.820) & 0.611 (0.543--0.678) & 0.422 (0.385--0.458) & 0.747 (0.713--0.781) & 0.565 (0.488--0.639) \\
H-Optimus-0 & 0.721 (0.656--0.783) & 0.591 (0.524--0.659) & 0.408 (0.370--0.448) & 0.735 (0.699--0.770) & 0.547 (0.468--0.622) \\
GigaPath & 0.773 (0.708--0.829) & 0.620 (0.553--0.683) & 0.428 (0.392--0.462) & 0.752 (0.719--0.783) & 0.568 (0.487--0.642) \\
Virchow2 & \underline{0.801 (0.745--0.854)} & 0.615 (0.548--0.683) & 0.424 (0.391--0.460) & 0.751 (0.720--0.783) & 0.563 (0.485--0.643) \\
Virchow & \textbf{0.805 (0.751--0.857)} & \textbf{0.707 (0.639--0.764)} & \textbf{0.493 (0.451--0.535)} & \textbf{0.812 (0.773--0.848)} & \textbf{0.688 (0.619--0.750)} \\
UNI & 0.767 (0.709--0.824) & 0.639 (0.572--0.702) & 0.443 (0.404--0.480) & 0.765 (0.730--0.800) & 0.602 (0.523--0.674) \\
Phikon-v2 & 0.709 (0.646--0.769) & 0.601 (0.534--0.668) & 0.415 (0.379--0.450) & 0.740 (0.708--0.773) & 0.549 (0.471--0.623) \\
\hline
\end{tabular}
\par\smallskip
\noindent\parbox{\textwidth}{\fontsize{6.5}{7.8}\selectfont Values are estimate (95\% CI). External predictions were used when available, otherwise test predictions. Binary AUC CIs use DeLong's method; all other CIs use 2,000-patient bootstrap percentile intervals. Multiclass Sensitivity, Specificity, PPV, and NPV are macro-averaged. Bold and underlining indicate the highest and second-highest values, respectively.}\end{table}
\begin{table}[p]
\centering
\caption{External-set metrics for COADREAD (Benign vs. malignant).}
\label{tab:dsmil_metrics_coadread_benign_malignant}
\fontsize{8.0}{9.5}\selectfont
\setlength{\tabcolsep}{4pt}
\begin{tabular}{@{}lccccc@{}}
\hline
Model & AUC & Accuracy & Sensitivity & Specificity & F1 score \\
\hline
FOCUS & \underline{0.969 (0.945--0.993)} & \textbf{0.848 (0.794--0.897)} & 0.977 (0.949--1.000) & \textbf{0.333 (0.174--0.486)} & \underline{0.912 (0.875--0.943)} \\
H-Optimus-1 & 0.962 (0.935--0.989) & \textbf{0.848 (0.794--0.897)} & 0.977 (0.949--1.000) & \textbf{0.333 (0.174--0.486)} & \underline{0.912 (0.875--0.943)} \\
H-Optimus-0 & 0.959 (0.930--0.987) & \underline{0.842 (0.782--0.897)} & \underline{0.985 (0.962--1.000)} & 0.273 (0.132--0.429) & 0.909 (0.871--0.942) \\
GigaPath & 0.947 (0.914--0.979) & 0.836 (0.782--0.891) & 0.970 (0.937--0.993) & \underline{0.303 (0.152--0.464)} & 0.905 (0.867--0.938) \\
Virchow2 & \textbf{0.978 (0.959--0.998)} & \textbf{0.848 (0.794--0.903)} & \underline{0.985 (0.962--1.000)} & \underline{0.303 (0.152--0.464)} & \textbf{0.912 (0.877--0.944)} \\
Virchow & 0.953 (0.924--0.983) & 0.830 (0.770--0.885) & 0.977 (0.949--1.000) & 0.242 (0.107--0.393) & 0.902 (0.863--0.936) \\
UNI & 0.965 (0.939--0.992) & \underline{0.842 (0.788--0.891)} & 0.970 (0.937--0.993) & \textbf{0.333 (0.174--0.486)} & 0.908 (0.871--0.940) \\
Phikon-v2 & 0.965 (0.938--0.991) & 0.824 (0.764--0.879) & \textbf{0.992 (0.977--1.000)} & 0.152 (0.036--0.289) & 0.900 (0.862--0.934) \\
\hline
\end{tabular}
\par\smallskip
\noindent\parbox{\textwidth}{\fontsize{6.5}{7.8}\selectfont Values are estimate (95\% CI). External predictions were used when available, otherwise test predictions. Binary AUC CIs use DeLong's method; all other CIs use 2,000-patient bootstrap percentile intervals. Multiclass Sensitivity, Specificity, PPV, and NPV are macro-averaged. Bold and underlining indicate the highest and second-highest values, respectively.}\end{table}
\begin{table}[p]
\centering
\caption{External-set metrics for HNSC (Benign vs. malignant).}
\label{tab:dsmil_metrics_hnsc_benign_malignant}
\fontsize{8.0}{9.5}\selectfont
\setlength{\tabcolsep}{4pt}
\begin{tabular}{@{}lccccc@{}}
\hline
Model & AUC & Accuracy & Sensitivity & Specificity & F1 score \\
\hline
FOCUS & 0.978 (0.952--1.000) & 0.962 (0.939--0.985) & \underline{0.994 (0.981--1.000)} & 0.902 (0.840--0.957) & 0.971 (0.953--0.988) \\
H-Optimus-1 & \textbf{0.983 (0.960--1.000)} & \textbf{0.985 (0.970--0.996)} & \underline{0.994 (0.981--1.000)} & \textbf{0.967 (0.928--1.000)} & \textbf{0.988 (0.975--0.997)} \\
H-Optimus-0 & \underline{0.980 (0.955--1.000)} & 0.962 (0.939--0.985) & 0.977 (0.952--0.994) & 0.935 (0.884--0.980) & 0.971 (0.952--0.988) \\
GigaPath & 0.977 (0.948--1.000) & 0.970 (0.947--0.989) & 0.982 (0.960--1.000) & 0.946 (0.894--0.989) & 0.977 (0.959--0.991) \\
Virchow2 & 0.977 (0.948--1.000) & \underline{0.977 (0.958--0.992)} & 0.988 (0.970--1.000) & \underline{0.957 (0.911--0.990)} & \underline{0.983 (0.968--0.995)} \\
Virchow & 0.973 (0.944--1.000) & 0.951 (0.924--0.977) & 0.988 (0.970--1.000) & 0.880 (0.814--0.942) & 0.963 (0.942--0.982) \\
UNI & 0.954 (0.923--0.984) & 0.821 (0.776--0.867) & 0.965 (0.935--0.989) & 0.554 (0.452--0.655) & 0.875 (0.840--0.909) \\
Phikon-v2 & 0.976 (0.954--0.998) & 0.654 (0.593--0.711) & \textbf{1.000 (1.000--1.000)} & 0.011 (0.000--0.035) & 0.790 (0.745--0.829) \\
\hline
\end{tabular}
\par\smallskip
\noindent\parbox{\textwidth}{\fontsize{6.5}{7.8}\selectfont Values are estimate (95\% CI). External predictions were used when available, otherwise test predictions. Binary AUC CIs use DeLong's method; all other CIs use 2,000-patient bootstrap percentile intervals. Multiclass Sensitivity, Specificity, PPV, and NPV are macro-averaged. Bold and underlining indicate the highest and second-highest values, respectively.}\end{table}
\begin{table}[p]
\centering
\caption{Test-set metrics for LIHC (Tumour grading).}
\label{tab:dsmil_metrics_lihc_tumor_grading}
\fontsize{8.0}{9.5}\selectfont
\setlength{\tabcolsep}{4pt}
\begin{tabular}{@{}lccccc@{}}
\hline
Model & AUC & Accuracy & Sensitivity & Specificity & F1 score \\
\hline
FOCUS & \underline{0.634 (0.553--0.709)} & 0.462 (0.370--0.555) & 0.290 (0.229--0.358) & 0.764 (0.729--0.799) & 0.430 (0.333--0.525) \\
H-Optimus-1 & 0.629 (0.552--0.706) & 0.471 (0.378--0.555) & \underline{0.306 (0.246--0.374)} & \textbf{0.773 (0.738--0.810)} & \textbf{0.444 (0.347--0.539)} \\
H-Optimus-0 & 0.612 (0.530--0.688) & \textbf{0.504 (0.412--0.588)} & 0.284 (0.242--0.328) & \underline{0.771 (0.743--0.800)} & 0.424 (0.328--0.522) \\
GigaPath & 0.606 (0.527--0.686) & 0.445 (0.353--0.538) & 0.264 (0.214--0.321) & 0.752 (0.719--0.786) & 0.397 (0.301--0.491) \\
Virchow2 & \textbf{0.656 (0.576--0.733)} & \underline{0.496 (0.403--0.588)} & \textbf{0.306 (0.244--0.371)} & 0.771 (0.739--0.802) & \underline{0.443 (0.343--0.541)} \\
Virchow & 0.601 (0.518--0.683) & \underline{0.496 (0.403--0.580)} & 0.287 (0.241--0.340) & 0.767 (0.739--0.799) & 0.426 (0.327--0.525) \\
UNI & 0.621 (0.544--0.698) & 0.445 (0.353--0.538) & 0.262 (0.212--0.319) & 0.751 (0.720--0.784) & 0.393 (0.295--0.494) \\
Phikon-v2 & 0.568 (0.484--0.646) & 0.454 (0.361--0.538) & 0.259 (0.214--0.307) & 0.755 (0.723--0.787) & 0.391 (0.297--0.484) \\
\hline
\end{tabular}
\par\smallskip
\noindent\parbox{\textwidth}{\fontsize{6.5}{7.8}\selectfont Values are estimate (95\% CI). External predictions were used when available, otherwise test predictions. Binary AUC CIs use DeLong's method; all other CIs use 2,000-patient bootstrap percentile intervals. Multiclass Sensitivity, Specificity, PPV, and NPV are macro-averaged. Bold and underlining indicate the highest and second-highest values, respectively.}\end{table}
\begin{table}[p]
\centering
\caption{Test-set metrics for LIHC (Benign vs. malignant).}
\label{tab:dsmil_metrics_lihc_benign_malignant}
\fontsize{8.0}{9.5}\selectfont
\setlength{\tabcolsep}{4pt}
\begin{tabular}{@{}lccccc@{}}
\hline
Model & AUC & Accuracy & Sensitivity & Specificity & F1 score \\
\hline
FOCUS & \textbf{1.000 (1.000--1.000)} & \textbf{1.000 (1.000--1.000)} & \textbf{1.000 (1.000--1.000)} & \textbf{1.000 (1.000--1.000)} & \textbf{1.000 (1.000--1.000)} \\
H-Optimus-1 & \textbf{1.000 (1.000--1.000)} & \textbf{1.000 (1.000--1.000)} & \textbf{1.000 (1.000--1.000)} & \textbf{1.000 (1.000--1.000)} & \textbf{1.000 (1.000--1.000)} \\
H-Optimus-0 & \textbf{1.000 (1.000--1.000)} & \textbf{1.000 (1.000--1.000)} & \textbf{1.000 (1.000--1.000)} & \textbf{1.000 (1.000--1.000)} & \textbf{1.000 (1.000--1.000)} \\
GigaPath & \textbf{1.000 (1.000--1.000)} & \textbf{1.000 (1.000--1.000)} & \textbf{1.000 (1.000--1.000)} & \textbf{1.000 (1.000--1.000)} & \textbf{1.000 (1.000--1.000)} \\
Virchow2 & \textbf{1.000 (1.000--1.000)} & \textbf{1.000 (1.000--1.000)} & \textbf{1.000 (1.000--1.000)} & \textbf{1.000 (1.000--1.000)} & \textbf{1.000 (1.000--1.000)} \\
Virchow & \textbf{1.000 (1.000--1.000)} & \textbf{1.000 (1.000--1.000)} & \textbf{1.000 (1.000--1.000)} & \textbf{1.000 (1.000--1.000)} & \textbf{1.000 (1.000--1.000)} \\
UNI & \textbf{1.000 (1.000--1.000)} & \textbf{1.000 (1.000--1.000)} & \textbf{1.000 (1.000--1.000)} & \textbf{1.000 (1.000--1.000)} & \textbf{1.000 (1.000--1.000)} \\
Phikon-v2 & \textbf{1.000 (1.000--1.000)} & \textbf{1.000 (1.000--1.000)} & \textbf{1.000 (1.000--1.000)} & \textbf{1.000 (1.000--1.000)} & \textbf{1.000 (1.000--1.000)} \\
\hline
\end{tabular}
\par\smallskip
\noindent\parbox{\textwidth}{\fontsize{6.5}{7.8}\selectfont Values are estimate (95\% CI). External predictions were used when available, otherwise test predictions. Binary AUC CIs use DeLong's method; all other CIs use 2,000-patient bootstrap percentile intervals. Multiclass Sensitivity, Specificity, PPV, and NPV are macro-averaged. Bold and underlining indicate the highest and second-highest values, respectively.}\end{table}
\begin{table}[p]
\centering
\caption{Test-set metrics for OV (Benign vs. malignant).}
\label{tab:dsmil_metrics_ov_benign_malignant}
\fontsize{8.0}{9.5}\selectfont
\setlength{\tabcolsep}{4pt}
\begin{tabular}{@{}lccccc@{}}
\hline
Model & AUC & Accuracy & Sensitivity & Specificity & F1 score \\
\hline
FOCUS & \textbf{0.999 (0.998--1.000)} & \textbf{0.995 (0.988--1.000)} & \textbf{0.994 (0.986--1.000)} & \textbf{1.000 (1.000--1.000)} & \textbf{0.997 (0.993--1.000)} \\
H-Optimus-1 & \underline{0.999 (0.997--1.000)} & \textbf{0.995 (0.988--1.000)} & \textbf{0.994 (0.986--1.000)} & \textbf{1.000 (1.000--1.000)} & \textbf{0.997 (0.993--1.000)} \\
H-Optimus-0 & 0.999 (0.996--1.000) & \underline{0.993 (0.983--1.000)} & \textbf{0.994 (0.986--1.000)} & \underline{0.980 (0.933--1.000)} & \underline{0.996 (0.990--1.000)} \\
GigaPath & 0.998 (0.995--1.000) & 0.990 (0.981--0.998) & \underline{0.992 (0.981--1.000)} & \underline{0.980 (0.933--1.000)} & 0.994 (0.989--0.999) \\
Virchow & 0.998 (0.996--1.000) & 0.990 (0.981--0.998) & \underline{0.992 (0.981--1.000)} & \underline{0.980 (0.933--1.000)} & 0.994 (0.989--0.999) \\
UNI & 0.998 (0.994--1.000) & 0.990 (0.981--0.998) & \underline{0.992 (0.981--1.000)} & \underline{0.980 (0.933--1.000)} & 0.994 (0.989--0.999) \\
Phikon-v2 & 0.998 (0.995--1.000) & \underline{0.993 (0.983--1.000)} & \textbf{0.994 (0.986--1.000)} & \underline{0.980 (0.933--1.000)} & \underline{0.996 (0.990--1.000)} \\
\hline
\end{tabular}
\par\smallskip
\noindent\parbox{\textwidth}{\fontsize{6.5}{7.8}\selectfont Values are estimate (95\% CI). External predictions were used when available, otherwise test predictions. Binary AUC CIs use DeLong's method; all other CIs use 2,000-patient bootstrap percentile intervals. Multiclass Sensitivity, Specificity, PPV, and NPV are macro-averaged. Bold and underlining indicate the highest and second-highest values, respectively.}\end{table}
\begin{table}[p]
\centering
\caption{External-set metrics for PAAD (Benign vs. malignant).}
\label{tab:dsmil_metrics_paad_benign_malignant}
\fontsize{8.0}{9.5}\selectfont
\setlength{\tabcolsep}{4pt}
\begin{tabular}{@{}lccccc@{}}
\hline
Model & AUC & Accuracy & Sensitivity & Specificity & F1 score \\
\hline
FOCUS & \textbf{0.968 (0.949--0.987)} & 0.908 (0.880--0.936) & 0.959 (0.934--0.981) & 0.803 (0.734--0.871) & 0.934 (0.912--0.955) \\
H-Optimus-1 & 0.962 (0.941--0.983) & \underline{0.916 (0.885--0.941)} & 0.917 (0.883--0.949) & \textbf{0.913 (0.859--0.959)} & 0.937 (0.913--0.957) \\
H-Optimus-0 & 0.956 (0.932--0.979) & 0.906 (0.875--0.934) & 0.940 (0.909--0.968) & \underline{0.835 (0.767--0.898)} & 0.931 (0.908--0.953) \\
GigaPath & 0.958 (0.934--0.982) & 0.891 (0.860--0.919) & 0.985 (0.969--0.996) & 0.693 (0.610--0.772) & 0.924 (0.901--0.946) \\
Virchow2 & \underline{0.966 (0.945--0.988)} & \textbf{0.921 (0.893--0.947)} & 0.977 (0.958--0.993) & 0.803 (0.728--0.872) & \textbf{0.944 (0.923--0.963)} \\
Virchow & 0.960 (0.938--0.983) & 0.913 (0.885--0.941) & \underline{0.996 (0.988--1.000)} & 0.740 (0.661--0.816) & \underline{0.940 (0.919--0.959)} \\
UNI & 0.923 (0.894--0.952) & 0.781 (0.740--0.822) & 0.981 (0.964--0.996) & 0.362 (0.279--0.450) & 0.859 (0.828--0.886) \\
Phikon-v2 & 0.885 (0.849--0.920) & 0.684 (0.639--0.728) & \textbf{1.000 (1.000--1.000)} & 0.024 (0.000--0.052) & 0.811 (0.777--0.842) \\
\hline
\end{tabular}
\par\smallskip
\noindent\parbox{\textwidth}{\fontsize{6.5}{7.8}\selectfont Values are estimate (95\% CI). External predictions were used when available, otherwise test predictions. Binary AUC CIs use DeLong's method; all other CIs use 2,000-patient bootstrap percentile intervals. Multiclass Sensitivity, Specificity, PPV, and NPV are macro-averaged. Bold and underlining indicate the highest and second-highest values, respectively.}\end{table}
\begin{table}[p]
\centering
\caption{Test-set metrics for PRAD (Gleason grade group).}
\label{tab:dsmil_metrics_prad_gleason_grade_group}
\fontsize{8.0}{9.5}\selectfont
\setlength{\tabcolsep}{4pt}
\begin{tabular}{@{}lccccc@{}}
\hline
Model & AUC & Accuracy & Sensitivity & Specificity & F1 score \\
\hline
FOCUS & \textbf{0.774 (0.731--0.818)} & \textbf{0.505 (0.434--0.577)} & \textbf{0.393 (0.331--0.455)} & \textbf{0.867 (0.848--0.885)} & \textbf{0.476 (0.399--0.551)} \\
H-Optimus-1 & 0.744 (0.696--0.792) & \underline{0.489 (0.418--0.560)} & 0.348 (0.306--0.391) & 0.859 (0.842--0.876) & 0.415 (0.335--0.494) \\
H-Optimus-0 & 0.730 (0.681--0.778) & 0.478 (0.407--0.549) & 0.333 (0.296--0.371) & 0.857 (0.840--0.874) & 0.400 (0.320--0.477) \\
GigaPath & 0.743 (0.693--0.792) & 0.467 (0.396--0.538) & 0.338 (0.291--0.385) & 0.856 (0.838--0.873) & 0.417 (0.338--0.493) \\
Virchow2 & 0.748 (0.698--0.796) & 0.473 (0.401--0.544) & 0.353 (0.299--0.411) & 0.856 (0.838--0.874) & 0.433 (0.355--0.507) \\
Virchow & 0.741 (0.693--0.788) & 0.484 (0.412--0.549) & 0.351 (0.305--0.402) & 0.859 (0.842--0.876) & 0.428 (0.347--0.504) \\
UNI & \underline{0.752 (0.704--0.800)} & \textbf{0.505 (0.434--0.577)} & \underline{0.388 (0.331--0.446)} & \underline{0.865 (0.847--0.884)} & \underline{0.470 (0.395--0.546)} \\
Phikon-v2 & 0.689 (0.634--0.742) & 0.451 (0.379--0.522) & 0.309 (0.273--0.346) & 0.845 (0.830--0.862) & 0.356 (0.283--0.434) \\
\hline
\end{tabular}
\par\smallskip
\noindent\parbox{\textwidth}{\fontsize{6.5}{7.8}\selectfont Values are estimate (95\% CI). External predictions were used when available, otherwise test predictions. Binary AUC CIs use DeLong's method; all other CIs use 2,000-patient bootstrap percentile intervals. Multiclass Sensitivity, Specificity, PPV, and NPV are macro-averaged. Bold and underlining indicate the highest and second-highest values, respectively.}\end{table}
\begin{table}[p]
\centering
\caption{Test-set metrics for PRAD (Benign vs. malignant).}
\label{tab:dsmil_metrics_prad_benign_malignant}
\fontsize{8.0}{9.5}\selectfont
\setlength{\tabcolsep}{4pt}
\begin{tabular}{@{}lccccc@{}}
\hline
Model & AUC & Accuracy & Sensitivity & Specificity & F1 score \\
\hline
FOCUS & 0.990 (0.971--1.000) & \textbf{0.986 (0.968--1.000)} & \textbf{1.000 (1.000--1.000)} & \textbf{0.914 (0.810--1.000)} & \textbf{0.992 (0.981--1.000)} \\
H-Optimus-1 & 0.996 (0.990--1.000) & 0.968 (0.945--0.991) & \underline{0.995 (0.983--1.000)} & 0.829 (0.700--0.946) & 0.981 (0.966--0.994) \\
H-Optimus-0 & 0.992 (0.984--1.000) & 0.963 (0.935--0.986) & \underline{0.995 (0.983--1.000)} & 0.800 (0.667--0.929) & 0.978 (0.962--0.992) \\
GigaPath & 0.993 (0.982--1.000) & \underline{0.972 (0.949--0.991)} & \textbf{1.000 (1.000--1.000)} & 0.829 (0.690--0.944) & \underline{0.984 (0.969--0.995)} \\
Virchow2 & 0.978 (0.938--1.000) & \underline{0.972 (0.949--0.991)} & \underline{0.995 (0.983--1.000)} & \underline{0.857 (0.731--0.969)} & 0.984 (0.969--0.995) \\
Virchow & 0.995 (0.989--1.000) & 0.963 (0.935--0.986) & 0.989 (0.972--1.000) & 0.829 (0.692--0.943) & 0.978 (0.961--0.992) \\
UNI & \textbf{0.997 (0.994--1.000)} & \underline{0.972 (0.949--0.991)} & \textbf{1.000 (1.000--1.000)} & 0.829 (0.700--0.946) & \underline{0.984 (0.970--0.995)} \\
Phikon-v2 & \underline{0.996 (0.991--1.000)} & 0.963 (0.935--0.986) & 0.989 (0.972--1.000) & 0.829 (0.697--0.943) & 0.978 (0.961--0.992) \\
\hline
\end{tabular}
\par\smallskip
\noindent\parbox{\textwidth}{\fontsize{6.5}{7.8}\selectfont Values are estimate (95\% CI). External predictions were used when available, otherwise test predictions. Binary AUC CIs use DeLong's method; all other CIs use 2,000-patient bootstrap percentile intervals. Multiclass Sensitivity, Specificity, PPV, and NPV are macro-averaged. Bold and underlining indicate the highest and second-highest values, respectively.}\end{table}
\begin{table}[p]
\centering
\caption{Test-set metrics for STAD (Benign vs. malignant).}
\label{tab:dsmil_metrics_stad_benign_malignant}
\fontsize{8.0}{9.5}\selectfont
\setlength{\tabcolsep}{4pt}
\begin{tabular}{@{}lccccc@{}}
\hline
Model & AUC & Accuracy & Sensitivity & Specificity & F1 score \\
\hline
FOCUS & \textbf{1.000 (1.000--1.000)} & \textbf{1.000 (1.000--1.000)} & \textbf{1.000 (1.000--1.000)} & \textbf{1.000 (1.000--1.000)} & \textbf{1.000 (1.000--1.000)} \\
H-Optimus-1 & 0.999 (0.998--1.000) & 0.991 (0.978--1.000) & 0.989 (0.973--1.000) & \textbf{1.000 (1.000--1.000)} & 0.995 (0.986--1.000) \\
H-Optimus-0 & 0.999 (0.998--1.000) & 0.982 (0.965--0.996) & 0.995 (0.984--1.000) & 0.919 (0.818--1.000) & 0.989 (0.978--0.997) \\
GigaPath & 0.999 (0.998--1.000) & 0.982 (0.965--0.996) & \underline{0.995 (0.984--1.000)} & 0.919 (0.818--1.000) & 0.990 (0.978--0.997) \\
Virchow2 & \textbf{1.000 (1.000--1.000)} & \underline{0.996 (0.987--1.000)} & \textbf{1.000 (1.000--1.000)} & \underline{0.973 (0.909--1.000)} & \underline{0.997 (0.992--1.000)} \\
Virchow & 0.999 (0.998--1.000) & 0.982 (0.965--0.996) & \underline{0.995 (0.984--1.000)} & 0.919 (0.822--1.000) & 0.990 (0.978--0.997) \\
UNI & 1.000 (0.999--1.000) & 0.987 (0.969--1.000) & 0.995 (0.984--1.000) & 0.946 (0.862--1.000) & 0.992 (0.982--1.000) \\
Phikon-v2 & \underline{1.000 (0.999--1.000)} & 0.987 (0.969--1.000) & \textbf{1.000 (1.000--1.000)} & 0.919 (0.828--1.000) & 0.992 (0.982--1.000) \\
\hline
\end{tabular}
\par\smallskip
\noindent\parbox{\textwidth}{\fontsize{6.5}{7.8}\selectfont Values are estimate (95\% CI). External predictions were used when available, otherwise test predictions. Binary AUC CIs use DeLong's method; all other CIs use 2,000-patient bootstrap percentile intervals. Multiclass Sensitivity, Specificity, PPV, and NPV are macro-averaged. Bold and underlining indicate the highest and second-highest values, respectively.}\end{table}
\begin{table}[p]
\centering
\caption{Test-set metrics for THCA (Benign vs. malignant).}
\label{tab:dsmil_metrics_thca_benign_malignant}
\fontsize{8.0}{9.5}\selectfont
\setlength{\tabcolsep}{4pt}
\begin{tabular}{@{}lccccc@{}}
\hline
Model & AUC & Accuracy & Sensitivity & Specificity & F1 score \\
\hline
FOCUS & \textbf{1.000 (1.000--1.000)} & \underline{0.995 (0.984--1.000)} & \underline{0.994 (0.981--1.000)} & \textbf{1.000 (1.000--1.000)} & \underline{0.997 (0.990--1.000)} \\
H-Optimus-1 & \textbf{1.000 (1.000--1.000)} & \textbf{1.000 (1.000--1.000)} & \textbf{1.000 (1.000--1.000)} & \textbf{1.000 (1.000--1.000)} & \textbf{1.000 (1.000--1.000)} \\
H-Optimus-0 & \textbf{1.000 (1.000--1.000)} & \textbf{1.000 (1.000--1.000)} & \textbf{1.000 (1.000--1.000)} & \textbf{1.000 (1.000--1.000)} & \textbf{1.000 (1.000--1.000)} \\
GigaPath & \textbf{1.000 (1.000--1.000)} & \textbf{1.000 (1.000--1.000)} & \textbf{1.000 (1.000--1.000)} & \textbf{1.000 (1.000--1.000)} & \textbf{1.000 (1.000--1.000)} \\
Virchow2 & \textbf{1.000 (1.000--1.000)} & \textbf{1.000 (1.000--1.000)} & \textbf{1.000 (1.000--1.000)} & \textbf{1.000 (1.000--1.000)} & \textbf{1.000 (1.000--1.000)} \\
Virchow & \textbf{1.000 (1.000--1.000)} & \textbf{1.000 (1.000--1.000)} & \textbf{1.000 (1.000--1.000)} & \textbf{1.000 (1.000--1.000)} & \textbf{1.000 (1.000--1.000)} \\
UNI & \textbf{1.000 (1.000--1.000)} & \textbf{1.000 (1.000--1.000)} & \textbf{1.000 (1.000--1.000)} & \textbf{1.000 (1.000--1.000)} & \textbf{1.000 (1.000--1.000)} \\
Phikon-v2 & \textbf{1.000 (1.000--1.000)} & \textbf{1.000 (1.000--1.000)} & \textbf{1.000 (1.000--1.000)} & \textbf{1.000 (1.000--1.000)} & \textbf{1.000 (1.000--1.000)} \\
\hline
\end{tabular}
\par\smallskip
\noindent\parbox{\textwidth}{\fontsize{6.5}{7.8}\selectfont Values are estimate (95\% CI). External predictions were used when available, otherwise test predictions. Binary AUC CIs use DeLong's method; all other CIs use 2,000-patient bootstrap percentile intervals. Multiclass Sensitivity, Specificity, PPV, and NPV are macro-averaged. Bold and underlining indicate the highest and second-highest values, respectively.}\end{table}
\begin{table}[p]
\centering
\caption{Test-set metrics for UCEC/UCS (Benign vs. malignant).}
\label{tab:dsmil_metrics_ucec_ucs_benign_malignant}
\fontsize{8.0}{9.5}\selectfont
\setlength{\tabcolsep}{4pt}
\begin{tabular}{@{}lccccc@{}}
\hline
Model & AUC & Accuracy & Sensitivity & Specificity & F1 score \\
\hline
FOCUS & \textbf{1.000 (1.000--1.000)} & \underline{0.996 (0.985--1.000)} & \underline{0.996 (0.984--1.000)} & \textbf{1.000 (1.000--1.000)} & \underline{0.998 (0.992--1.000)} \\
H-Optimus-1 & \textbf{1.000 (1.000--1.000)} & \textbf{1.000 (1.000--1.000)} & \textbf{1.000 (1.000--1.000)} & \textbf{1.000 (1.000--1.000)} & \textbf{1.000 (1.000--1.000)} \\
H-Optimus-0 & \textbf{1.000 (1.000--1.000)} & \textbf{1.000 (1.000--1.000)} & \textbf{1.000 (1.000--1.000)} & \textbf{1.000 (1.000--1.000)} & \textbf{1.000 (1.000--1.000)} \\
GigaPath & \textbf{1.000 (1.000--1.000)} & \underline{0.996 (0.985--1.000)} & \underline{0.996 (0.984--1.000)} & \textbf{1.000 (1.000--1.000)} & \underline{0.998 (0.992--1.000)} \\
Virchow2 & \textbf{1.000 (1.000--1.000)} & \underline{0.996 (0.985--1.000)} & \underline{0.996 (0.984--1.000)} & \textbf{1.000 (1.000--1.000)} & \underline{0.998 (0.992--1.000)} \\
Virchow & \textbf{1.000 (1.000--1.000)} & \textbf{1.000 (1.000--1.000)} & \textbf{1.000 (1.000--1.000)} & \textbf{1.000 (1.000--1.000)} & \textbf{1.000 (1.000--1.000)} \\
UNI & \textbf{1.000 (1.000--1.000)} & \textbf{1.000 (1.000--1.000)} & \textbf{1.000 (1.000--1.000)} & \textbf{1.000 (1.000--1.000)} & \textbf{1.000 (1.000--1.000)} \\
Phikon-v2 & \textbf{1.000 (1.000--1.000)} & \textbf{1.000 (1.000--1.000)} & \textbf{1.000 (1.000--1.000)} & \textbf{1.000 (1.000--1.000)} & \textbf{1.000 (1.000--1.000)} \\
\hline
\end{tabular}
\par\smallskip
\noindent\parbox{\textwidth}{\fontsize{6.5}{7.8}\selectfont Values are estimate (95\% CI). External predictions were used when available, otherwise test predictions. Binary AUC CIs use DeLong's method; all other CIs use 2,000-patient bootstrap percentile intervals. Multiclass Sensitivity, Specificity, PPV, and NPV are macro-averaged. Bold and underlining indicate the highest and second-highest values, respectively.}\end{table}

\begin{table}[p]
\centering
\caption{External-set metrics for LUAD/LUSC (Histological subtype).}
\label{tab:dsmil_metrics_luad_lusc_histological_subtype}
\fontsize{8.0}{9.5}\selectfont
\setlength{\tabcolsep}{4pt}
\begin{tabular}{@{}lccccc@{}}
\hline
Model & AUC & Accuracy & Sensitivity & Specificity & F1 score \\
\hline
FOCUS & 0.862 (0.840--0.884) & 0.762 (0.737--0.788) & \underline{0.635 (0.579--0.691)} & 0.808 (0.783--0.836) & 0.588 (0.540--0.632) \\
H-Optimus-1 & 0.875 (0.854--0.896) & 0.825 (0.802--0.847) & 0.580 (0.522--0.638) & 0.914 (0.894--0.932) & \textbf{0.639 (0.590--0.685)} \\
H-Optimus-0 & \textbf{0.884 (0.864--0.904)} & 0.830 (0.808--0.852) & 0.430 (0.373--0.488) & 0.975 (0.964--0.985) & 0.574 (0.514--0.629) \\
GigaPath & 0.872 (0.851--0.894) & \textbf{0.840 (0.820--0.862)} & 0.444 (0.386--0.502) & \textbf{0.985 (0.976--0.993)} & \underline{0.598 (0.541--0.653)} \\
Virchow2 & 0.847 (0.823--0.870) & \underline{0.834 (0.812--0.857)} & 0.423 (0.366--0.483) & \underline{0.984 (0.975--0.992)} & 0.577 (0.519--0.633) \\
Virchow & \underline{0.883 (0.863--0.903)} & 0.813 (0.791--0.837) & 0.352 (0.296--0.407) & 0.981 (0.971--0.990) & 0.501 (0.438--0.559) \\
UNI & 0.878 (0.857--0.898) & 0.823 (0.801--0.847) & 0.382 (0.325--0.439) & \underline{0.984 (0.975--0.992)} & 0.536 (0.474--0.593) \\
Phikon-v2 & 0.841 (0.817--0.865) & 0.683 (0.655--0.711) & \textbf{0.805 (0.758--0.850)} & 0.638 (0.607--0.672) & 0.576 (0.533--0.615) \\
\hline
\end{tabular}
\par\smallskip
\noindent\parbox{\textwidth}{\fontsize{6.5}{7.8}\selectfont Values are estimate (95\% CI). External predictions were used when available, otherwise test predictions. Binary AUC CIs use DeLong's method; all other CIs use 2,000-patient bootstrap percentile intervals. Multiclass Sensitivity, Specificity, PPV, and NPV are macro-averaged. Bold and underlining indicate the highest and second-highest values, respectively.}\end{table}
\begin{table}[p]
\centering
\caption{External-set metrics for LUNG (Benign vs. malignant).}
\label{tab:dsmil_metrics_lung_benign_malignant}
\fontsize{8.0}{9.5}\selectfont
\setlength{\tabcolsep}{4pt}
\begin{tabular}{@{}lccccc@{}}
\hline
Model & AUC & Accuracy & Sensitivity & Specificity & F1 score \\
\hline
FOCUS & 0.992 (0.985--0.998) & \textbf{0.987 (0.980--0.994)} & \textbf{0.991 (0.983--0.997)} & \underline{0.981 (0.967--0.993)} & \textbf{0.990 (0.984--0.995)} \\
H-Optimus-1 & \underline{0.992 (0.986--0.998)} & \underline{0.985 (0.978--0.992)} & \underline{0.990 (0.981--0.997)} & 0.979 (0.965--0.991) & \underline{0.988 (0.982--0.993)} \\
H-Optimus-0 & \textbf{0.993 (0.987--0.998)} & 0.983 (0.974--0.990) & 0.988 (0.979--0.995) & 0.974 (0.958--0.988) & 0.986 (0.979--0.992) \\
GigaPath & 0.989 (0.982--0.997) & 0.975 (0.965--0.985) & 0.982 (0.971--0.991) & 0.965 (0.946--0.981) & 0.980 (0.972--0.987) \\
Virchow2 & 0.991 (0.985--0.998) & \underline{0.985 (0.978--0.992)} & 0.985 (0.975--0.994) & \textbf{0.986 (0.973--0.995)} & 0.988 (0.982--0.993) \\
Virchow & 0.992 (0.985--0.998) & 0.983 (0.974--0.990) & \textbf{0.991 (0.983--0.997)} & 0.970 (0.952--0.984) & 0.986 (0.979--0.992) \\
UNI & 0.989 (0.982--0.997) & 0.976 (0.967--0.985) & 0.985 (0.976--0.994) & 0.963 (0.944--0.980) & 0.981 (0.973--0.988) \\
Phikon-v2 & 0.989 (0.981--0.997) & 0.967 (0.956--0.978) & \underline{0.990 (0.981--0.997)} & 0.932 (0.907--0.956) & 0.973 (0.964--0.982) \\
\hline
\end{tabular}
\par\smallskip
\noindent\parbox{\textwidth}{\fontsize{6.5}{7.8}\selectfont Values are estimate (95\% CI). External predictions were used when available, otherwise test predictions. Binary AUC CIs use DeLong's method; all other CIs use 2,000-patient bootstrap percentile intervals. Multiclass Sensitivity, Specificity, PPV, and NPV are macro-averaged. Bold and underlining indicate the highest and second-highest values, respectively.}\end{table}
\begin{table}[p]
\centering
\caption{External-set metrics for KIRC/KIRP/KICH (Histological subtype).}
\label{tab:dsmil_metrics_kirc_kirp_kich_histological_subtype}
\fontsize{8.0}{9.5}\selectfont
\setlength{\tabcolsep}{4pt}
\begin{tabular}{@{}lccccc@{}}
\hline
Model & AUC & Accuracy & Sensitivity & Specificity & F1 score \\
\hline
FOCUS & \underline{0.964 (0.947--0.980)} & 0.920 (0.893--0.943) & 0.500 (0.340--0.658) & 0.960 (0.940--0.978) & 0.519 (0.369--0.651) \\
H-Optimus-1 & 0.960 (0.942--0.978) & \underline{0.928 (0.902--0.950)} & 0.425 (0.273--0.585) & \textbf{0.976 (0.961--0.990)} & 0.507 (0.346--0.648) \\
H-Optimus-0 & 0.961 (0.944--0.979) & \textbf{0.937 (0.913--0.959)} & 0.575 (0.415--0.729) & \underline{0.971 (0.955--0.986)} & 0.613 (0.464--0.733) \\
GigaPath & 0.935 (0.912--0.959) & 0.883 (0.850--0.911) & \underline{0.725 (0.578--0.857)} & 0.898 (0.867--0.927) & 0.518 (0.393--0.625) \\
Virchow2 & 0.950 (0.930--0.970) & 0.907 (0.878--0.933) & 0.575 (0.419--0.730) & 0.938 (0.914--0.961) & 0.517 (0.376--0.632) \\
Virchow & 0.943 (0.919--0.966) & 0.909 (0.880--0.935) & 0.700 (0.550--0.837) & 0.929 (0.904--0.953) & 0.571 (0.442--0.674) \\
UNI & \textbf{0.965 (0.949--0.981)} & 0.917 (0.891--0.941) & \textbf{0.850 (0.733--0.952)} & 0.924 (0.898--0.948) & \textbf{0.642 (0.524--0.737)} \\
Phikon-v2 & 0.956 (0.936--0.976) & 0.926 (0.900--0.948) & 0.700 (0.559--0.833) & 0.948 (0.925--0.967) & \underline{0.622 (0.494--0.727)} \\
\hline
\end{tabular}
\par\smallskip
\noindent\parbox{\textwidth}{\fontsize{6.5}{7.8}\selectfont Values are estimate (95\% CI). External predictions were used when available, otherwise test predictions. Binary AUC CIs use DeLong's method; all other CIs use 2,000-patient bootstrap percentile intervals. Multiclass Sensitivity, Specificity, PPV, and NPV are macro-averaged. Bold and underlining indicate the highest and second-highest values, respectively.}\end{table}
\begin{table}[p]
\centering
\caption{External-set metrics for KIRC/KIRP/KICH (Tumour grading).}
\label{tab:dsmil_metrics_kirc_kirp_kich_tumor_grading}
\fontsize{8.0}{9.5}\selectfont
\setlength{\tabcolsep}{4pt}
\begin{tabular}{@{}lccccc@{}}
\hline
Model & AUC & Accuracy & Sensitivity & Specificity & F1 score \\
\hline
FOCUS & \underline{0.616 (0.572--0.660)} & \underline{0.510 (0.462--0.560)} & \underline{0.277 (0.254--0.305)} & \underline{0.771 (0.757--0.787)} & \underline{0.437 (0.387--0.495)} \\
H-Optimus-1 & 0.614 (0.571--0.658) & 0.502 (0.452--0.552) & 0.268 (0.248--0.289) & 0.767 (0.753--0.782) & 0.429 (0.374--0.484) \\
H-Optimus-0 & 0.598 (0.556--0.643) & 0.507 (0.457--0.555) & 0.274 (0.251--0.299) & 0.765 (0.751--0.780) & 0.428 (0.372--0.484) \\
GigaPath & 0.541 (0.498--0.587) & 0.500 (0.455--0.548) & 0.260 (0.243--0.277) & 0.762 (0.750--0.774) & 0.403 (0.347--0.460) \\
Virchow2 & 0.587 (0.543--0.632) & 0.493 (0.445--0.543) & 0.262 (0.243--0.287) & 0.761 (0.749--0.774) & 0.403 (0.348--0.461) \\
Virchow & 0.609 (0.567--0.654) & \underline{0.510 (0.462--0.560)} & 0.268 (0.249--0.287) & 0.761 (0.748--0.774) & 0.417 (0.365--0.473) \\
UNI & 0.613 (0.571--0.658) & 0.495 (0.448--0.543) & 0.269 (0.246--0.295) & 0.770 (0.755--0.785) & 0.429 (0.374--0.483) \\
Phikon-v2 & \textbf{0.660 (0.621--0.700)} & \textbf{0.533 (0.486--0.581)} & \textbf{0.304 (0.274--0.332)} & \textbf{0.787 (0.770--0.803)} & \textbf{0.479 (0.427--0.531)} \\
\hline
\end{tabular}
\par\smallskip
\noindent\parbox{\textwidth}{\fontsize{6.5}{7.8}\selectfont Values are estimate (95\% CI). External predictions were used when available, otherwise test predictions. Binary AUC CIs use DeLong's method; all other CIs use 2,000-patient bootstrap percentile intervals. Multiclass Sensitivity, Specificity, PPV, and NPV are macro-averaged. Bold and underlining indicate the highest and second-highest values, respectively.}\end{table}
\begin{table}[p]
\centering
\caption{External-set metrics for KIRC/KIRP/KICH (Benign vs. malignant).}
\label{tab:dsmil_metrics_kirc_kirp_kich_benign_malignant}
\fontsize{8.0}{9.5}\selectfont
\setlength{\tabcolsep}{4pt}
\begin{tabular}{@{}lccccc@{}}
\hline
Model & AUC & Accuracy & Sensitivity & Specificity & F1 score \\
\hline
FOCUS & \textbf{1.000 (0.999--1.000)} & \textbf{0.994 (0.987--0.999)} & \textbf{0.996 (0.989--1.000)} & \underline{0.991 (0.976--1.000)} & \textbf{0.996 (0.990--0.999)} \\
H-Optimus-1 & 0.999 (0.999--1.000) & 0.979 (0.968--0.990) & 0.972 (0.955--0.986) & \textbf{0.995 (0.984--1.000)} & 0.985 (0.975--0.992) \\
H-Optimus-0 & 0.999 (0.997--1.000) & 0.975 (0.963--0.985) & 0.978 (0.964--0.990) & 0.968 (0.943--0.990) & 0.981 (0.972--0.989) \\
GigaPath & 0.998 (0.996--0.999) & 0.965 (0.950--0.978) & 0.954 (0.934--0.973) & 0.986 (0.967--1.000) & 0.973 (0.962--0.983) \\
Virchow2 & \underline{1.000 (0.999--1.000)} & \underline{0.991 (0.984--0.997)} & \textbf{0.996 (0.989--1.000)} & 0.982 (0.961--0.996) & \underline{0.993 (0.988--0.998)} \\
Virchow & 0.999 (0.998--1.000) & 0.978 (0.966--0.988) & 0.974 (0.958--0.987) & 0.986 (0.970--1.000) & 0.984 (0.975--0.991) \\
UNI & 0.999 (0.999--1.000) & 0.984 (0.974--0.993) & \underline{0.993 (0.985--1.000)} & 0.964 (0.937--0.986) & 0.988 (0.981--0.995) \\
Phikon-v2 & 0.997 (0.995--0.999) & 0.966 (0.953--0.978) & 0.980 (0.967--0.992) & 0.936 (0.900--0.966) & 0.975 (0.965--0.984) \\
\hline
\end{tabular}
\par\smallskip
\noindent\parbox{\textwidth}{\fontsize{6.5}{7.8}\selectfont Values are estimate (95\% CI). External predictions were used when available, otherwise test predictions. Binary AUC CIs use DeLong's method; all other CIs use 2,000-patient bootstrap percentile intervals. Multiclass Sensitivity, Specificity, PPV, and NPV are macro-averaged. Bold and underlining indicate the highest and second-highest values, respectively.}\end{table}
\endgroup

\begingroup
\renewcommand{\arraystretch}{1.16}
\begin{table}[p]
\centering
\caption{Five-fold patch-level classification performance.}
\label{tab:patchcls-fold-results}
\SuppTableStandard
\setlength{\tabcolsep}{4pt}
\begin{tabular}{@{}lccc@{}}
\toprule
Model & accuracy & Macro-F1 (\%) & Balanced accuracy \\
\midrule
\textbf{FOCUS} & \textbf{73.04 $\pm$ 0.96} & \textbf{72.82 $\pm$ 0.99} & \textbf{73.04 $\pm$ 0.96} \\
UNI & \underline{70.84 $\pm$ 0.86} & \underline{70.61 $\pm$ 0.86} & \underline{70.84 $\pm$ 0.86} \\
Virchow2 & 70.68 $\pm$ 0.83 & 70.38 $\pm$ 0.85 & 70.68 $\pm$ 0.83 \\
Virchow & 70.23 $\pm$ 0.41 & 70.00 $\pm$ 0.42 & 70.23 $\pm$ 0.41 \\
H-Optimus-0 & 70.06 $\pm$ 0.98 & 69.81 $\pm$ 0.98 & 70.06 $\pm$ 0.98 \\
GigaPath & 69.99 $\pm$ 0.63 & 69.80 $\pm$ 0.62 & 69.99 $\pm$ 0.63 \\
CRISP & 70.28 $\pm$ 0.69 & 70.07 $\pm$ 0.67 & 70.28 $\pm$ 0.69 \\
Phikon-v2 & 69.48 $\pm$ 0.97 & 69.28 $\pm$ 0.99 & 69.48 $\pm$ 0.97 \\
H-Optimus-1 & 69.81 $\pm$ 0.64 & 69.56 $\pm$ 0.64 & 69.81 $\pm$ 0.64 \\
\bottomrule
\end{tabular}
\par\smallskip
\noindent\parbox{\textwidth}{\SuppTableNotes Values are mean $\pm$ SD across five cross-validation folds. All models used the same SGD linear-probe configuration (learning rate $5\times10^{-5}$, weight decay $10^{-4}$, 200 epochs, batch size 512, z-score normalization). Each fold used 14,400 training and 3,600 validation samples. Bold indicates the highest value and underlining the second-highest value for each metric.}
\end{table}
\endgroup

\begingroup
\renewcommand{\arraystretch}{1.16}
\begin{table}[p]
\centering
\caption{Task-level patch-size robustness of FOCUS.}
\label{tab:multisize-stability}
\SuppTableCompact
\setlength{\tabcolsep}{3pt}
\begin{tabular}{@{}lccccc@{}}
\toprule
Task & AUC, 224 px & AUC, 448 px & AUC, 896 px & $\Delta$AUC, 448--224 px & $\Delta$AUC, 896--224 px \\
\midrule
T1  & 0.964 & 0.971 & 0.971 & +0.007 & +0.007 \\
T2  & 0.995 & 0.997 & 0.993 & +0.001 & -0.003 \\
T3  & 0.992 & 0.988 & 0.983 & -0.004 & -0.009 \\
T4  & 0.987 & 0.985 & 0.981 & -0.002 & -0.006 \\
T5  & 0.898 & 0.917 & 0.935 & +0.018 & +0.037 \\
T6  & 0.992 & 0.997 & 0.987 & +0.005 & -0.005 \\
T7  & 0.977 & 0.986 & 0.986 & +0.009 & +0.009 \\
T8  & 0.956 & 0.966 & 0.953 & +0.010 & -0.003 \\
T9  & 0.932 & 0.938 & 0.935 & +0.006 & +0.003 \\
T10 & 0.988 & 0.988 & 0.984 & +0.000 & -0.004 \\
T11 & 0.973 & 0.972 & 0.970 & -0.001 & -0.003 \\
\midrule
T12 & 0.834 & 0.817 & 0.823 & -0.017 & -0.012 \\
T13 & 0.967 & 0.968 & 0.962 & +0.001 & -0.005 \\
T14 & 0.905 & 0.904 & 0.897 & -0.001 & -0.008 \\
T15 & 0.981 & 0.975 & 0.976 & -0.006 & -0.005 \\
T16 & 0.836 & 0.860 & 0.840 & +0.024 & +0.004 \\
T17 & 0.823 & 0.839 & 0.830 & +0.016 & +0.007 \\
T18 & 0.866 & 0.862 & 0.862 & -0.005 & -0.005 \\
T19 & 0.955 & 0.952 & 0.950 & -0.003 & -0.005 \\
T20 & 0.736 & 0.734 & 0.723 & -0.002 & -0.012 \\
T21 & 0.884 & 0.881 & 0.862 & -0.004 & -0.022 \\
T22 & 0.855 & 0.850 & 0.832 & -0.005 & -0.023 \\
T23 & 0.877 & 0.893 & 0.882 & +0.016 & +0.004 \\
\midrule
\textbf{Mean} & \textbf{0.921} & \textbf{0.923} & \textbf{0.918} & \textbf{+0.003} & \textbf{-0.002} \\
\textbf{$P$} & -- & -- & -- & 0.321 & 0.275 \\
\bottomrule
\end{tabular}
\par\smallskip
\noindent\parbox{\textwidth}{\SuppTableNotes
 $\Delta$AUC is the within-task difference from 224 px. The mean changes were +0.003 (bootstrap 95\% CI, -0.001 to +0.007) for 448 px and -0.002 (bootstrap 95\% CI, -0.007 to +0.003) for 896 px. $P$ values are two-sided paired Wilcoxon signed-rank tests across 23 tasks.}
\end{table}
\endgroup

\begingroup
\renewcommand{\arraystretch}{1.13}
\begin{table}[p]
\centering
\caption{Matched-task data-pipeline ablation results for FOCUS.}
\label{tab:ablation-results}
\SuppTableDense
\setlength{\tabcolsep}{2.4pt}

\begin{tabular}{@{}p{3.55cm}cccc@{}}
\toprule
Contrast (Full FOCUS minus counterpart) & Tasks with $\Delta$AUC $>0$ & Mean $\Delta$AUC & Bootstrap 95\% CI & Wilcoxon $P$ \\
\midrule
FAMV & 17/23 & +0.0012 & +0.0000 to +0.0022 & 0.0327 \\
HCDC & 22/23 & +0.0125 & +0.0085 to +0.0173 & $<0.0001$ \\
\bottomrule
\end{tabular}

\par\medskip

\begin{tabular}{@{}lccccc@{}}
\toprule
Task & Full FOCUS & FAMV & HCDC & $\Delta$AUC FAMV & $\Delta$AUC HCDC \\
\midrule
T1 & 0.9642 & 0.9658 & 0.9603 & -0.0016 & +0.0038 \\
T2 & 0.9952 & 0.9959 & 0.9871 & -0.0008 & +0.0081 \\
T3 & 0.9920 & 0.9910 & 0.9905 & +0.0010 & +0.0015 \\
T4 & 0.9873 & 0.9869 & 0.9758 & +0.0004 & +0.0116 \\
T5 & 0.8985 & 0.8928 & 0.8823 & +0.0057 & +0.0162 \\
T6 & 0.9919 & 0.9885 & 0.9850 & +0.0035 & +0.0069 \\
T7 & 0.9771 & 0.9763 & 0.9706 & +0.0008 & +0.0065 \\
T8 & 0.9559 & 0.9620 & 0.9569 & -0.0061 & -0.0011 \\
T9 & 0.9319 & 0.9316 & 0.9185 & +0.0003 & +0.0133 \\
T10 & 0.9878 & 0.9860 & 0.9720 & +0.0017 & +0.0157 \\
T11 & 0.9729 & 0.9686 & 0.9540 & +0.0043 & +0.0190 \\
\midrule
T12 & 0.8343 & 0.8316 & 0.8342 & +0.0028 & +0.0002 \\
T13 & 0.9675 & 0.9681 & 0.9544 & -0.0006 & +0.0131 \\
T14 & 0.9047 & 0.9018 & 0.8559 & +0.0029 & +0.0488 \\
T15 & 0.9812 & 0.9762 & 0.9697 & +0.0051 & +0.0116 \\
T16 & 0.8359 & 0.8349 & 0.8285 & +0.0010 & +0.0074 \\
T17 & 0.8231 & 0.8256 & 0.8136 & -0.0025 & +0.0095 \\
T18 & 0.8664 & 0.8663 & 0.8483 & +0.0001 & +0.0181 \\
T19 & 0.9549 & 0.9561 & 0.9453 & -0.0012 & +0.0095 \\
T20 & 0.7356 & 0.7317 & 0.7213 & +0.0039 & +0.0143 \\
T21 & 0.8845 & 0.8830 & 0.8568 & +0.0014 & +0.0277 \\
T22 & 0.8553 & 0.8518 & 0.8549 & +0.0034 & +0.0003 \\
T23 & 0.8772 & 0.8761 & 0.8519 & +0.0011 & +0.0254 \\
\bottomrule
\end{tabular}

\par\smallskip
\noindent\parbox{\textwidth}{\SuppTableNotes
All entries are test AUCs; positive $\Delta$AUC values favour full FOCUS. Each contrast includes the 23 tasks with results available for all three supplied configurations. The confidence intervals are percentile task-bootstrap 95\% intervals from 2,000 resamples; $P$ values are two-sided paired Wilcoxon signed-rank tests.}
\end{table}
\endgroup

\end{document}